\documentclass[acmsmall, noncam, screen, authorversion, natbib=false]{acmart}
\pdfoutput=1

\usepackage[utf8]{inputenc}
\usepackage[T1]{fontenc}

\usepackage{pdflscape}

\usepackage{hyperref}
\hypersetup{pagebackref}

\newcommand*{\fullref}[1]{\hyperref[{#1}]{\nameref*{#1} \S\ref{#1}}} 

\usepackage{color}
\usepackage{booktabs}
\usepackage{tabularx}
\usepackage{multirow}
\usepackage{comment}

\usepackage{textcomp}
\usepackage{lipsum}

\usepackage{rotating, graphicx}
\usepackage[normalem]{ulem}

\usepackage{sidecap}
\sidecaptionvpos{figure}{c}

\usepackage{titlesec}
\titlespacing*{\paragraph}{2ex}{0ex}{1ex}[.]
\titlespacing*{\subsubsection}{0ex}{.5ex}{1ex}[.]
\titlespacing*{\subsection}{0ex}{1ex}{0em}
\usepackage{titletoc,tocloft}
\usepackage{enumitem}
\setlist[itemize]{noitemsep, topsep=0pt}
\setlist[itemize]{noitemsep, topsep=0pt}

\newcommand{\cliprow}[1]{\begin{tabular}[c]{@{}l@{}}#1\end{tabular}}

\usepackage[firstinits=true, maxbibnames=1,maxcitenames=1,doi=false,url=false,isbn=false,backend=bibtex]{biblatex}
\renewbibmacro{in:}{}
\AtEveryBibitem{\clearfield{pages}}
\AtEveryBibitem{\clearfield{volume}}
\AtEveryBibitem{\clearfield{number}}
\AtEveryBibitem{\clearfield{howpublished}}

\usepackage{xr}
\usepackage{filecontents}

\begin{document}

\title{Trustworthy AI: From Principles to Practices}
\author{Bo Li}

\author{Peng Qi}

\author{Bo Liu}

\author{Shuai Di}

\author{Jingen Liu}

\author{Jiquan Pei}

\author{Jinfeng Yi}

\author{Bowen Zhou}
\authorsaddresses{Contacts: Bo~Li, libo427@jd.com; Peng~Qi, peng.qi@jd.com; Bo~Liu, bo.liu2@jd.com; Shuai~Di, dishuai@jd.com; Jingen~Liu, jingen.liu@jd.com; Jiquan~Pei, peijiquan@jd.com; Jinfeng~Yi, yijinfeng@jd.com; Bowen~Zhou (*corresponding author), zhoubowen@jd.com, zhoubw@gmail.com.}

\begin{abstract}

The rapid development of Artificial Intelligence (AI) technology has enabled the deployment of various systems based on it. However, many current AI systems are found vulnerable to imperceptible attacks, biased against underrepresented groups, lacking in user privacy protection. These shortcomings degrade user experience and erode people's trust in all AI systems. In this review, we provide AI practitioners with a comprehensive guide for building trustworthy AI systems. We first introduce the theoretical framework of important aspects of AI trustworthiness, including robustness, generalization, explainability, transparency, reproducibility, fairness, privacy preservation, and accountability.
To unify currently available but fragmented approaches toward trustworthy AI, we organize them in a systematic approach that considers the entire lifecycle of AI systems, ranging from data acquisition to model development, to system development and deployment, finally to continuous monitoring and governance. In this framework, we offer concrete action items for practitioners and societal stakeholders (e.g., researchers, engineers, and regulators) to improve AI trustworthiness. Finally, we identify key opportunities and challenges for the future development of trustworthy AI systems, where we identify the need for a paradigm shift toward comprehensively trustworthy AI systems.

\end{abstract}

\keywords{trustworthy AI, robustness, generalization, explainability, transparency, reproducibility, fairness, privacy protection, accountability}

\ccsdesc[500]{Computing methodologies~Artificial intelligence}
\ccsdesc[500]{General and reference~Surveys and overviews}
\ccsdesc[500]{Computing methodologies~Machine learning}
\maketitle


\section{Introduction}\label{sec:introduction}



The rapid development of Artificial Intelligence (AI) continues to provide significant economic and social benefits to society.
With the widespread application of AI in areas such as transportation, finance, medicine, security, and entertainment, there is rising societal awareness that we need these systems to be trustworthy.
This is because the breach of stakeholders' trust can lead to severe societal consequences given the pervasiveness of these AI systems.
Such breaches can range from biased treatment by automated systems in hiring and loan decisions \cite{bogen2018help, hao2019ai} to the loss of human life \cite{boudette2021happened}.
By contrast, AI practitioners, including researchers, developers, and decision-makers, have traditionally considered system performance (i.e., accuracy) to be the main metric in their workflows.
This metric is far from sufficient to reflect the trustworthiness of AI systems. 
Various aspects of AI systems beyond system performance should be considered to improve their trustworthiness, including but not limited to their robustness, algorithmic fairness, explainability, and transparency.

While most active academic research on AI trustworthiness has focused on the algorithmic properties of models, advancements in algorithmic research alone is insufficient for building trustworthy AI products. From an industrial perspective, the lifecycle of an AI product consists of multiple stages, including data preparation, algorithmic design, development, and deployment as well as operation, monitoring, and governance. Improving trustworthiness in any single aspect (e.g., robustness) involves efforts at multiple stages in this lifecycle, e.g., data sanitization, robust algorithms, anomaly monitoring, and risk auditing. On the contrary, the breach of trust in any single link or aspect can undermine the trustworthiness of the entire system. Therefore, AI trustworthiness should be established and assessed systematically throughout the lifecycle of an AI system. 

In addition to taking a holistic view of the trustworthiness of AI systems over all stages of their lifecycle, it is important to understand the big picture of different aspects of AI trustworthiness. In addition to pursuing AI trustworthiness by establishing requirements for each specific aspect, we call attention to the combination of and interaction between these aspects, which are important and underexplored topics for trustworthy real-world AI systems.
For instance, the need for data privacy might interfere with the desire to explain the system output in detail, and the pursuit of algorithmic fairness may be detrimental to the accuracy and robustness experienced by some groups~\cite{xu2021robust, roh2020fr}.
As a result, trivially combining systems to separately improve each aspect of trustworthiness does not guarantee a more trustworthy and effective end result.
Instead, elaborated joint optimization and trade-offs between multiple aspects of trustworthiness are necessary~\cite{xu2021robust, hong2021federated, tsipras103robustness, zhang2019theoretically, bodria2021benchmarking}.

These facts suggest that a systematic approach is necessary to shift the current AI paradigm toward trustworthiness.
This requires awareness and cooperation from multi-disciplinary stakeholders who work on different aspects of trustworthiness and different stages of the system’s lifecycle. We have recently witnessed important developments in multi-disciplinary research on trustworthy AI. From the perspective of technology, trustworthy AI has promoted the development of adversarial learning, private learning, and the fairness and explainability of machine learning (ML) . Some recent studies have organized these developments from the perspective of either research~\cite{liu2021trustworthy, wing2021trustworthy, kaur2020requirements} or engineering~\cite{cammarota2020trustworthy, brundage2020toward, varshney2019trustworthy, wickramasinghe2020trustworthy, kumar2020trustworthy}. Developments in non-technical areas have also been reviewed in a few studies, and involve guidelines~\cite{jobin2019global, schiff2021ai, hagendorff2020ethics}, standardization~\cite{lewis2021ontology}, and management processes~\cite{shneiderman2020bridging, baker2021trustworthy, rakova2021responsible}. We have conducted a detailed analysis of the various reviews, including algorithmic research, engineering practices, and institutionalization, in Section~\ref{sec:recent-multidisciplinary} in the appendix. These fragmented reviews have mostly focused on specific views of trustworthy AI.
To synchronize these diverse developments in a systematic view, we organize multi-disciplinary knowledge in an accessible manner for AI practitioners, and provide actionable and systematic guidance in the context of the lifecycle of an industrial system to build trustworthy AI systems.
Our main contributions are as follows:


\begin{itemize}
    \item We dissect the entire lifecycle of the development and deployment of AI systems in industrial applications, and discuss how AI trustworthiness can be enhanced at each stage---from data to AI models, and from system deployment to its operation. We propose a systematic framework to organize the multi-disciplinary and fragmented approaches toward trustworthy AI, and propose pursuing it as a continuous workflow to incorporate feedback at each stage of the lifecycle of the AI system.

    \item We dissect the entire development and deployment lifecycle of AI systems in industrial applications and discuss how AI trustworthiness can be enhanced at each stage -- from data to AI models, from system deployment to its operation. We propose a systematic framework to organize the multi-disciplinary and fragmented approaches towards trustworthy AI, and further propose to pursue AI trustworthiness as a continuous workflow to incorporate feedback at each stage of the AI system lifecycle.
    We also analyze the relationship between different aspects of trustworthiness in practice (mutual enhancement and, sometimes, trade-offs).
    The aim is to provide stakeholders of AI systems, such as researchers, developers, operators, and legal experts, with an accessible and comprehensive guide to quickly understand the approaches toward AI trustworthiness (Section \ref{sec:systematic-approach}).
    \item We discuss outstanding challenges facing trustworthy AI on which the research community and industrial practitioners should focus in the near future.
    We identify several key issues, including the need for a deeper and fundamental understanding of several aspects of AI trustworthiness (e.g., robustness, fairness, and explainability), the importance of user awareness, and the promotion of inter-disciplinary and international collaboration (Section \ref{sec:challenges-and-opportunities}).
\end{itemize}

With these contributions, we aim to provide the practitioners and stakeholders of AI systems not only with a comprehensive introduction to the foundations and future of AI trustworthiness, but also with an operational guidebook for how to construct AI systems that are trustworthy.

\begin{SCfigure*}
    \centering
    \includegraphics[width=0.7\linewidth]{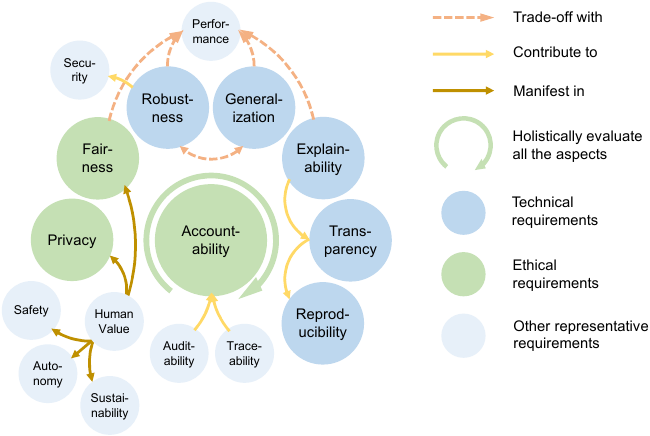}
    \caption{The relation between different aspects of AI trustworthiness discussed in this survey. Note that implicit interaction widely exists between aspects, and we cover only representative explicit interactions.}
    \label{fig:aspects}
\end{SCfigure*}

\section{AI Trustworthiness: Beyond Predictive Accuracy} \label{sec:definition-evaluation}


The success of ML technology in the last decades has largely benefited from the accuracy-based performance measurements.
By assessing task performance based on quantitative accuracy or loss, training AI models becomes tractable in the sense of optimization.
Meanwhile, predictive accuracy is widely adopted to indicate the superiority of an AI product over others. 
However, with the recent widespread applications of AI, the limitation of an accuracy-only measurement has been exposed to a number of new challenges, ranging from malicious attacks against AI systems to misuses of AI that violate human values.
To solve these problems, the AI community has realized in the last decade that factors beyond accuracy should be considered and improved when building an AI system.
A number of enterprises~\cite{google-ai, openai2018openai, ibm-trusting, cammarota2020trustworthy, brundage2020toward, varshney2019trustworthy}, academia~\cite{kumar2020trustworthy, thiebes2020trustworthy, floridi2019unified, shneiderman2020bridging,liu2021trustworthy}, public sectors, and organizations~\cite{aihleg2018ethics, union2017top, lewis2021ontology} have recently identified these factors and summarized them as principles of AI trustworthiness.
They include robustness, security, transparency, fairness, and safety~\cite{jobin2019global}.
Comprehensive statistics relating to and comparisons between these principles have been provided in~\cite{jobin2019global, hagendorff2020ethics}. In this paper, we study the representative principles that have recently garnered wide interest and are closely related to practical applications.
These principles can be categorized as follows:
\begin{itemize}
    \item We consider representative requirements that pertain to technical challenges faced by current AI systems. We review aspects that have sparked wide interest in recent technical studies, including robustness, explainability, transparency\footnote{We follow~\cite{turilli2009ethics} to exclude transparency as an ethical requirement.}, reproducibility, and generalization.
    \item We consider ethical requirements with broad concerns in recent literature~\cite{jobin2019global, aihleg2018ethics, floridi2019establishing, brundage2020toward, liu2021trustworthy, kumar2020trustworthy, shneiderman2020bridging, union2017top, hagendorff2020ethics}, including fairness, privacy, and accountability.
\end{itemize}
In this section, we illustrate the motivation for and definition of each requirement. We also survey approaches to the evaluation of each requirement.
It should also be noted that the selected requirements are not orthogonal, and some of them are closely correlated. We explain the relationships with the corresponding requirements in this section. We also use Figure~\ref{fig:aspects} to visualize the relationships between aspects, including trade-offs, contributions, and manifestation. 

\subsection{Robustness}\label{sec:robustness}
In general, robustness refers to the ability of an algorithm or system to deal with execution errors, erroneous inputs, or unseen data. Robustness is an important factor affecting the performance of AI systems in empirical environments. The lack of robustness might also cause unintended or harmful behavior by the system, thus diminishing its safety and trustworthiness. 
In the context of ML systems, the term robustness is applicable to a diversity of situations.
In this review, we non-exhaustively summarize the robustness of an AI system by categorizing vulnerabilities at the levels of data, algorithms, and systems, respectively.


\textbf{Data.} With the widespread application of AI systems, the environment in which an AI model is deployed becomes more complicated and diverse. If an AI model is trained without considering the diverse distributions of data in different scenarios, its performance might be significantly affected. Robustness against distributional shifts has been a common problem in various applications of AI~\cite{amodei2016concrete}. In high-stake applications, this problem is even more critical owing to its negative effect on safety and security. For example, in the field of autonomous driving, besides developing a perceptual system working in sunny scenes, academia and the industry are using numerous development and testing strategies to enhance the perceptual performance of the vehicles in nighttime/rainy scenes to guarantee the system’s reliability under a variety of weather conditions~\cite{zhang2018deeproad, tan2020night}.

\textbf{Algorithms.} 
It is widely recognized that AI models might be vulnerable to attacks by adversaries with malicious intentions. Among the various forms of attacks, the adversarial attack and defenses against it have raised concerns in both academia and the industry in recent years.
Literature has categorized the threat of adversarial attacks in several typical aspects and proposed various defense approaches~\cite{chakraborty2018adversarial, silva2020opportunities, akhtar2018threat, yuan2019adversarial, li2020backdoor}.
For example, in~\cite{vorobeychik2018adversarial}, adversarial attacks were categorized with respect to the attack timing.
\textit{Decision-time attack} perturbs input samples to mislead the prediction of a given model so that adversary could evade security checks or impersonate victims. \textit{Training-time attack} injects carefully designed samples into the training data to change the system's response to specific patterns, and is also known as \textit{poisoning attack}. 
Considering the practicality of attacks, it is also useful to note the differences of attacks in terms of the spaces in which they are carried out. Conventional studies have mainly focused on \textit{feature space attacks}, which are generated directly as the input features of a model.
In many practical scenarios, the adversaries can modify only the input entity to indirectly produce attack-related features. For example, it is easy for someone to wear adversarial pattern glasses to evade a face verification system but difficult to modify the image data in memory. 
Studies on producing realizable entity-based attacks (\textit{problem space attacks}) have recently garnered increasing interest~\cite{tong2019improving, wu2019defending}.
Algorithm-level threats might exist in various forms, in addition to directly misleading AI models. \textit{Model stealing} (a.k.a. the \textit{exploratory attack}) attempts to steal knowledge about models. Although it does not directly change model behavior, the stolen knowledge has significant value for generating adversarial samples~\cite{tramer2016stealing}. 


\textbf{Systems.} System-level robustness against illegal inputs should also be carefully considered in realistic AI products. The cases of illegal inputs can be extremely diverse in practical situations. For example, an image with a very high resolution might cause an imperfect image recognition system to hang. A lidar perception system for an autonomous vehicle might perceive laser beams emitted by lidars in other vehicles and produce corrupted inputs.
The \textit{presentation attack}~\cite{ramachandra2017presentation} (a.k.a. \textit{spoof attack}) is another example that has generated wide concerns in recent years. It fakes inputs by, for example, photos or masks to fool biometric systems.

Various approaches have been explored to prevent vulnerabilities in AI systems. The objective of defense can be either proactive or reactive~\cite{machado2021adversarial}. A proactive defense attempts to optimize the AI system to be more robust to various inputs while a reactive defense aims at detecting potential security issues, such as changing distributions or adversarial samples. Representative approaches to improve the robustness of an AI system are introduced in Section~\ref{sec:systematic-approach}.

\paragraph{Evaluation} Evaluating the robustness of an AI system serves as an important means of avoiding vulnerabilities and controlling risks. We briefly describe two groups of evaluations: robustness test and mathematical verification.


\textbf{Robustness test}. Testing has served as an essential approach to evaluate and enhance the robustness not only of conventional software, but also of AI systems. Conventional functional test methodologies, such as \textit{the monkey test}~\cite{exforsys2011what}, provide effective approaches to evaluating the system-level robustness. Moreover, as will be introduced in Section~\ref{ssec:testing}, software testing methodologies have recently been extended to evaluate robustness against adversarial attacks~\cite{pei2017deepxplore, ma2018deepgauge}. 

In comparison with functional test, performance test, i.e., benchmarking, are more widely adopted approach in the area of ML to evaluate system performance along various dimensions. Test datasets with various distributions are used to evaluate the robustness of data in ML research. 
In the context of adversarial attacks, the minimal adversarial perturbation is a core metric of robustness, and its empirical upper bound, a.k.a. empirical robustness, on a test dataset has been widely used~\cite{su2018robustness,carlini2017towards}. From the attacker's perspective, the rate of success of an attack also intuitively measures the robustness of the system~\cite{su2018robustness}.


\textbf{Mathematical verification.} 
Inherited from the theory of formal method, certified verification of the adversarial robustness of an AI model has led to growing interest in research on ML. For example, adversarial robustness can be reflected by deriving a non-trivial and certified lower bound of the minimum distortion to an attack on an AI model~\cite{zhang2018efficient, boopathy2019cnn}. We introduce this direction in Section~\ref{sec:adversarial-robustness}.

\subsection{Generalization}\label{sec:generalization}
Generalization has long been a source of concern in ML models. It represents the capability to distill knowledge from limited training data to make accurate predictions regarding unseen data~\cite{goodfellow2016machine}. Although generalization is not a frequently mentioned direction in the context of trustworthy AI, we find that its impact on AI trustworthiness should not be neglected and deserves specific discussion.  
On one hand, generalization requires that AI systems make predictions on realistic data, even on domains or distributions on which they are not trained~\cite{goodfellow2016machine}. This significantly affects the reliability and risk of practical systems. 
On the other hand, AI models should be able to generalize without the need to exhaustively collect and annotate large amounts of data for various domains~\cite{zhou2021domain, wang2021generalizing}, so that the deployment of AI systems in a wide range of applications is more affordable and sustainable.


In the field of ML, the canonical research on generalization theory has focused on the prediction of unseen data, which typically share the same distribution as the training data~\cite{goodfellow2016machine}. Although AI models can achieve a reasonable accuracy on training datasets, it is known that a gap (a.k.a generalization gap) exists between their training and testing accuracies. Approaches in different areas, ranging from statistic learning to deep learning, have been studied to analyze this problem and enhance the model generalization. Typical representatives like cross-validation, regularization, and data augmentation can be found in many ML textbooks~\cite{goodfellow2016machine}.

The creation of a modern data-driven AI model requires a large amount of data and annotations in the training stage. This leads to a high cost for manufacturers and users for re-collecting and re-annotating data to train the model for each task. The cost highlights the need to generalize the knowledge of a model to different tasks, which not only reduces data cost, but also improves model performance in many cases. Various directions of research have been explored to address knowledge generalization under different scenarios and configurations within the paradigm of transfer learning~\cite{pan2009survey, weiss2016survey}. We review the representative approaches in Section~\ref{sec:model-generalization}.

The inclusive concept of generalization is closely related to other aspects of AI trustworthiness, especially robustness. In the context of ML, the robustness against distributional shifts mentioned in Section~\ref{sec:robustness} is also considered a problem of generalization. This implies that the requirements of robustness and generalization have some overlapping aspects. 
The relationship between adversarial robustness and generalization is more complicated. As demonstrated in~\cite{xu2012robustness}, an algorithm that is robust against small perturbations has better generalization. Recent research~\cite{tsipras103robustness,raghunathan2019adversarial}, however, has noted that improving adversarial robustness by adversarial training may reduce the testing accuracy and leads to worse generalization. To explain this phenomenon, \cite{fawzi2018adversarial} has argued that the adversarial robustness corresponds to different data distributions that may hurt a model's generalization.


\paragraph{Evaluation} Benchmarking on test datasets with various distributions is a widely used approach to evaluate the generalization of an AI model in realistic scenarios. 
A summary of commonly used datasets and benchmarks for domain generalization can be found in~\cite{zhou2021domain}, and covers the tasks of object recognition, action recognition, segmentation, and face recognition. 

In terms of theoretical evaluation, past ML studies have developed rich approaches to measure the bounds of a model's generalization error. For example, Rademacher complexity \cite{bartlett2002rademacher} is commonly used to determine how well a model can fit a random assignment of class labels. In addition, the Vapnik-Chervonenkis (VC) dimension \cite{vapnik2013nature} is a measure of the capacity/complexity of a learnable function set. A larger number of VC dimensions indicates a higher capacity. 



Advances in the DNN has led to new developments in the theory of generalization. \cite{zhang2017understanding} observed that modern deep learning models can achieve a generalization gap despite their massive capacity. This phenomenon has led to academic discussions on the generalization of the DNN~\cite{arpit2017closer, belkin2019reconciling}. For example, \cite{belkin2019reconciling} examined generalization from the perspective of the bias--variance trade-off to explain and evaluate the generalization of the DNN.

\subsection{Explainability and Transparency}\label{sec:explainability-transparency}

The opaqueness of complex AI systems has led to widespread concerns in academia, the industry, and society at large. The problem of how DNNs outperform other conventional ML approaches has been puzzling researchers~\cite{arrieta2020explainable}. 
From the perspective of practical systems, there is a demand among users for the right to know the intention, business model, and technological mechanism of AI products~\cite{aihleg2018ethics, goodman2017european}.
A variety of studies have addressed these problems in terms of nomenclature including interpretability, explainability, and transparency~\cite{arrieta2020explainable, guidotti2018survey, bodria2021benchmarking, lipton2018mythos, molnar2020interpretable, adadi2018peeking}, and have delved into different definitions.
To make our discussion more concise and targeted, we narrow the coverage of explainability and transparency to address the above concerns in theoretical research and practical systems, respectively.

\begin{itemize}
    \item \textbf{Explainability} addresses to understand how an AI model makes decision~\cite{arrieta2020explainable}.
    \item \textbf{Transparency} considers AI as a software system, and seeks to disclose information regarding its entire lifecycle (c.f., ``operate transparently'' in~\cite{aihleg2018ethics}).
\end{itemize}

\subsubsection{Explainability}\label{sec:def-explainability}
Explainability, i.e., understanding how an AI model makes its decision, stays at the core place of modern AI research and serves as a fundamental factor that determines the trust in AI technology. The motivation for the explainability of AI comes from various aspects~\cite{arya2019one, arrieta2020explainable}. 
From the perspective of scientific research, it is meaningful to understand all intrinsic mechanisms of the data, parameters, procedures, and outcomes in an AI system. The mechanisms also fundamentally determine AI trustworthiness. From the perspective of building AI products, there exist various practical requirements on explainability.
For operators like bank executives, explainability helps understand the AI credit system to prevent potential defects in it~\cite{arya2019one, kim2018introduction}.
Users like loan applicants are interested to know why they are rejected by the model, and what they can do to qualify~\cite{arya2019one}.
See~\cite{arya2019one} for a detailed analysis of the various motivations of explainability.


Explaining ML models has been an active topic not only in ML research, but also in psychological research in the past five years~\cite{arrieta2020explainable, guidotti2018survey, bodria2021benchmarking, lipton2018mythos, molnar2020interpretable, adadi2018peeking}. Although the definition of the explainability of an AI model is still an open question, research has sought to address this problem from the perspectives of AI~\cite{rosenfeld2019explainability, guidotti2018survey} and psychology~\cite{gunning2019darpa, miller2019explanation}. In summary, the relevant studies divide explainability into two levels to explain it.

\begin{itemize}
    \item \textbf{Model explainability by design.} A series of fully or partially explainable ML models have been designed in the past half-century of ML research. Representatives include linear regression, trees, the KNN, rule-based learners, generalized additive model (GAM), and Bayesian models~\cite{arrieta2020explainable}. The design of explainable models remains an active area in ML.
    
    \item \textbf{Post-hoc model explainability.} Despite the good explainability of the above conventional models, more complex models such as the DNN or GDBT have exhibited better performance in recent industrial AI systems. Because the relevant approaches still cannot holistically explain these complex models, researchers have turned to post-hoc explanation. It addresses a model's behavior by analyzing its input, intermediate result, and output. A representative category in this vein approximates the decision surface either globally or locally by using an explainable ML model, i.e., \textit{explainer}, such as a linear model~\cite{ribeiro2016should, lundberg2017unified} and rules~\cite{ribeiro2018anchors, guidotti2018local}. For deep learning models like the CNN or the transformer, the inspection of intermediate features is a widely used means of explaining model behavior~\cite{tu2020select, yang2018hotpotqa}.
\end{itemize}
Approaches to explainability are an active area of work in ML and have been comprehensively surveyed in a variety of studies~\cite{arrieta2020explainable, guidotti2018survey, bodria2021benchmarking, molnar2020interpretable}. Representative algorithms to achieve the above two levels of explainability are reviewed in Section~\ref{sec:model-explainability}.


\paragraph{Evaluation}
In addition to the problem of explaining AI models, a unified evaluation of explainability has been recognized as a challenge.
A major reason for this lies in the ambiguity of the psychological outlining of explainability. To sidestep this problem, a variety of studies have used qualitative metrics to evaluate explainability with human participation. Representative approaches include the following:

\begin{itemize}
    \item \textbf{Subjective human evaluation.} The methods of evaluation in this context include interviews, self-reports, questionnaires, and case studies that measure, e.g., user satisfaction, mental models, and trust~\cite{gunning2019darpa, hoffman2018metrics, poursabzi2021manipulating}.
    \item \textbf{Human--AI task performance.} In tasks performed with human--AI collaboration, the collaborative performance is significantly affected by the human understanding of the AI collaborator, and can be viewed as a reflection of the quality of explanation ~\cite{mohseni2021multidisciplinary}. This evaluation has been used for the development of, for instance, recommendation systems~\cite{kulesza2012tell} and data analysis~\cite{goodall2018situ}.
\end{itemize}
In addition, if explainability can be achieved by an explainer, the performance of the latter, such as in terms of the precision of approximation (\textit{fidelity}~\cite{guidotti2018local, ribeiro2018anchors, ribeiro2016should}), can be used to indirectly and quantitatively evaluate explainability~\cite{alvarez2018towards}.

Despite the above evaluations, a direct quantitative measurement of explainability remains a problem. Some naive measurements of model complexity, like tree depth~\cite{blanco2020machine} and the size of the rule set ~\cite{lakkaraju2017interpretable}, have been studied as surrogate explainability metrics in previous work. We believe that a unified quantitative metric lies at the very heart of fundamental AI research. Recent research on the complexity of ML models~\cite{hu2021model} and their cognitive functional complexity~\cite{wang2003measurement} may inspire future research on a unified quantitative evaluation metric.


\subsubsection{Transparency}\label{sec:def-transparency}
Transparency requires the disclosure of the information on a system, and has long been a recognized requirement in software engineering~\cite{leite2010software, cysneiros2009initial}. In the AI industry, this requirement naturally covers the lifecycle of an AI system and helps stakeholders confirm that appropriate design principles are reflected in it. Consider a biometric system for identification as an example. Users are generally concerned with the purpose for which their biometric information is collected and how it is used. Business operators are concerned with the accuracy and robustness against attacks so that they can control risks. Government sectors are concerned with whether the AI system follows guidelines and regulations. On the whole, transparency serves as a basic requirement to build the public's trust in AI systems~\cite{jobin2019global, arnold2019factsheets, knowles2021sanction}. 

To render the lifecycle of an AI system transparent, a variety of information regarding its creation needs to be disclosed, including the design purposes, data sources, hardware requirements, configurations, working conditions, expected usage, and system performance. A series of studies have examined disclosing this information through appropriate documentation~\cite{arnold2019factsheets, gebru2021datasheets, holland2018dataset, mitchell2019model, piorkowski2020towards}. This is discussed in Section~\ref{sec:documentation}. The recent trend of open-source systems also significantly contributes to the algorithmic transparency of AI systems.

Owing to the complex and dynamic internal procedure of an AI system, facts regarding its creation are not sufficient in many cases to fully reveal its mechanism. Hence, the transparency of the runtime process and decision-making should also be considered in various scenarios. For an interactive AI system, an appropriately designed user interface serves as an important means to disclose the underlying decision procedure~\cite{aihleg2018assessment}. In many safety-critical systems, such as autonomous driving vehicles, logging systems~\cite{ayvaz2019witness, perez2010argos, yao2020smart} are widely adopted to trace and analyze the system execution.

\paragraph{Evaluation}
Although a unified quantitative evaluation is not yet available, the qualitative evaluation of transparency has undergone recent advances in the AI industry. Assessment checklists~\cite{aihleg2018assessment, schelenz2020applying} have been regarded as an effective means to evaluate and enhance the transparency of a system. In the context of the psychology of users or the public, user studies or A/B test can provide a useful evaluation based on user satisfaction~\cite{mohseni2021multidisciplinary}.

Quality evaluations of AI documentation have also been explored in recent years. Some studies~\cite{raji2019ml, arnold2019factsheets, gebru2021datasheets, holland2018dataset, mitchell2019model} have proposed standard practices to guide and evaluate the documentation of an AI system. \cite{piorkowski2020towards} summarized the general qualitative dimensions for more specified evaluations.





\subsection{Reproducibility}
Modern AI research involves both mathematical derivation and computational experiments. The reproducibility of these computational procedures serves as an essential step to verify AI research. In terms of AI trustworthiness, this verification facilitates the detection, analysis, and mitigation of potential risks in an AI system, such as a vulnerability on specific inputs or unintended bias.
With the gradual establishment of the open cooperative ecosystem in the AI research community, reproducibility is emerging as a concern among researchers and developers. In addition to enabling the effective verification of research, reproducibility allows the community to quickly convert the latest approaches into practice or conduct follow-up research.

There has been a new trend in the AI research community to regard reproducibility as a requirement when publicizing research~\cite{gundersen2018reproducible}. We have seen major conferences, such as NeurIPS, ICML, and ACMMM, introduce reproducibility-related policies or programs~\cite{pineau2021improving} to encourage the reproducibility of works. To obtain a clear assessment, degrees of reproducibility have been studied in such works as the ACM Artifact Review and~\cite{gundersen2018state, drummond2009replicability}. For example, in~\cite{gundersen2018state}, the lowest degree of reproducibility requires the exact replication of an experiment with the same implementation and data, while a higher degree requires using different implementations or data. Beyond the basic verification of research, a higher degree of reproducibility promotes a better understanding of the research by distinguishing among the key factors influencing effectiveness.

Some recently developed large scale pre-trained AI models, such as GPT-3 and BERT, are representative of the challenges to the reproducibility of AI research. The creation of these models involves specifically designed data collection strategies, efficient storage of big data, communication and scheduling between distributed clusters, algorithm implementation, appropriate software and hardware environments, and other kinds of knowhow. The reproducibility of such a model should be considered over its entire lifecycle. In recent studies on the reproducibility of ML, this requirement has been decomposed into the reproducibility of data, methods, and experiments~\cite{gundersen2018state, gundersen2018reproducible, isdahl2019out}, where the latter range over a series of lifecycle artifacts such as code, documentation, software, hardware, and deployment configuration. Based on this methodology, an increasing number of ML platforms are being developed that assist researchers and developers better track the lifecycle in a reproducible manner~\cite{isdahl2019out, zaharia2018accelerating}.

\paragraph{Evaluation}

Reproducibility checklists have been recently widely adopted in ML conferences to assess the reproducibility of submissions~\cite{pineau2021improving}. Beyond the replication of experiments in publication, \cite{gundersen2018reproducible, gundersen2018state} also specified checklists to evaluate reproducibility at varying degrees. In addition to checklists, mechanisms such as challenges to reproducibility and paper tracks of reproducibility have been adopted to evaluate the reproducibility of publications~\cite{pineau2021improving, ferro2018overview}. To quantitatively evaluate reproducibility in the context of challenges, a series of quantitative metrics have been studied. For example, \cite{ferro2018overview, breuer2020measure} designed metrics to quantify how closely an information retrieval system can be reproduced to its origins.

\subsection{Fairness}
When AI systems help us in areas such as hiring, financial risk assessment, and face identification, systematic unfairness in their decisions might have negative social ramifications (e.g., underprivileged groups might experience systematic disadvantage in hiring decisions ~\cite{bogen2018help}, or be disproportionally impacted in criminal risk profiling~\cite{dressel2018accuracy, howard2018ugly, hao2019ai}).
This not only damages the trust that various stakeholders have in AI, but also hampers the development and application of AI technology for the greater good.
Therefore, it is important that practitioners keep in mind the {fairness} of AI systems to avoid instilling or exacerbating social bias~\cite{mehrabi2021survey, caton2020fairness, drozdowski2020demographic}.
A common objective of fairness in AI systems is to eliminate or mitigate the effects of biases
The mitigation is non-trivial because the biases can take various forms, such as  data bias, model bias, and procedural bias, in the process of developing and applying AI systems~\cite{mehrabi2021survey}.
Bias often manifests in the form of unfair \emph{treatment} of different \emph{groups} of people based on their protected information (e.g., gender, race, and ethnicity). Therefore, group identity (sometimes also called \emph{sensitive variables}) and system response (prediction) are two factors influencing bias.
Some cases also involve objective \emph{ground truths} of a given task that one should consider when evaluating system fairness, e.g., whether a person’s speech is correctly recognized or their face correctly identified.


Fairness can be applicable at multiple granularities of system behavior~\cite{mehrabi2021survey, caton2020fairness, verma2018fairness}. 
At each granularity, we might be concerned with \emph{distributive fairness} or fairness of the outcome, or \emph{procedural fairness} or fairness of the process (we refer the reader to \cite{grgic2018beyond} for a more detailed discussion).
In each case, we are commonly concerned with the aggregated behavior and the bias therein of an AI system, which is referred to as \emph{statistical fairness} or \emph{group fairness}.
In certain applications, it is also helpful to consider \emph{individual fairness} or \emph{counterfactual fairness}, especially when the sensitive variable can be more easily decoupled from the other features that should justifiably determine the system's prediction~\cite{mehrabi2021survey}.
While the former is more widely applicable to various ML tasks, e.g., speech recognition and face identification, the latter can be critical in cases like resume reviewing for candidate screening \cite{bertrand2004emily}.

At the group level, researchers have identified three abstract principles to categorize different types of fairness~\cite{caton2020fairness}.
We illustrate them with a simple example of hiring applicants from a population consisting of 50\% male and 50\% female applicants, where gender is the sensitive variable (examples adapted from~\cite{verma2018fairness, zhong2018tutorial}):

\begin{itemize}
\item \textbf{Independence.} This requires for the system outcome to be statistically independent of sensitive variables.
In our example, this requires that the rate of admission of male and female candidates be equal (known as \textit{demographic parity}~\cite{zemel2013learning}; see also \textit{disparate impact} \cite{feldman2015certifying}).
\item \textbf{Separation.} Independence does not account for a justifiable correlation between the ground truth and the sensitive variable (e.g., fewer female candidates might be able to lift 100-lb goods more easily than male candidates).
Separation therefore requires that the independence principle hold, conditioned on the underlying ground truth.
That is, if the job requires strength qualifications, the rate of admission for qualified male and female candidates should be equal (known as \textit{equal opportunity}~\cite{hardt2016equality}; see also \textit{equal odds}~\cite{berk2021fairness} and \textit{accuracy equity}~\cite{dieterich2016compas}).
\item \textbf{Sufficiency.} Sufficiency similarly considers the ground truth but requires that the true outcome and the sensitive variable be independent when conditioned on the same system prediction.
That is, given the same hiring decision predicted by the model, we want the same ratio of qualified candidates among male and female candidates (known as \textit{test fairness}~\cite{chouldechova2017fair, hardt2016equality}).
This is closely related to model calibration \cite{pleiss2017fairness}.
\end{itemize}

Note that these principles are mutually exclusive under certain circumstances (e.g., independence and separation cannot both hold when the sensitive variable is correlated with the ground truth). \cite{kleinberg2016inherent} has discussed the trade-off between various fairness metrics.
Furthermore, \cite{corbett2018measure} advocated an extended view of these principles, where the utility of the predicted and true outcomes is factored into consideration (e.g., the risk and cost of recidivism in violent crimes compared with the cost of detention), and can be correlated with the sensitive variables. 
We refer the reader to this work for a more detailed discussion.






\paragraph{Evaluation}

Despite the simplicity of the abstract criteria outlined in the previous section, fairness can manifest in many different forms following these principles (see \cite{wilson2021building, caton2020fairness} for comprehensive surveys, and \cite{madaio2020co} for AI ethics’ checklists).
%
We categorize metrics of fairness according to the properties of models and tasks to help the reader choose appropriate ones for their application:

    \textbf{Discrete vs. continuous variables.} The task output, model prediction, and sensitive variables can all be discrete (e.g., classification and nationality), ranked (e.g., search engines, recommendation systems), or continuous (e.g., regression, classifier scores, age, etc.) in nature.
    An empirical correlation of discrete variables can be evaluated with standard statistical tools, such as correlation coefficients (Pearson/Kendall/Spearman) and ANOVA), while continuous variables often further require binning, quantization, or loss functions to evaluate fairness~\cite{caton2020fairness}.
    
    \textbf{Loss function.} The criteria of fairness often cannot be exactly satisfied given the limitations of empirical data (e.g., demographic parity between groups when hiring only three candidates). Loss functions are useful in this case to gauge how far we are from empirical fairness.
    The choice of loss function can be informed by the nature of the variables of concern: If the variables represent probabilities, likelihood ratios are more meaningful (e.g., {disparate impact}~\cite{feldman2015certifying}); for real-valued regression, the difference between mean distances to the true value aggregated over each group might be used instead to indicate whether we model one group significantly better than another~\cite{calders2013controlling}.
    
    \textbf{Multiple sensitive variables.}
    In many applications, the desirable AI system should be fair to more than one sensitive variables (e.g., the prediction of risk posed by a loan should be fair in terms of both gender and ethnicity; \textit{inter alia}, a recommendation system should ideally be fair to both users and recommendees).
    One can either form a trade-off of ``marginal fairness'' between these variables when they are considered one at a time, i.e., evaluate the fairness of each variable separately and combine the loss functions for final evaluation, or explore the full Cartesian product~\cite{sixta2020fairface} of all variables to achieve joint fairness, which typically requires more empirical observations but tends to satisfy stronger ethical requirements.

\subsection{Privacy protection}
\label{ssect:privacy}
Privacy protection mainly refers to protecting against unauthorized use of the data that can directly or indirectly identify a person or household. These data cover a wide range of information, including name, age, gender, face image, fingerprints, etc. Commitment to privacy protection is regarded as an important factor determining the trustworthiness of an AI system.
The recently released AI ethics guidelines also highlight privacy as one of the key concerns~\cite{jobin2019global, aihleg2018ethics}. 
Government agencies are formulating a growing number of policies to regulate the privacy of data. The General Data Protection Regulation (GDPR) is a representative legal framework, which pushes enterprises to take effective measures for user privacy protection.

In addition to internal privacy protection within an enterprise, recent developments in data exchange across AI stakeholders has yielded new scenarios and challenges for privacy protection.
For example, when training a medical AI model, each healthcare institution typically only has data from the local residents, which might be insufficient. This leads to the demand to collaborate with other institutions and jointly train a model~\cite{sheller2020federated} without leaking private information across institutions.

Existing protection techniques penetrate the entire lifecycle of AI systems to address rising concerns about privacy. In Section~\ref{ssec:privacy-computing} we briefly review techniques for protecting the privacy of data used in data collection and processing (data anonymization and differential privacy in Section~\ref{sect:PrivacyDataProcessing}), model training (secure multi-party computing, and federated learning, in Section~\ref{sec:flandsmpc}), and model deployment (hardware security in Section~\ref{ssect:hardwaresecurity}).  
The realization of privacy protection is also closely related to other aspects of trustworthy AI. For example, the transparency principle is widely used in AI systems. It informs users of personal data collection and enables privacy settings. In the development of privacy-preserving ML software, such as federated learning (e.g., FATE and PySyft), open source is a common practice to increase transparency and certify the protectiveness of the system.

\paragraph{Evaluation}
Laws for data privacy protection like the GDPR require data protection impact assessment (DPIA) if any data processing poses a risk to data privacy. Measures have to be taken to address the risk-related concerns and demonstrate compliance with the law~\cite{aihleg2018assessment}. Data privacy protection professionals and other stakeholders need to be involved in the assessment to evaluate it. 


Previous research has devised various mathematical methods to formally verify the protectiveness of privacy-preserving approaches. Typical verification can be conducted under assumptions such as semi-honest security, which implies all participating parties follow a protocol to perform the computational task but may try to infer the data of other parties from the intermediate results of computation (e.g., \cite{lindell2009proof}).
A stricter assumption is the malicious attack assumption, where each participating party need not follow the given protocol, and can take any possible measure to infer the data~\cite{lindell2020secure}.

In practical scenarios, the empirical evaluation of the risk of leakage of privacy is usually considered~\cite{xia2021enabling, rocher2019estimating}. For example, \cite{rocher2019estimating} showed that 15 demographic attributes were sufficient to render 99\% of the participants unique. An assessment of such data re-identification intuitively reflects protectiveness when designing a data collection plan.

\subsection{Accountability: A Holistic Evaluation throughout the Above Requirements}\label{sec:def-accountability}

We have described a series of requirements to build trustworthy AI. Accountability addresses the regulation on AI systems to follow these requirements. With gradually improving legal and institutional norms on AI governance, accountability becomes a crucial factor for AI to sustainably benefit society with trustworthiness~\cite{doshi2018accountability}.

Accountability runs through the entire lifecycle of an AI system, and requires that the stakeholders of an AI system be obligated to justify their design, implementation, and operation as aligned with human values. At the level of executive, this justification is realized by considerate product design, reliable technique architecture, a responsible assessment of the potential impacts, and the disclosure of information on these aspects~\cite{leslie2019understanding}. Note that in terms of information disclosure, transparency contributes the basic mechanism used to facilitate the accountability of an AI system~\cite{doshi2018accountability, diakopoulos2016accountability}. 

From accountability is also derived the concept of auditability, which requires the justification of a system to be reviewed, assessed, and audited~\cite{leslie2019understanding}. Algorithmic auditing is a recognized approach to ensure the accountability of an AI system and assess its impact on multiple dimensions of human values~\cite{raji2020closing}. See also Section~\ref{sec:auditing}. 



\paragraph{Evaluation}
Checklist-based assessments have been studied to qualitatively evaluate accountability and auditability~\cite{aihleg2018assessment, sai2020audting}. As mentioned in this section, we consider accountability to be the comprehensive justification of each concrete requirement of trustworthy AI. Its realization is composed of evaluations of these requirements over the lifecycle of an AI system~\cite{raji2020closing}. Hence, the evaluation of accountability is reflected by the extent to which these requirements of trustworthiness and their impact can be evaluated.





\begin{figure}[t]
    \centering
    \includegraphics[width=0.9\linewidth]{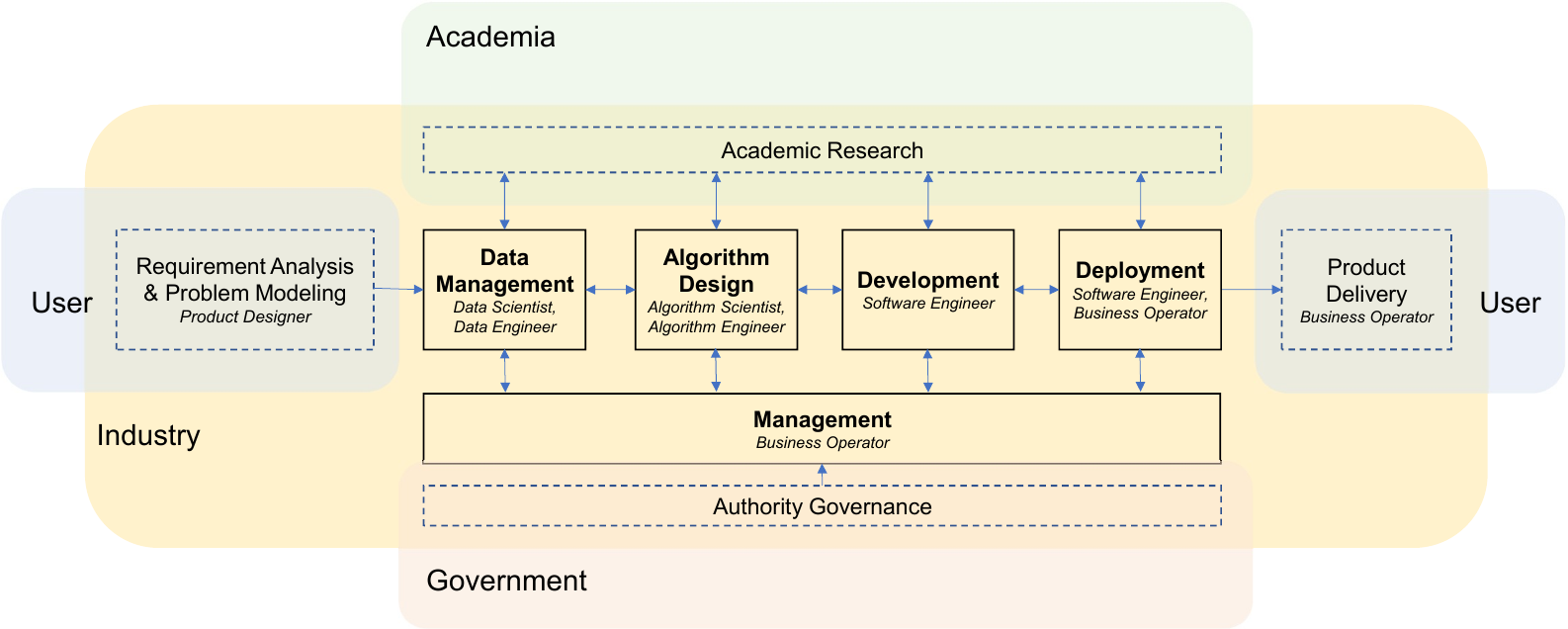}
    \caption{The AI industry holds a connecting position to organize multi-disciplinary practitioners, including users, academia, and government in the establishment of trustworthy AI. In Section~\ref{sec:systematic-approach}, we discuss the current approaches towards trustworthy AI in the five main stages of the lifecycle of an AI system, i.e., data preparation, algorithm design, development, deployment, and management. }
    \label{fig:lifecycle}
\end{figure}


\section{Trustworthy AI: A Systematic Approach} \label{sec:systematic-approach}
We have introduced the concepts relevant to trustworthy AI in Section~\ref{sec:definition-evaluation}. In the last decade, diverse AI stakeholders have made efforts to enhance AI trustworthiness. In Section~\ref{sec:practices} in our appendix, we briefly review their recent practices in multi-disciplinary areas, including research, engineering, and regulation, and studies that are exemplars of industrial applications, including on face recognition, autonomous driving, and NLP. These practices have made important progress in improving AI trustworthiness. However, we find that this work remains insufficient from the industrial perspective. As depicted in Section~\ref{sec:introduction} and Figure~\ref{fig:lifecycle}, the AI industry holds a position to connect multi-disciplinary fields for the establishment of trustworthy AI. This position requires that industrial stakeholders learn and organize these multi-disciplinary approaches and ensure trustworthiness throughout the lifecycle of AI.  

\begin{figure}
    \centering
    \includegraphics[width=1.35\linewidth, angle=-90]{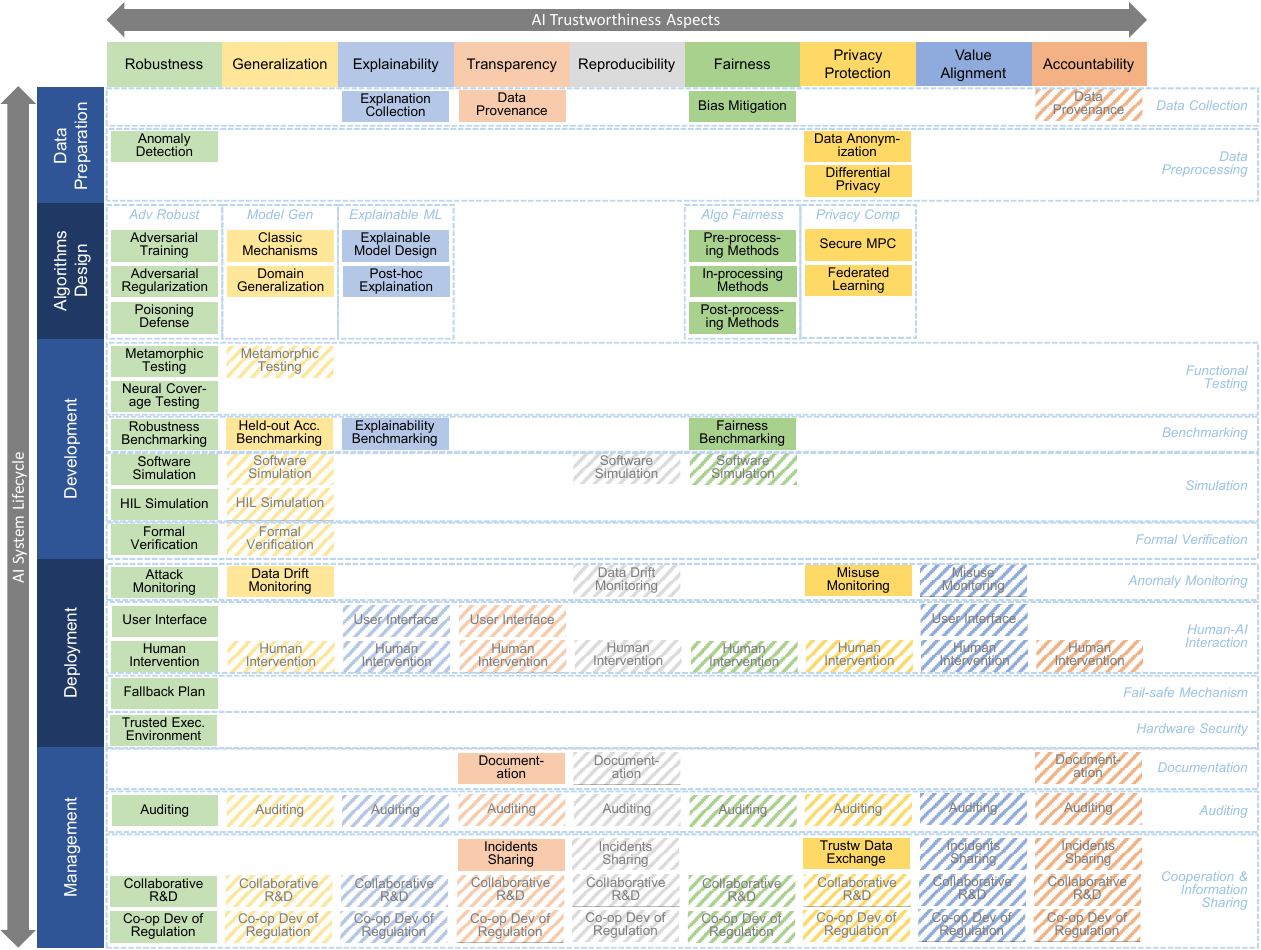}
    \caption{A look-up table that organizes surveyed approaches to AI trustworthiness from different perspectives and in different stages of the lifecycle of the AI system. Some approaches can be used to improve AI trustworthiness from multiple aspects, and are in multiple columns. We dim these duplicated blocks here through stride-filling for better visualization. See the corresponding paragraphs in Section~\ref{sec:systematic-approach} for the details of the approaches.}
    \label{fig:lut}
\end{figure}

In this section, we provide a brief survey of techniques used for building trustworthy AI products and organize them across the lifecycle of product development from an industrial perspective.
As shown by the solid-lined boxes in Figure~\ref{fig:lifecycle}, the lifecycle of the development of a typical AI product can be partitioned into data preparation, algorithm design, development--deployment, and management~\cite{ashmore2021assuring}. We review several critical algorithms, guidelines, and government regulations that are closely relevant to the trustworthiness of AI products in each stage of their lifecycle, with the aim of providing a systematic approach and an easy-to-follow guide to practitioners from varying backgrounds to establish trustworthy AI. The approaches and literature mentioned in this section are summarized in Figure~\ref{fig:lut} and Table~\ref{tab:papers}.

\subsection{Data Preparation}\label{sec:data-management}
Current AI technology is largely data driven. The appropriate management and exploitation of data not only improves the performance of an AI system but also affects its trustworthiness. In this section,
we consider two major aspects of data preparation, i.e., data collection and data preprocessing. We also discuss the corresponding requirements of trustworthy AI.

\subsubsection{Data Collection}
Data collection is a fundamental stage of the lifecycle of AI systems.
An elaborately designed data collection strategy can conduce to the enhancement of AI trustworthiness, such as in terms of fairness and explainability.


\textbf{Bias mitigation.} Training and evaluation data are recognized as a common source of bias for AI systems. 
Many types of bias might exist and plague fairness in data collection, requiring different processes and techniques to combat it (see \cite{mehrabi2021survey} for a comprehensive survey).
Bias mitigation techniques during data collection can be divided into two broad categories: debias sampling and debias annotation.
The former concerns the identification of data points to use or annotate, while the latter focuses on choosing the appropriate annotators.

When sampling data points to annotate, we note that a dataset reflecting the user population does not guarantee fairness because statistical methods and metrics might favor majority groups.
This bias can be further amplified if the majority group is more homogeneous for the task (e.g., recognizing speech in less-spoken accents can be naturally harder due to data scarcity~\cite{koenecke2020racial}).
System developers should therefore take task difficulty into consideration when developing and evaluating fair AI systems. On the other hand, choosing the appropriate annotators is especially crucial for underrepresented data (e.g., when annotating speech recognition data, most humans are also poor at recognizing rarely heard accents).
Therefore, care must be taken in selecting the correct experts, especially when annotating data from underrepresented groups, to prevent human bias from creeping into the annotated data.


\textbf{Explanation collection.} Aside from model design and development, data collection is also integral to building explainable AI systems.
As will be mentioned in Section~\ref{sec:model-explainability}, adding an explanation task to the AI model can help explain the intermediate features of the model. This strategy is used in tasks like NLP-based reading comprehension by generating supporting sentences~\cite{tu2020select, yang2018hotpotqa}. To train the explanation task, it is helpful to consider collecting explanations or information that may not directly be part of the end task, either directly from annotators \cite{wiegreffe2021teach} or with the help of automated approaches \cite{kim2016examples}.

\textbf{Data Provenance.}
Data provenance requires recording the data lineage, including the source, dependencies, contexts, and steps of processing ~\cite{singh2018decision}. By tracking the data lineage at the highest resolution, data provenance can enhance the transparency, reproducibility, and accountability of an AI system~\cite{herschel2017survey, janssen2020data}. Moreover, recent research has shown that data provenance can be used to mitigate data poisoning~\cite{baracaldo2017mitigating}, thus enhancing the robustness and security of an AI system. The technical realization of data provenance has been provided in~\cite{herschel2017survey}. Tool chains~\cite{schelter2017automatically} and documentation~\cite{gebru2021datasheets} guides have also been studied for specific scenarios involving AI systems. To ensure that the provenance is tamper proof, the blockchain has been recently considered as a promising tool to certify data provenance in AI~\cite{dillenberger2019blockchain, alshamsi2021artificial}.

\begin{table}[]
\begin{tabular}{l|l|l}
\hline
Lifecycle                          & Approaches                              & Literature               \\ \hline
\multirow{5}{*}{\cliprow{Data\\Preparation}} & \multirow{2}{*}{Data Collection}        & Bias Mitigation~\cite{mehrabi2021survey, caton2020fairness, verma2018fairness}   \\ \cline{3-3} 
                                    &                                         & Data Provenance~\cite{herschel2017survey, janssen2020data}         \\ \cline{2-3} 
    & \multirow{3}{*}{\cliprow{Data\\Preprocessing}}  & Anomaly Detection \cite{tae2019data, chu2016data, chalapathy2019deep, pang2021deep}\\ \cline{3-3} 
                                    &                                         & Data Anonymozation ~\cite{al2015literature,majeed2020anonymization, majeed2020anonymization,beigi2020survey}      \\ \cline{3-3} 
                                    &                                         & Differential Privacy ~\cite{dwork2008differential,dwork2014algorithmic,ji2014differential}    \\ \hline
\multirow{8}{*}{\cliprow{Algorithm\\Design}} & \multirow{2}{*}{\cliprow{Adversarial\\Robustness}} & Adversarial Robustness~\cite{silva2020opportunities, akhtar2018threat, chakraborty2018adversarial, amodei2016concrete, yuan2019adversarial, bhambri2019survey, wang2019convergence} \\ \cline{3-3} 
                                    &                                         & Poisoning defense~\cite{li2020backdoor}        \\ \cline{2-3} 
                                    & Explainability ML                       & Explainability ML~\cite{arrieta2020explainable, guidotti2018survey, bodria2021benchmarking, dovsilovic2018explainable, molnar2020interpretable, dovsilovic2018explainable}        \\ \cline{2-3} 
                                    & \multirow{2}{*}{\cliprow{Model\\Generalization}} & Model Generalization ~\cite{friedman2001elements,goodfellow2016machine,zhang2017understanding,arpit2017closer,krueger2017deep, kawaguchi2017generalization, belkin2018reconciling}  \\ \cline{3-3} 
                                    &                                         & Domain Generalization~\cite{zhou2021domain, wang2021generalizing}    \\ \cline{2-3} 
                                    & Algorithmic Fairness                    &     Fairness and bias mitigation~\cite{mehrabi2021survey, caton2020fairness, verma2018fairness}                     \\ \cline{2-3} 
                                    & \multirow{2}{*}{Privacy   Computing}    & SMPC  ~\cite{zhao2019secure,evans2018pragmatic}           \\ \cline{3-3} 
                                    &                                         & Federated Learning ~\cite{yang2019federated,kairouz2021advances}\\ \hline
\multirow{4}{*}{Development}        & Functional Testing                      & \cite{huang2020survey, zhang2020machine, marijan2020software}                         \\ \cline{2-3} 
                                    & \cliprow{Performance\\Benchmarking}                &                  \cite{croce2021robustbench, hendrycks2019benchmarking, brun2018software,galhotra2017fairness,tramer2017fairtest,bellamy2018ai, friedler2019comparative, sixta2020fairface,mathew2021hatexplain,deyoung2020eraser,mohseni2018human}        \\ \cline{2-3} 
                                    & Simulation                              & \cite{koenig2004design,todorov2012mujoco,dreossi2019verifai,wymann2000torcs,benekohal1988carsim,tideman2010scenario,lee2019convlab,bokc2007validation, brogle2019hardware}                         \\ \cline{2-3} 
                                    & Formal Verification        & \cite{urban2021review,huang2020survey,seshia2016towards}                          \\ \hline
\multirow{4}{*}{Deployment}         & Anomaly Monitoring & \cite{chalapathy2019deep, pang2021deep, vzliobaite2016overview} \\ \cline{2-3} 
                                    & Human-AI Interaction & \cite{shneiderman2020bridging, chatzimparmpas2020state, kafle2020artificial, torres2018accessibility} \\ \cline{2-3} 
                                    & Fail-Safe Mechanism & \cite{fu2014monocular, muller2016increased, horwick2010strategy, magdici2016fail} \\ \cline{2-3} 
                                    & Hardware Security & \cite{xu2021security, sabt2015trusted, pinto2019demystifying} \\ \hline
\multirow{3}{*}{Management}         & Documentation & \cite{aimar1998introduction, mitchell2019model, gebru2021datasheets,bender2018data,holland2018dataset, arnold2019factsheets} \\ \cline{2-3} 
                                    & Auditing &  \cite{raji2020closing, buolamwini2018gender, madaio2020co, schelenz2020applying, reisman2018algorithmic, sandvig2014auditing, wilson2021building} \\ \cline{2-3} 
                                    & Cooperation & \cite{askell2019role, dafoe2020open} \\ \hline
Workflow                            & MLOps & \cite{makinen2021needs, hummer2019modelops, siddique2020safetyops} \\ \hline
\end{tabular}
\caption{Representative papers of approaches or research directions mentioned in Section~\ref{sec:systematic-approach}. For research directions that are widely studied, we provide the corresponding surveys for readers to refer to. For approaches or research directions without surveys available, we provide representative technical papers. }
\label{tab:papers}
\end{table}

\subsubsection{Data Preprocessing}\label{sect:PrivacyDataProcessing}
Before feeding data into an AI model, data preprocessing helps remove inconsistent pollution of the data that might harm model behavior and sensitive information that might compromise user privacy.


\textbf{Anomaly Detection.} Anomaly detection is also known as \textit{outlier detection}, and has long been an active area of research in ML~\cite{tae2019data, chu2016data, chalapathy2019deep, pang2021deep}. Due to the sensitivity of ML models to outlier data, data cleaning by anomaly detection serves as an effective approach to enhance performance. In recent studies, anomaly detection has been shown to be useful in addressing some requirements of AI trustworthiness. For example, fraudulent data can challenge the robustness and security of systems in areas such as banking and insurance. Various approaches have been proposed to address this issue by using anomaly detection~\cite{chalapathy2019deep}. The detection and mitigation of adversarial inputs is also considered to be a means to defend against evasion attacks and data poisoning attacks~\cite{silva2020opportunities, li2020backdoor, akhtar2018threat}.
It is noteworthy that the effectiveness of detection in high dimensions (e.g., images) is still limited~\cite{carlini2017adversarial}.
The mitigation of adversarial attacks is also referred to as \textit{data sanitization}~\cite{chan2018data, paudice2018label, cretu2008casting}.

\textbf{Data Anonymization (DA).} DA modifies the data so that the protected private information cannot be recovered. Different principles of quantitative data anonymization have been developed, such as $k$-anonymity~\cite{samarati1998protecting}, $(c,k)$-safety~\cite{martin2007worst}, and $\delta$-presence~\cite{nergiz2007hiding}. Data format-specific DA approaches have been studied for decades~\cite{yuan2011protecting, zhang2013scalable,jakob2020design}. For example, private information in the form of graph data for social networks can be potentially contained in properties of the vertices of the graph, its link relationships, weights, or other graph metrics~\cite{zhou2008brief}. Ways of anonymizing such data have been considered in the literature~\cite{liu2008towards,beigi2020survey}. Specific DA methods have also been designed for relational data~\cite{pervaiz2013accuracy}, set-valued data~\cite{terrovitis2008privacy, he2009anonymization}, and image data~\cite{maximov2020ciagan,dirik2014analysis}. Guidelines and standards have been formulated for data anonymization, such as the US HIPAA and the UK ISB1523. Data pseudonymization~\cite{mourby2018pseudonymised} is also a relevant technique promoted by the GDPR. It replaces private information with non-identifying references.
    
    
Desirable data anonymization is expected to be immune from data de-anonymization or re-identification attacks that try to recover private information from anonymized data~\cite{el2011systematic,ji2014structural}. For example, \cite{ji2016graph} introduces several approaches into de-anonymize user information from graph data. To mitigate the risk of privacy leakage, an open-source platform was provided in~\cite{ji2015secgraph} to evaluate the privacy protection-related performance of graph data anonymization algorithms against de-anonymization attacks.
    
    

\textbf{Differential privacy~(DP).}
DP shares information of groups within datasets while withholding individual samples~\cite{dwork2006calibrating,dwork2014algorithmic,dwork2016calibrating}.
Typical DP can be formally defined by $\epsilon$-differential privacy. It measures the extent to which a (randomized) statistical function on the dataset reflects whether an element has been removed~\cite{dwork2006calibrating}.
DP has been explored in various data publishing tasks, such as log data~\cite{hong2014collaborative,zhang2016anonymizing}, set-valued data~\cite{chen2011publishing}, correlated network data~\cite{chen2014correlated}, and crowdsourced data~\cite{wang2016real,ren2018textsf}. 
It has also been applied to single- and multi-machine computing environments, and integrated with ML models for protecting model privacy~\cite{fletcher2019decision,abadi2016deep,wei2020federated}.
Enterprises like Apple have used DP to transform user data into a form from which the true data cannot be reproduced~\cite{apple2017apple}.
In~\cite{erlingsson2014rappor}, researchers proposed the RAPPOR algorithm that satisfies the definition of DP. The algorithm is used for crowdsourcing statistical analyses of user software.  
DP is also used to improve the robustness of AI models against adversarial samples~\cite{lecuyer2019certified}.

\subsection{Algorithm Design} \label{sec:r-d-of-an-ai-model}
A number of aspects of trustworthy AI have been addressed as algorithmic problems in the context of AI research, and have attracted widespread interest.
We organize recent technical approaches by their corresponding aspects of AI trustworthiness, including robustness, explainability, fairness, generalization, and privacy protection, to provide a quick reference for practitioners.

\subsubsection{Adversarial Robustness}\label{sec:adversarial-robustness}
The robustness of an AI model is significantly affected by the training data and the algorithms used. 
We describe several representative directions in this section. Comprehensive surveys can be found in the literature such as~\cite{silva2020opportunities, akhtar2018threat, chakraborty2018adversarial, amodei2016concrete, li2020backdoor, yuan2019adversarial, bhambri2019survey}.

\paragraph{Adversarial training}
Since the discovery of adversarial attacks, it has been recognized that augmenting the training data with adversarial samples provides an intuitive approach for defense against them. This is typically referred to as adversarial training~\cite{goodfellow2014explaining, wang2019convergence, li2016general}. The augmentation can be carried out in a brute-force manner by feeding both the original data and adversarial samples during training~\cite{kurakin2018adversarial}, or by using a regularization term to implicitly represent adversarial samples~\cite{goodfellow2014explaining}. Conventional adversarial training augments data with respect to specific attacks. It can defend against the corresponding attack but is vulnerable to other kinds of attacks. Various improvements have been studied to improve this defense~\cite{bhambri2019survey, madry2018towards, silva2020opportunities}. \cite{tramer2018ensemble} augmented the training data with adversarial perturbations transferred from other models. It is shown to further provide defense against black-box attacks which do not require knowledge of the model parameters 
This can help defend against black-box attacks, which do not require knowledge of the model parameters. \cite{maini2020adversarial} combined multiple types of perturbations into adversarial training to improve the robustness of the model against multiple types of attacks. 

\paragraph{Adversarial regularization}
In addition to the regularization term that implicitly represents adversarial samples, recent research has further explored network structures or regularization to overcome the vulnerabilities of the DNN to adversarial attacks. An intuitive motivation for this regularization is to prevent the outcome of the network from changing dramatically in case of small perturbations. 
For example, \cite{gu2015towards} penalized large partial derivatives of each layer to improve the stability of its output. A similar regularization on gradients was adopted by~\cite{ross2018improving}. Parseval networks~\cite{cisse2017parseval} train the network by imposing a regularization on the Lipschitz constant in each layer.

\paragraph{Certified robustness}

Adversarial training and regularization improve the robustness of AI models in practice but cannot theoretically guarantee that the models work reliably. This problem has prompted research to formally verify the robustness of models (a.k.a. \textit{certified robustness}). Recent research on certified robustness has focused on robust training to deal with perturbations. 
For example, CNN-Cert \cite{boopathy2019cnn}, CROWN~\cite{zhang2018efficient}, Fast-lin, and Fast-lip \cite{weng2018evaluating} aim to minimize an upper bound of the worst-case loss under given input perturbations.  \cite{hein2017formal} instead derives a lower bound on the norm of the input manipulation required to change the decision of the classifier, and uses it as a regularization term for robust training. To address the issue that the exact computation of such bounds is intractable for large networks, various relaxations or approximations, such as~\cite{zhang2019towards, weng2018evaluating}, have been proposed as alternatives to regularization. Note that the above research mainly optimizes robustness only locally near the given training data. To achieve certified robustness on unseen inputs as well, \textit{global robustness} has recently attracted the interest of the AI community~\cite{chen2021learning, leino2021globally}. 

It is also worth noting the recent trend of the intersection of certified robustness and the perspective of formal verification, which aims at developing rigorous mathematical specification and techniques of verification for assurances of software correctness~\cite{clarke1996formal}. 
A recent survey by~\cite{urban2021review} provided a thorough review of the formal verification of neural networks. 



\paragraph{Poisoning defense} Typical poisoning or backdoor attacks contaminate the training data to mislead model behavior. Besides avoiding suspicious data in the data sanitization stage, defensive algorithms against poisoning data are an active area of recent ML research~\cite{li2020backdoor}. The defense has been studied at different stages of a DNN model. For example, based on the observation that backdoor-related neurons are usually inactivated for benign samples, \cite{liu2018fine} proposed pruning these neurons from a network to remove the hidden backdoor. 
Neural Cleanse~\cite{wang2019neural} proactively discovers backdoor patterns in a model. The backdoor can then be avoided by the early detection of backdoor patterns from the data or retraining the model to mitigate the backdoor. The detection of backdoor attacks can also be carried out by analyzing the prediction on the model on specifically designed benchmarking inputs~\cite{kolouri2020universal}.

\subsubsection{Model Generalization}\label{sec:model-generalization}
Techniques of model generalization not only aim to improve model performance, but also explore training AI models with limited data and at limited cost. We review representative approaches to model generalization, categorized as classic generalization and domain generalization.

\paragraph{Classic generalization mechanisms}
As a fundamental principle of model generalization theory, the bias--variance trade-off indicates that a generalized model should maintain a balance between underfitting and overfitting \cite{belkin2019reconciling, friedman2001elements}. For an overfitted model, reducing complexity/capacity might lead to better generalization. Consider the neural network as an example. Adding a bottleneck layer, which has fewer neurons than the layers both below and above, to it can help reduce model complexity and reduce overfitting.

Other than adjusting the architecture of the model, one can mitigate overfitting to obtain better generalization via various explicit or implicit regularizations, such as early stopping \cite{yao2007early}, batch normalization~\cite{ioffe2015batch}, dropout~\cite{srivastava2014dropout}, data augmentation, and weight decay \cite{krogh1992simple}. These regularizations are standard techniques to improve model generalization when the training data are much smaller in size than the number of model parameters \cite{vapnik2013nature}. They aim to push learning to a sub-space of a hypothesis with manageable complexity and reduce model complexity \cite{zhang2017understanding}. However, \cite{zhang2017understanding} also observed that explicit regularization may improve generalization performance but is insufficient to reduce generalization error. The generalization of the deep neural network is thus still an open problem.

\paragraph{Domain generalization}
A challenge for modern deep neural networks is their generalization of out-of-distribution data. This challenge arises from various practical AI tasks~\cite{zhou2021domain, wang2021generalizing} in the area of transfer learning~\cite{pan2009survey, weiss2016survey}. 
Domain adaptation~\cite{zhou2021domain, wang2021generalizing} aims to find domain-invariant features such that an algorithm can achieve similar performances across domains. As another example, the goal of few-shots learning is to generalize model to new tasks using only a few examples~\cite{wang2020generalizing, cheng2019few, yu2018diverse}.
Meta-learning~\cite{vanschoren2018meta} attempts to learn prior knowledge of generalization from a large number of similar tasks. Feature similarity~\cite{koch2015siamese, snell2017prototypical} has been used as a representative type of knowledge prior in works such as the MAML \cite{finn2017model}, reinforcement learning ~\cite{li2016learning}, and memory-augmented NN~\cite{santoro2016meta}.

Model pre-training is a popular mechanism to leverage knowledge learned from other domains, and has recently achieved growing success in both academia and the industry. For example, in computer vision, an established successful paradigm involves pre-training models on a large-scale dataset, such as ImageNet, and then fine-tuning them on target tasks with fewer training data~\cite{girshick2014rich, zeiler2014visualizing, long2015fully}. This is because pre-trained feature representation can be used to transfer information to the target tasks \cite{zeiler2014visualizing}. Unsupervised pre-training has recently been very successful in language processing (e.g., BERT \cite{devlin2019bert} and GPT \cite{radford2018improving}) and computer vision tasks (e.g., MoCo~\cite{he2020momentum} and SeCo~\cite{yao2021seco}). In addition, self-supervised learning provides a good mechanism to learn a cross-modal feature representation. These include the vision and language models VL-BERT~\cite{su2019vl} and Auto-CapTIONs \cite{pan2020auto}. To explain the effectiveness of unsupervised pre-training, \cite{erhan2010does} conducted a series of experiments to illustrate that it can drive learning to the basins of minima that yield better generalization.

\subsubsection{Explainable ML}\label{sec:model-explainability}
In this section, we review representative approaches for the two aspects of ML explainability mentioned in Section~\ref{sec:def-explainability} and their application to different tasks.

\paragraph{Explainable ML model design}
Although they have been recognized as being disadvantageous in terms of performance, explainable models have been actively researched in recent years, and a variety of fully or partially explainable ML models have been studied to push their performance limit.



\textbf{Self-explainable ML models.} A number of self-explainable models have been studied in ML over the years.  The representative ones include the k-nearest neighbors (KNN), linear/logistic regression, decision trees/rules, and probabilistic graphical models ~\cite{arrieta2020explainable, guidotti2018survey, bodria2021benchmarking, molnar2020interpretable}. Note that the self-explainability of these models is sometimes undermined by their complexity. For example, very complex tree or rule structures might sometimes be considered incomprehensible or unexplainable.

Some learning paradigms other than conventional models are also considered to be explainable, such as causal inference~\cite{kuang2020causal, pearl2009causal} and the knowledge graph~\cite{wang2019explainable}. These methods are also expected to provide valuable inspiration to solve the problem of explainability for ML.

\textbf{Beyond self-explainable ML models.} Compared with black-box models, such as the DNN, conventional self-explainable models have poor performance on complex tasks, such as image classification and text comprehension. To achieve a compromise between explainability and performance, hybrid combinations of self-explainable models and black-box models have been proposed. A typical design involves embedding an explainable bottleneck model into a DNN. For example, previous research has embedded linear models and prototype selection to the DNN~\cite{alvarez2018towards, chen2019looks, angelov2020towards}.
In the well-known class activation mapping (CAM)~\cite{zhou2016learning}, an average pooling layer at the end of a DNN can also be regarded as an explainable linear bottleneck. Attention mechanisms~\cite{bahdanau2014neural, xu2015show} have also attracted recent interest, and have been regarded as an explainable bottleneck in a DNN in some studies~\cite{choi2016retain, martins2016softmax}. However, this claim continues to be debated because attention weights representing different explanations can produce similar final predictions~\cite{jain2019attention, wiegreffe2019attention}.

\paragraph{Post-hoc model explanation} 
In addition to designing self-explainable models, understanding how a specific decision is made by black-box models is also an important problem.
A major part of research on this problem has addressed the methodology of post-hoc model explanation and proposed various approaches.

\textbf{Explainer approximation} aims to mimic the behavior of a given model with explainable models. This is also referred to as the global explanation of a model. Various approaches have been proposed to approximate ML models, such as random forests~\cite{zhou2016interpreting, tan2020tree} and neural networks~\cite{craven1995extracting, augasta2012reverse, zhou2003extracting}. With the rise of deep learning in the past decade, explainer approximation on the DNN has advanced as the knowledge distillation problem on explainers such as trees~\cite{zhang2019interpreting,frosst2017distilling}.

\textbf{Feature importance} has been a continually active area of research on explainability. A representative aspect uses local linear approximation to model the contribution of each feature to the prediction. LIME~\cite{ribeiro2016should} and SHAP~\cite{lundberg2017unified} are influential approaches that can be used for predictions on tabular data, computer vision, and NLP. Gradients can reflect how features contribute to the predicted outcome, and have drawn great interest in work on the explainability of the DNN~\cite{simonyan2014deep, selvaraju2017grad}. In NLP or CV, gradients or their variants are used to back-trace the decision of the model to the location of the most intimately related input, in the form of saliency maps and sentence highlights~\cite{sundararajan2017axiomatic, shrikumar2017learning, zeiler2014visualizing, molnar2020interpretable}.

\textbf{Feature introspection} aims to provide a semantic explanation of intermediate features. A representative aspect attaches an extra branch to a model to generate an explanatory outcome that is interpretable by a human. For example, in NLP-based reading comprehension, the generation of a supporting sentence serves as an explanatory task in addition to answer generation~\cite{tu2020select, yang2018hotpotqa}. In image recognition, part-template masks can be used to regularize feature maps to focus on local semantic parts~\cite{zhang2018interpretable}. Concept attribution~\cite{bodria2021benchmarking} is another aspect that maps a given feature space to human-defined concepts. Similar ideas have been used in generative networks to gain control over attributes, such as the gender, age, and ethnicity, in a face generator~\cite{liu2019stgan}.



\textbf{Example-based explanation} explains outcomes of the AI model by using the sample data. For example, an influential function was borrowed from robust statistics in~\cite{koh2017understanding} to find the most influential data instance for a given outcome. Counterfactual explanation~\cite{wachter2017counterfactual, kim2016examples, akula2022cx} works in a contrary way by finding the boundary case to flip the outcome. This helps users better understand the decision surface of the model.





\subsubsection{Algorithmic Fairness} 
Methods to reduce bias in AI models during algorithm development can intervene before the data are fed into the model (pre-processing), when the model is being trained (in-processing), or into model predictions after it has been trained (post-processing).

\paragraph{Pre-processing methods} Aside from debiasing the data collection process, we can debias data before model training. Common approaches include the following:
    
        \textbf{Adjusting sample importance}. This is helpful especially if debiasing the data collection is not sufficient or no longer possible.
        Common approaches include resampling \cite{adler2018auditing}, which involves selecting a subset of the data, reweighting \cite{calders2010three}, which involves assigning different importance values to data examples, and adversarial learning \cite{madry2018towards}, which can be achieved through resampling or reweighting with the help of a trained model to find the offending cases.

        Aside from helping to balance the classification accuracy, these approaches can be applied to balance the cost of classification errors to improve performance on certain groups~\cite{huang2016learning} (e.g., for the screening of highly contagious and severe diseases, false negatives can be costlier than false positives; see \textit{cost-sensitive learning} \cite{thai2010cost}).

        \textbf{Adjusting feature importance}. Inadvertent correlation between features and sensitive variables can lead to unfairness. 
        Common approaches of debiasing include representation transformation \cite{calmon2017optimized}, which can help adjust the relative importance of features, and blinding \cite{chen2018my}, which omits features that are directly related to the sensitive variables.
        
        \textbf{Data augmentation.} 
        Besides making direct use of the existing data samples, it is possible to introduce additional samples that typically involves making changes to the available samples, including through perturbation and relabeling \cite{calders2010three, cowgill2017algorithmic}.
\paragraph{In-processing methods} 
Pre-processing techniques are not guaranteed to have the desired effect during model training because different models can leverage features and examples in different ways. This is where in-processing techniques can be helpful:

        \textbf{Adjusting sample importance.} 
        Similar to pre-processing methods, reweighting \cite{krasanakis2018adaptive} and adversarial learning \cite{celis2019improved} can be used for in-processing, with the potential of either making use of the model parameters or predictions that are not yet fully optimized to more directly debias the model.        
        
        \textbf{Optimization-related techniques.}
        Alternatively, model fairness can be enforced more directly via optimization techniques. For instance, quantitative fairness metrics can be used as regularization \cite{aghaei2019learning} or constraints for the optimization of the model parameters \cite{celis2019classification}.

\paragraph{Post-processing methods}
    Even if all precautions have been taken with regard to data curation and model training, the resulting models might still exhibit unforeseen biases. 
    Post-processing techniques can be applied for debiasing, often with the help of auxiliary models or hyperparameters to adjust the model output. 
    For instance, optimization techniques (e.g., constraint optimization) can be applied to train a smaller model to transform model outputs or calibrate model confidence \cite{kim2018fairness}. 
    Reweighting the predictions of multiple models can also help reduce bias \cite{iosifidis2019fae}.

\subsubsection{Privacy Computing}\label{ssec:privacy-computing}
Apart from privacy-preserving data processing methods, which were introduced in Section~\ref{sect:PrivacyDataProcessing}, another line of methods preserve data privacy during model learning. In this part, we briefly review the two popular categories of such algorithms: secure multi-party computing, and federated learning.

\textbf{Secure Multi-party Computing (SMPC)} \label{sec:mpc-he}
deals with the task whereby multiple data owners compute a function, with the privacy of the data protected and no trusted third party serving as coordinator. A typical SMPC protocol satisfies properties of privacy, correctness, independence of inputs, guaranteed output delivery, and fairness~\cite{zhao2019secure,evans2018pragmatic}. The garbled circuit is a representative paradigm for secure two-party computation~\cite{yao1982protocols,micali1987play}. Oblivious transfer is among the key techniques. It guarantees that the sender does not know what information the receiver obtains from the transferred message. For the multi-party condition, secret sharing is one of the generic frameworks~\cite{karnin1983secret}. Each data instance is treated as a secret and split into several shares. These shares are then distributed to the multiple participating parties. The computation of the function value is decomposed into basic operations that are computed following the given protocol. 
   
The use of SMPC in ML tasks has been studied in the context of both model-specific learning tasks, e.g., linear regression~\cite{gascon2017privacy} and logistic regression~\cite{shi2016secure}, and generic model learning tasks~\cite{mohassel2017secureml}. Secure inference is the an emerging topic that tailors the SMPC for ML use. Its application to ML is as a service in which the server holds the model and clients hold the private data. To reduce the high costs of computation and communication of the SMPC, parameter quantization and function approximation were used together with cryptographic protocols in~\cite{ball2019garbled, agrawal2019quotient}. Several tools have been open-sourced, such as MP2ML~\cite{boemer2020mp2ml}, CryptoSPN~\cite{treiber2020cryptospn}, CrypTFlow~\cite{kumar2020cryptflow,rathee2020cryptflow2}, and CrypTen~\cite{knott2021crypten}.

\label{sec:flandsmpc}
\textbf{Federated learning (FL)} was initially proposed as a secure scheme to collaboratively train an ML model on a data of user interactions with their devices~\cite{mcmahan2017communication}. It quickly gained extensive interest in academia and the industry as a solution to collaborative model training tasks by using data from multiple parties. It aims to address data privacy concerns that hinder ML algorithms from properly using multiple data sources. It has been applied to numerous domains, such as healthcare~\cite{rieke2020future,sheller2020federated} and finance~\cite{long2020federated}. 

Existing FL algorithms can be categorized into horizontal FL, vertical FL, and federated transfer learning algorithms~\cite{yang2019federated}. Horizontal FL refers to the scenario in which each party has different samples but the samples share the same feature space. A training step is decomposed as to first compute optimization updates on each client and then aggregate them on a centralized server without knowing the clients’ private data~\cite{mcmahan2017communication}.
Vertical FL refers to the setting in which all parties share the same sample ID space but have different features. \cite{hardy2017private} used homomorphic encryption for vertical logistic regression-based model learning. In~\cite{gu2020federated}, an efficient method of kernel learning was proposed. Federated transfer learning is applicable to the condition in which none of the parties overlaps in either the sample or the feature space~\cite{liu2020secure}. The connection between FL and other research topics, such as multi-task learning, meta-learning, and fairness learning, has been discussed in~\cite{kairouz2021advances}. To expedite FL-related research and development, many open-source libraries have been released, such as FATE, FedML~\cite{he2020fedml}, and Fedlearn-Algo~\cite{liu2021fedlearn}.

    
    

\subsection{Development} \label{sec:development}

The manufacture of reliable products requires considerable effort in software engineering, and this is sometimes overlooked by AI developers. This lack of diligence, such as insufficient testing and monitoring, may incur long-term costs in the subsequent lifecycle of AI products (a.k.a. \textit{technical debt}~\cite{sculley2015hidden}). Software engineering in the stages of development and deployment has recently aroused wide concern as an essential condition for reliable AI systems~\cite{amershi2019software, lavin2021technology}. Moreover, various techniques researched for this stage can contribute to the trustworthiness of an AI system~\cite{amershi2019software}. In this section, we survey the representative techniques.

\subsubsection{Functional Testing}\label{ssec:testing}
Inherited from the workflow of canonical software engineering, the testing methodology has drawn growing attention in the development of an AI system. In terms of AI trustworthiness, testing serves as an effective approach to certify that the system is fulfilling specific requirements. 
Recent research has explored new approaches to adapt functional testing to AI systems. This has been reviewed in the literature, such as~\cite{huang2020survey, zhang2020machine, marijan2020software}. We describe two aspects of adaption from the literature that are useful to enhance the trustworthiness of an AI system.

\textbf{Test criteria.} Different from canonical software engineering where exact equity is tested between the actual and the expected outputs of a system, an AI system is usually tested by its predictive accuracy on a specific testing dataset. Beyond accuracy, various test criteria have been studied to further reflect and test more complex properties of an AI system. The concept of test coverage in software testing has been transplanted into highly entangled DNN models~\cite{pei2017deepxplore, ma2018deepgauge}. The name of a representative metric---neuron coverage~\cite{pei2017deepxplore}---figuratively illustrates that it measures the coverage of activated neurons in a DNN in analogy to code branches in canonical software testing. Such coverage criteria are effective for certifying the robustness of a DNN against adversarial attacks~\cite{ma2018deepgauge}.

\textbf{Test case generation.} Human-annotated datasets are insufficient for thoroughly testing an AI system, and large-scale automatically generated test cases are widely used. Similar to canonical software testing, the problem to automatically generate the expected ground truth, known as the \textit{oracle problem}~\cite{barr2014oracle}, also occurs in the AI software testing scenario. The hand-crafted test case template is an intuitive but effective approach in applications of NLP~\cite{ribeiro2020beyond}. Metamorphic testing is also a practical approach that converts the input/output pairs into new test cases. For example, \cite{zhang2018deeproad} transfers images of road scenes taken in daylight to rainy images by using GAN as new test cases, and re-uses the original, invariant annotation to test an autonomous driving systems. These testing cases are useful for evaluating the generalization performance of an AI model.
A similar methodology was adopted by adding adversarial patterns to normal images to test adversarial robustness~\cite{ma2018deepgauge}. Simulated environments are also widely used to test applications such as computer vision and reinforcement learning. We further review this topic in Section~\ref{sec:simulation}.



\subsubsection{Performance Benchmarking} \label{sec:benchmark}
Unlike conventional software, the functionality of AI systems is often not easily captured in functional test alone.
To ensure that systems are trustworthy in terms of different aspects of interest, benchmarking (a.k.a. performance testing in software engineering) is often applied to ensure system performance and stability when these characteristics can be automatically measured.

Robustness is an important aspect of trustworthiness that is relatively amenable to automatic evaluation.
\cite{croce2021robustbench} and \cite{hendrycks2019benchmarking} introduced a suite of black-box and white-box attacks to automatically evaluate the robustness of AI systems. It can potentially be performed as a sanity check before such systems are deployed to affect millions of users.
Software fairness has also been a concern since conventional software testing~\cite{brun2018software,galhotra2017fairness}. 
Criteria for AI systems have been studied to spot issues of unfairness by investigating the correlation among sensitive attributes, system outcomes, and the true label when applicable to well-designed diagnostic datasets~\cite{tramer2017fairtest}.  
Well-curated datasets and metrics have been proposed in the literature to evaluate performance on fairness metrics that are of interest for different tasks~\cite{bellamy2018ai, friedler2019comparative, sixta2020fairface}.

More recently, there has been growing interest in benchmarking explainability when models output explanations in NLP applications.
For instance, \cite{mathew2021hatexplain} asks crowd workers to annotate salient pieces of text that lead to their belief that the text is hateful or offensive, and examines how well model-predicted importance fits human annotations.
\cite{deyoung2020eraser} instead introduces partial perturbations to the text for human annotators, and observes if the system’s explanations match perturbations that change human decisions.
In the meantime, \cite{poursabzi2021manipulating} reported that explainability benchmarking remains relatively difficult because visual stimuli are higher dimensional and continuous.


\subsubsection{Development by Simulation}\label{sec:simulation}
While benchmarks serve to evaluate AI systems in terms of predictive behavior given static data, the behavior of many systems is deeply rooted in their interactions with the world.
For example, benchmarking autonomous vehicle systems on static scenarios is insufficient to help us evaluate their performance on dynamic roadways.
For these systems, simulation often plays an important role in ensuring their trustworthiness before deployment.

Robotics is a sub-fields of AI where simulations are most commonly used.
Control systems for robots can be compared and benchmarked in simulated environments such as Gazebo \cite{koenig2004design}, MuJoCo \cite{todorov2012mujoco}, and VerifAI \cite{dreossi2019verifai}.
Similarly, simulators for autonomous driving vehicles have been widely used, including CARLA~\cite{dosovitskiy2017carla}, TORCS \cite{wymann2000torcs}, CarSim \cite{benekohal1988carsim}, and PRESCAN \cite{tideman2010scenario}.
These software platforms simulate the environment in which robots and vehicles operate as well as the actuation of controls on the simulated robots or cars.
In NLP, especially conversational AI, simulators are widely used to simulate user behavior to test system ability and fulfill user needs by engaging in a dialog \cite{lee2019convlab}.
These simulators can help automatically ensure the performance of AI systems in an interactive environment and diagnose issues before their deployment.

Despite the efficiency, flexibility, and replicability afforded by software simulators, they often still fall short of perfectly simulating constraints faced by the AI system when deployed as well as environmental properties or variations in them. 
For AI systems that are deployed on embedded or otherwise boxed hardware, it is important to understand the system behavior when they are run on the hardware used in real-world scenarios.
Hardware-in-the-loop simulations can help developers understand system performance when it is run on chips, sensors, and actuators in a simulated environment, and is particularly helpful for latency- and power-critical systems like autonomous driving systems \cite{bokc2007validation, brogle2019hardware}.
By taking real-world simulations a step further, one can also construct controlled real-world environments for fully integrated AI systems to roam around in (e.g., test tracks for self-driving cars with road signs and dummy obstacles). This provides more realistic measurements and assurances of performance before releasing such systems to users.

\subsection{Deployment} \label{sec:deployment}

After development, AI systems are deployed on realistic products, and interact with the environment and users.
To guarantee that the systems are trustworthy, a number of approaches should be considered at the deployment stage, such as adding additional components to monitor anomalies, and developing specific human--AI interaction mechanisms for transparency and explainability.

\subsubsection{Anomaly Monitoring}
Anomaly monitoring has become a well-established methodology in software engineering. In terms of AI systems, the range of monitoring has been further extended to cover data outliers, data drifts, and model performance. As a keying safeguard for the successful operation of an AI system, monitoring provides the means to enhance the system's trustworthiness in multiple aspects. Some representative examples are discussed below.

\textit{Attack monitoring} has been widely adopted in conventional SaaS, such as fraud detection~\cite{abdallah2016fraud} in e-commerce systems. In terms of the recent emerging adversarial attacks, detection and monitoring~\cite{metzen2017detecting} of such attack inputs is also recognized as an important means to ensure system robustness. \textit{Data drift monitoring}~\cite{rabanser2019failing} provides important means to maintain the generalization of an AI system under concept change~\cite{vzliobaite2016overview} caused by dynamic environment such as market change~\cite{samuylova2020machine}. \textit{Misuse monitoring} is recently also adopted in several cloud AI services~\cite{javadi2020monitoring} to avoid improper use such as unauthorized population surveillance or individual tracking by face recognition, which helps ensure the proper alignment of ethical values.


\subsubsection{Human--AI Interaction}
As an extension of human--computer interaction (HCI), human--AI interaction has aroused wide attention in the AI industry~\cite{abdul2018trends, amershi2019guidelines}. Effective human--AI interaction affects the trustworthiness of an AI system in multiple aspects. We briefly illustrate two topics.

\textbf{User interface} serves as the most intuitive factor affecting user experience. It is a major medium for an AI system to disclose its internal information and decision-making procedure to users, and thus has an important effect on the transparency and explainability of the system~\cite{shneiderman2020bridging, weld2019challenge}. Various approaches of interaction have been studied to enhance the explainability of AI, including the visualization of ML models~\cite{chatzimparmpas2020state} and interactive parameter-tuning~\cite{weld2019challenge}. In addition to transparency and explainability, the accessibility of the interface also significantly affects user experience of trustworthiness. AI-based techniques of interaction have enabled various new forms of human--machine interfaces, such as chatbots, audio speech recognition, and gesture recognition, and might result in accessibility problems for disabled people. Mitigating such unfairness has aroused concerns in recent research~\cite{kafle2020artificial, torres2018accessibility}.

\textbf{Human intervention}, such as by monitoring failure or participating in decisions~\cite{schmidt2017intervention}, has been applied to various AI systems to compensate for limited performance. Advanced Driving Assistance System (ADAS) can be considered a typical example of systems involving human intervention, where the AI does the low-level driving work and the human makes the high-level decision. In addition to compensating for decision-making, human intervention provides informative supervision to train or fine-tune AI systems in many scenarios, such as the shadow mode~\cite{templaton2019tesla} of autonomous driving vehicles. To minimize and make the best use of human effort in such interaction, efficient design of patterns of human--machine cooperation is an emerging topic in inter-disciplinary work on HCI and AI, and is referred to as \textit{human-in-the-loop} or \textit{interactive machine learning}~\cite{holzinger2016interactive} in the literature.


\subsubsection{Fail-Safe Mechanisms}
Considering the imperfection of current AI systems, it is important to avoid harm when the system fails in exceptional cases. By learning from conventional real-time automation systems, the AI community has realized that a fail-safe mechanism or fallback plan should be an essential part of the design of an AI system if its failure can cause harm or loss. This mechanism is also emerging as an important requirement in recent AI guidelines, such as~\cite{aihleg2018ethics}.
The fail-safe design has been observed in multiple areas of robotics in the past few years. In the area of UAV, the fail-safe algorithm has been studied for a long time to avoid frequent collision of quadrocopters~\cite{fu2014monocular} and to ensure safe landing upon system failure~\cite{muller2016increased}. 
In autonomous driving where safety is critical, a fail-safe mechanism like standing still has become an indispensable component in ADAS products~\cite{horwick2010strategy}, and is being researched at a higher level of automation~\cite{magdici2016fail}.


\subsubsection{Hardware Security}
\label{ssect:hardwaresecurity}

AI systems are widely deployed on various hardware platforms to cope with the diverse scenarios, ranging from servers in computing centers to cellphones and embedded systems. Attacks on OS and hardware lead to new risks, such as data tampering or stealing, and threaten the robustness, security, and privacy of AI systems. Various approaches have been studied to address this new threat~\cite{xu2021security}. From the perspective of hardware security, the concept of a trusted execution environment (TEE) is a recent representative technique that has been adopted by many hardware manufacturers~\cite{sabt2015trusted}. The general mechanism of the TEE is to provide a secure area for the data and the code. This area is not interfered with by the standard OS such that the protected program cannot be attacked. ARM processors support TEE implementation using the TrustZone design~\cite{pinto2019demystifying}. They simultaneously run a secure OS and a normal OS on a single core. The secure part provides a safe environment for sensitive information. The Intel Software Guard Extensions (SGX) implement the TEE by hardware-based memory encryption~\cite{mckeen2016intel}. Its enclave mechanism allows for the allocation of protected memory to hold private information. Such security mechanisms have been used to protect sensitive information like biometric ID and financial account passwords, and are applicable to other AI use cases.

\subsection{Management}\label{sec:governance}

AI practitioners such as researchers and developers have studied various techniques to improve AI trustworthiness in the aforementioned stages of the data, algorithm, development, and deployment stages.
Beyond these concrete approaches, appropriate management and governance provide a holistic guarantee that trustworthiness is consistently aligned throughout the lifecycle of a AI system. In this section, we introduce several executable approaches which facilitate the AI community to improve management and governance on AI trustworthiness.

\subsubsection{Documentation}\label{sec:documentation}

Conventional software engineering has accumulated a wealth of experience in leveraging documentation to assist in development. Representative documentation types include requirement documents, product design documents, architecture documents, code documents, and test documents ~\cite{aimar1998introduction}. Beyond conventional software engineering, multiple new types of documents have been proposed to adapt to the ML training and testing mechanisms. The scope of these documents may include the purposes and characteristics of the model~\cite{mitchell2019model}, datasets~\cite{gebru2021datasheets,bender2018data,holland2018dataset}, and services~\cite{arnold2019factsheets}. As mentioned in Sections~\ref{sec:def-transparency} and~\ref{sec:def-accountability}, documentation is an effective and important approach to enhance the system’s transparency and accountability by tracking, guiding, and auditing its entire lifecycle~\cite{raji2020closing}, and serves as a cornerstone of building a trustworthy AI system.






\subsubsection{Auditing}\label{sec:auditing}
With lessons learned from safety-critical industries, e.g., finance and aerospace, auditing has been recently recognized as an effective mechanism to examine whether an AI system complies with specific principles~\cite{buolamwini2018gender, wilson2021building}. 
In terms of the position of auditors, the auditing process can be categorized as internal or external. Internal auditing enables self-assessment and iterative improvement for manufacturers to follow the principles of trustworthiness. It can cover the lifecycle of the system without the leakage of trade secrets~\cite{raji2020closing}. On the other hand, external auditing by independent parties is more effective in gaining public trust~\cite{buolamwini2018gender}.

Auditing might involve the entire or selective parts of the lifecycle of an AI system. A comprehensive framework of internal auditing can be found in \cite{raji2020closing}.
The means of auditing might include interviews, documented artifacts, checklists, code review, testing, and impact assessment. For example, documentation like product requirement documents (PRDs), model cards~\cite{mitchell2019model}, and datasheets~\cite{gebru2021datasheets} serve as important references to understand the principle alignment during development. Checklists are widely used as a straightforward qualitative approach to evaluate fairness~\cite{madaio2020co}, transparency~\cite{schelenz2020applying}, and reproducibility~\cite{pineau2021improving}. Quantitative testing also serves as a powerful approach, and has been successfully executed to audit fairness in, for example, the Gender Shade study~\cite{buolamwini2018gender}. Inspired by the EU’s Data Protection Impact Assessment (DPIA), the concept of Algorithmic Impact Assessment (AIA) has been proposed to evaluate the claims of trustworthiness and discover negative effects~\cite{reisman2018algorithmic}. Besides the above representatives, designs of approaches to algorithmic auditing can be found in~\cite{sandvig2014auditing, wilson2021building}.

\subsubsection{Cooperation and Information Sharing}\label{sec:cooperation}
As shown in Figure~\ref{fig:lifecycle}, the establishment of trustworthy AI requires cooperation between stakeholders. From the perspective of industry, cooperation with academia enables the fast application of new technology to enhance the performance of the product and reduce the risk posed by it. Cooperation with regulators certifies the products as appropriately following the principles of trustworthiness. Moreover, cooperation between industrial enterprises helps address consensus-based problems, such as data exchange, standardization, and ecosystem building~\cite{askell2019role}. Recent practices of AI stakeholders have shown the efficacy of cooperation in various dimensions. We summarize these practices in the following aspects below. 

\textbf{Collaborative research and development.} Collaboration has been a successful driving force in the development of AI technology. To promote research on AI trustworthiness, stakeholders are setting-up various forms of collaboration, such as research workshops on trustworthy AI and cooperative projects like DARPA XAI~\cite{gunning2019darpa}. 

\textbf{Trustworthy data exchange.} The increasing business value of data raises the demand to exchange them across companies in various scenarios (e.g., the medical AI system in Section~\ref{ssect:privacy}). Beyond privacy-based computing techniques, the cooperation between data owners, technology providers, and regulators is making progress in establishing an ecosystem of data exchange, and solving problems such as data pricing and data authorization.

\textbf{Cooperative development of regulation.} Active participation in the development of standards and regulations serves as an important means for academia, the industry, and regulators to align their requirements and situations.

\textbf{Incident sharing.} The AI community has recently recognized incident sharing as an effective approach to highlight and prevent potential risks to AI systems~\cite{brundage2020toward}. The AI Incident Database~\cite{DavidDao2020} provides an inspiring example for stakeholders to share negative AI incidents so that the industry can avoid similar problems.

\subsection{TrustAIOps: A Continuous Workflow toward Trustworthiness} \label{ssec:workflow}



The problem of trustworthy AI arises from the fast development of AI technology and its emerging applications. AI trustworthiness is not a well-studied static bar to reach by some specific solutions. The establishment of trustworthiness is a dynamic procedure. We have witnessed the evolution of different dimensions of trustworthiness over the past decade~\cite{jobin2019global}. For example, research on adversarial attacks has increased concerns regarding adversarial robustness. The applications of safety-critical scenarios have rendered more stringent the requirements of accountability of an AI system. The development of AI research, the evolution of the forms of AI products, and the changing perspectives of society imply the continual reformulation of the requirements of and solutions to trustworthiness. Therefore, we argue that beyond the requirements of an AI product, the AI industry should consider trustworthiness as an ethos of its operational routine and be prepared to continually enhance the trustworthiness of its products.

The constant enhancement of AI trustworthiness positions requirements on a new workflow for the AI industry. Recent studies on industrial AI workflow extend the mechanism of DevOps~\cite{bass2015devops} to MLOps~\cite{makinen2021needs}, to enable improvements in ML products. The concept of DevOps has been adopted in modern software development to continually deploy software features and improve their quality. MLOps~\cite{makinen2021needs} and its variants, such as ModelOps~\cite{hummer2019modelops} and SafetyOps~\cite{siddique2020safetyops}, extend DevOps to cover the ML lifecycle of data preparation, training, validation, and deployment, in its workflow. The workflow of MLOps provides a start point to build the workflow for trustworthy AI. By integrating the ML lifecycle, MLOps connects research, experimentation, and product development to enable the rapid leveraging of the theoretical development of trustworthy AI. A wealth of toolchains of MLOps have been released recently to track AI artifacts such as data, model, and meta-data to increase the accountability and reproducibility of products~\cite{hummer2019modelops}. Recent research has sought to extend MLOps to further integrate trustworthiness into the AI workflow. For example, \cite{siddique2020safetyops} extended MLOps with safety engineering as SafetyOps for autonomous driving.

As we have illustrated in this section, building trustworthiness requires the continual and systematic upgrade of the AI lifecycle. By extending MLOps, we summarize this upgrade of practices as a new workflow, \textit{TrustAIOps}, which focuses on imposing the requirements of trustworthiness over the entire AI lifecycle. This new workflow contains the following properties:

\begin{itemize}

    \item \textbf{Close collaboration between inter-disciplinary roles.} Building trustworthy AI requires organizing different roles, such as ML researchers, software engineers, safety engineers, and legal experts. Close collaboration mitigates the gap in knowledge between forms of expertise (e.g., \cite{lepri2018fair}, c.f., Sections~\ref{sec:cooperation} and \ref{sec:recent-multidisciplinary}).
    \item \textbf{Aligned principles of trustworthiness.} The risk of untrustworthiness exists in every stage in the lifecycle of an AI system. Mitigating such risks requires that all stakeholders in the AI industry be aware of and aligned with unified trustworthy principles (e.g., \cite{shneiderman2020bridging}, c.f., Section~\ref{sec:recent-multidisciplinary}).
    \item \textbf{Extensive management of artifacts.} An industrial AI system is built upon various artifacts such as data, code, models, configuration, product design, and operation manuals. The elaborate management of these artifacts helps assess risk and increases reproducibility and auditability (c.f., Section~\ref{sec:documentation}).
    \item \textbf{Continuous feedback loops.} Classical continuous integration and continuous development (CI/CD) workflows provide effective mechanisms to improve the software through feedback loops. In a trustworthy AI system, these feedback loops should connect and iteratively improve the five stages of its lifecycle, i.e., data, algorithm, development, deployment, and management (e.g., \cite{raji2020closing, stanton2021trust}).
\end{itemize}

The evolution of the industrial workflow of AI is a natural reflection of the dynamic procedure to establish its trustworthiness. By systematically organizing stages of the AI lifecycle and inter-disciplinary practitioners, the AI industry is able to understand the requirements of trustworthiness from various perspectives, including technology, law, and society, and deliver continual improvements.

\section{Conclusion, Challenges and Opportunities}\label{sec:challenges-and-opportunities}


In this survey, we outlined the key aspects of trustworthiness that we think are essential to AI systems.
We introduced how AI systems can be evaluated and assessed on each of these aspects, and reviewed current efforts in this direction in the industry.
We further proposed a systematic approach to consider these aspects of trustworthiness in the entire lifecycle of real-world AI systems, which offers recommendations for every step of the development and use of these systems.
We recognize that fully adopting this systematic approach to build trustworthy AI systems requires that practitioners embrace the concepts underlying the key aspects that we have identified.
More importantly, it requires a shift of focus from \emph{performance-driven} AI to \emph{trust-driven} AI.
In the short run, this shift will inevitably involve side-effects, such as longer learning time, slowed development, and/or increased cost to build AI systems.
However, we encourage practitioners to focus on the long-term benefits of gaining the trust of all stakeholders for the sustained use and development of these systems.
In this section, we conclude by discussing some of the open challenges and potential opportunities in the future development of trustworthy AI.

\subsection{AI Trustworthiness as Long-Term Research}
\label{ssec:long-term-research}

Our understanding of AI trustworthiness is far from complete or universal, and will inevitably evolve as we develop new AI technologies and understand their societal impact more clearly.
This procedure requires long-term research in multiple key areas of AI.
In this section, we discuss several open questions that we think are crucial to address for the future development of AI trustworthiness.

\subsubsection{Immaturity of Approaches to Trustworthiness.}
As mentioned in Section~\ref{sec:definition-evaluation}, several aspects of AI trustworthiness, such as explainability and robustness, address the limitation of current AI technologies. Despite wide interest in AI research, satisfactory solutions are still far from reach.

Consider explainability as an example. Despite being an active field of AI research, it remains poorly understood. Both current explanatory models and post-hoc model explanation techniques share a few common issues, e.g., 1) the explanation is fragile to perturbations~\cite{ghorbani2019interpretation}, 2) the explanation is not always consistent with human interpretation~\cite{bodria2021benchmarking}, and 3) it is difficult to judge if the explanation is correct or faithful~\cite{molnar2020interpretable}. These problems pose important questions in the study of explainability and provide valuable directions of research in theoretical research on AI.

Another example is robustness. The arms race between adversarial attack and defense reflects the immaturity of our understanding of the robustness of AI. As in other areas of security, attacks evolve along with the development of defenses. Conventional adversarial training~\cite{goodfellow2014explaining} has been shown to be easily fooled by subsequently developed attacks~\cite{tramer2018ensemble}. The corresponding defense~\cite{tramer2018ensemble} is later shown to be vulnerable against new attacks~\cite{dong2019evading}. This not only requires that practitioners be agile in adopting defensive techniques to mitigate the risk of new attacks in a process of long-term and continual development, but also poses long-term challenges to theoretical research~\cite{raghunathan2018certified}. 

\subsubsection{Frictional Impact of Trustworthy Aspects.}
As we have shown in Section~\ref{sec:definition-evaluation}, there are a wealth of connections and support between different aspects of trustworthiness. On the other hand, research has shown that there are frictions or trade-offs between these aspects in some cases, which we review here.

Increased transparency improves trust in AI systems through information disclosure. However, disclosing inappropriate information might increase potential risks. For example, excessive transparency on datasets and algorithms might leak private data and commercial intellectual property. Disclosure of detailed algorithmic mechanisms can also lead to the risk of targeted hacking~\cite{akhtar2018threat}. On the other hand, an inappropriate explanation might also cause users to overly rely on the system and follow wrong decisions of AI~\cite{stumpf2016explanations}. Therefore, the extent of transparency of an AI system should be specified carefully and differently for the roles of public users, operators, and auditors.

From an algorithmic perspective, the effects of different objectives of trustworthiness on model performance remain insufficiently understood. Adversarial robustness increases the model’s generalizability and reduces overfitting, but tends to negatively impact its overall accuracy~\cite{tsipras103robustness, zhang2019theoretically}. A similar loss of accuracy occurs in explainable models~\cite{bodria2021benchmarking}. Besides this trust--accuracy trade-off, algorithmic friction exists between the dimensions of trustworthiness. For example, adversarial robustness and fairness can negatively affect each other during training~\cite{xu2021robust, roh2020fr}. Furthermore, studies on fairness and explainability have shown several approaches to explanation to be unfair~\cite{dodge2019explaining}.


These frictional effects suggest that AI trustworthiness cannot be achieved through hillclimbing on a set of disjoint criteria. Compatibility should be carefully considered when integrating multiple requirements into a single system. Recent studies~\cite{zhang2019theoretically, xu2021robust} provide a good reference to start with.


\subsubsection{Limitations in Current Evaluations of Trustworthiness.}
Repeatable and quantitative measurements are the cornerstone of scientific and engineering progress.
However, despite increasing research interest and efforts, the quantification of many aspects of AI trustworthiness remains elusive.
Of the various aspects that we have discussed in this paper, the explainability, transparency, and accountability of AI systems are still seldom evaluated quantitatively, which makes it difficult to accurately compare systems.
Developing good methods of quantitative evaluation for these desiderata, we believe, will be an important first step in research on these aspects of AI trustworthiness as a scientific endeavor, rather than a purely philosophical one.

\subsubsection{Challenges and Opportunities in the Era of Large Scale Pre-trained Models.} 
Large scale pre-trained models have brought dramatic breakthroughs for AI. They not only show the potential to a more general form of AI~\cite{manning2022human}, but also bring new challenges and opportunities to the establishment of trustworthy AI. One of the most important properties of a large scale pre-trained model is the ability to transfer its learned knowledge to new tasks in a manner of few-shot or zero-shot learning~\cite{brown2020language}. This largely fulfills people’s requirement on the generalization of AI, and is recognized to hold great value in commercial applications. On the other hand, the risks of the large scale pre-trained models to be untrustworthy have been revealed by recent studies. For example, the training procedure is known to be costly and difficult to reproduce for third-parties, as mentioned in Section~\ref{sec:generalization}. Most downstream tasks have to directly adapt the pre-trained model without auditing its entire lifecycle. This business model poses a risk that downstream users might be affected by any biases presented in these models~\cite{manning2022human}. Privacy leakage is another issue that is recently revealed. Some pre-trained models are reported to output training text data containing private user information such as addresses~\cite{carlini2021extracting}. 

The development and application of the large scale pre-trained models is accelerating. To guarantee that this advancement will benefit the society without causing new risks, it is worthwhile for both academia and the industry to carefully study its potential impact in the perspective of AI trustworthiness.

\subsection{End-User Awareness of the Importance of AI Trustworthiness}\label{ssec:end-user}
Other than developers and providers of AI systems, end-users are an important yet opt-ignored group of stakeholders. 
%

Besides educating the general public about the basic concepts of AI trustworthiness, developers should consider how it can be presented to users to deliver hands-on experiences of trustworthy AI systems.
One positive step in this direction involves demonstrating system limitations (e.g., Google Translate displays translations in multiple gendered pronouns when the input text is ungendered) or explainable factors used for system prediction (e.g., recommendation systems can share user traits that are used in curating ads, like ``female aged 20--29'' and ``interested in technology'').
Taking this a step further, we believe that it would enable the user to directly control these factors (e.g., user traits) counterfactually, and judge for themselves whether the system is fair, robust, and trustworthy. 

Finally, we recognize that not all aspects of trustworthiness can be equally easily conveyed to end-users.
For instance, the impact of privacy preservation or transparency cannot be easily demonstrated to the end-user when AI systems are deployed.
We believe that media coverage and government regulations regarding these aspects can be very helpful in raising public awareness.

\subsection{Inter-disciplinary and International Cooperation}

An in-depth understanding of AI trustworthiness involves not only the development of better and newer AI technologies,
but also requires us to better understand the interactions between AI and human society.
We believe this calls for collaboration across various disciplines that reach far beyond computer science.
First, AI practitioners should work closely with domain experts whenever AI technologies are deployed to the real world and has impacts on people, e.g., in medicine, finance, transportation, and agriculture.
Second, AI practitioners should seek advice from social scientists to better understand the (often unintended) societal impacts of AI and work together to remedy them, e.g., the impact of AI-automated decisions, job displacement in AI-impacted sectors, and the effect of the use of AI systems in social networks.
Third, AI practitioners should carefully consider how the technology is presented to the public as well as inter-disciplinary collaborators, and make sure to communicate the known limitations of AI systems honestly and clearly.

In the meantime, the development of trustworthy AI is by no means a unique problem of any single country, nor does the potential positive or negative effect of AI systems respect geopolitical borders.
Despite the common portrayal of AI as a race between countries (see Section~3.2 of \cite{hagendorff2020ethics} for a detailed discussion), technological advances in the climate of increased international collaboration are far from a zero-sum game.
It not only allows us to build better technological solutions by combining diverse ideas from different backgrounds, but also helps us better serve the world's population by recognizing our shared humanity as well as our unique differences.
We believe that tight-knit inter-disciplinary and international cooperation will serve as the bedrock for rapid and steady developments in trustworthy AI technology, which will in turn benefit humanity at large.



\section{Acknowledgments}
The authors would like to thank Yanqing Chen, Jing Huang, Shuguang Zhang, and
Liping Zhang for their valuable suggestions. We also thank Yu He, Wenhan Xu, Xinyuan Shan, Chenliang Wang, Peng Liu, Jingling Fu, Baicun Zhou, Hongbao Tian and Qili Wang for their assistance with the experiment in the appendix. 



\printbibliography

@article{liu2021fedlearn,
  title={Fedlearn-Algo: A flexible open-source privacy-preserving machine learning platform},
  author={Liu, Bo and Tan, Chaowei and Wang, Jiazhou and Zeng, Tao and Shan, Huasong and Yao, Houpu and Heng, Huang and Dai, Peng and Bo, Liefeng and Chen, Yanqing},
  journal={arXiv:2107.04129},
  url={https://arxiv.org/abs/2107.04129},
  year={2021}
}

@article{he2020fedml,
  title={Fedml: A research library and benchmark for federated machine learning},
  author={He, Chaoyang and Li, Songze and So, Jinhyun and Zeng, Xiao and Zhang, Mi and Wang, Hongyi and Wang, Xiaoyang and Vepakomma, Praneeth and Singh, Abhishek and Qiu, Hang and others},
  journal={arXiv:2007.13518},
  year={2020}
}

@article{lindell2020secure,
  title={Secure multiparty computation (MPC)},
  author={Lindell, Yehuda},
  journal={Cryptology ePrint Archive},
  year={2020}
}

@article{mourby2018pseudonymised,
  title={Are ‘pseudonymised’data always personal data? Implications of the GDPR for administrative data research in the UK},
  author={Mourby, Miranda and Mackey, Elaine and Elliot, Mark and Gowans, Heather and Wallace, Susan E and Bell, Jessica and Smith, Hannah and Aidinlis, Stergios and Kaye, Jane},
  journal={Computer Law \& Security Review},
  volume={34},
  number={2},
  pages={222--233},
  year={2018},
  publisher={Elsevier}
}

@incollection{mckeen2016intel,
  title={Intel{\textregistered} software guard extensions (intel{\textregistered} sgx) support for dynamic memory management inside an enclave},
  author={McKeen, Frank and Alexandrovich, Ilya and Anati, Ittai and Caspi, Dror and Johnson, Simon and Leslie-Hurd, Rebekah and Rozas, Carlos},
  booktitle={Proceedings of the Hardware and Architectural Support for Security and Privacy 2016},
  pages={1--9},
  year={2016}
}

@article{pinto2019demystifying,
  title={Demystifying arm trustzone: A comprehensive survey},
  author={Pinto, Sandro and Santos, Nuno},
  journal={ACM Computing Surveys (CSUR)},
  volume={51},
  number={6},
  pages={1--36},
  year={2019},
  publisher={ACM New York, NY, USA}
}

@inproceedings{sabt2015trusted,
  title={Trusted execution environment: what it is, and what it is not},
  author={Sabt, Mohamed and Achemlal, Mohammed and Bouabdallah, Abdelmadjid},
  booktitle={2015 IEEE Trustcom/BigDataSE/ISPA},
  volume={1},
  pages={57--64},
  year={2015},
  organization={IEEE}
}

@article{lindell2009proof,
  title={A proof of security of Yao’s protocol for two-party computation},
  author={Lindell, Yehuda and Pinkas, Benny},
  journal={Journal of cryptology},
  volume={22},
  number={2},
  pages={161--188},
  year={2009},
  publisher={Springer}
}

@incollection{long2020federated,
  title={Federated learning for open banking},
  author={Long, Guodong and Tan, Yue and Jiang, Jing and Zhang, Chengqi},
  booktitle={Federated learning},
  pages={240--254},
  year={2020},
  publisher={Springer}
}

@inproceedings{mcmahan2017communication,
  title={Communication-efficient learning of deep networks from decentralized data},
  author={McMahan, Brendan and Moore, Eider and Ramage, Daniel and Hampson, Seth and y Arcas, Blaise Aguera},
  booktitle={Artificial intelligence and statistics},
  pages={1273--1282},
  year={2017},
  organization={PMLR}
}

@article{janssen2020data,
  title={Data governance: Organizing data for trustworthy Artificial Intelligence},
  author={Janssen, Marijn and Brous, Paul and Estevez, Elsa and Barbosa, Luis S and Janowski, Tomasz},
  journal={Government Information Quarterly},
  volume={37},
  number={3},
  pages={101493},
  year={2020},
  publisher={Elsevier}
}

@inproceedings{lecuyer2019certified,
  title={Certified robustness to adversarial examples with differential privacy},
  author={Lecuyer, Mathias and Atlidakis, Vaggelis and Geambasu, Roxana and Hsu, Daniel and Jana, Suman},
  booktitle={2019 IEEE Symposium on Security and Privacy (SP)},
  pages={656--672},
  year={2019},
  organization={IEEE}
}

@article{karnin1983secret,
  title={On secret sharing systems},
  author={Karnin, Ehud and Greene, Jonathan and Hellman, Martin},
  journal={IEEE Transactions on Information Theory},
  volume={29},
  number={1},
  pages={35--41},
  year={1983},
  publisher={IEEE}
}

@inproceedings{micali1987play,
  title={How to play any mental game},
  author={Micali, Silvio and Goldreich, Oded and Wigderson, Avi},
  booktitle={Proceedings of the Nineteenth ACM Symp. on Theory of Computing, STOC},
  pages={218--229},
  year={1987},
  organization={ACM}
}

@article{samarati1998protecting,
  title={Protecting privacy when disclosing information: k-anonymity and its enforcement through generalization and suppression},
  author={Samarati, Pierangela and Sweeney, Latanya},
  year={1998},
  publisher={technical report, SRI International}
}

@inproceedings{martin2007worst,
  title={Worst-case background knowledge for privacy-preserving data publishing},
  author={Martin, David J and Kifer, Daniel and Machanavajjhala, Ashwin and Gehrke, Johannes and Halpern, Joseph Y},
  booktitle={2007 IEEE 23rd International Conference on Data Engineering},
  pages={126--135},
  year={2007},
  organization={IEEE}
}

@inproceedings{nergiz2007hiding,
  title={Hiding the presence of individuals from shared databases},
  author={Nergiz, Mehmet Ercan and Atzori, Maurizio and Clifton, Chris},
  booktitle={Proceedings of the 2007 ACM SIGMOD international conference on Management of data},
  pages={665--676},
  year={2007}
}

@article{knott2021crypten,
  title={Crypten: Secure multi-party computation meets machine learning},
  author={Knott, Brian and Venkataraman, Shobha and Hannun, Awni and Sengupta, Shubho and Ibrahim, Mark and van der Maaten, Laurens},
  journal={Advances in Neural Information Processing Systems},
  volume={34},
  year={2021}
}

@inproceedings{rathee2020cryptflow2,
  title={CrypTFlow2: Practical 2-party secure inference},
  author={Rathee, Deevashwer and Rathee, Mayank and Kumar, Nishant and Chandran, Nishanth and Gupta, Divya and Rastogi, Aseem and Sharma, Rahul},
  booktitle={Proceedings of the 2020 ACM SIGSAC Conference on Computer and Communications Security},
  pages={325--342},
  year={2020}
}

@inproceedings{agrawal2019quotient,
  title={QUOTIENT: two-party secure neural network training and prediction},
  author={Agrawal, Nitin and Shahin Shamsabadi, Ali and Kusner, Matt J and Gasc{\'o}n, Adri{\`a}},
  booktitle={Proceedings of the 2019 ACM SIGSAC Conference on Computer and Communications Security},
  pages={1231--1247},
  year={2019}
}

@article{ball2019garbled,
  title={Garbled Neural Networks are Practical.},
  author={Ball, Marshall and Carmer, Brent and Malkin, Tal and Rosulek, Mike and Schimanski, Nichole},
  journal={IACR Cryptol. ePrint Arch.},
  volume={2019},
  pages={338},
  year={2019}
}

@article{zhao2019secure,
  title={Secure multi-party computation: theory, practice and applications},
  author={Zhao, Chuan and Zhao, Shengnan and Zhao, Minghao and Chen, Zhenxiang and Gao, Chong-Zhi and Li, Hongwei and Tan, Yu-an},
  journal={Information Sciences},
  volume={476},
  pages={357--372},
  year={2019},
  publisher={Elsevier}
}

@inproceedings{yao1982protocols,
  title={Protocols for secure computations},
  author={Yao, Andrew C},
  booktitle={23rd annual symposium on foundations of computer science (sfcs 1982)},
  pages={160--164},
  year={1982},
  organization={IEEE}
}

@article{rieke2020future,
  title={The future of digital health with federated learning},
  author={Rieke, Nicola and Hancox, Jonny and Li, Wenqi and Milletari, Fausto and Roth, Holger R and Albarqouni, Shadi and Bakas, Spyridon and Galtier, Mathieu N and Landman, Bennett A and Maier-Hein, Klaus and others},
  journal={NPJ digital medicine},
  volume={3},
  number={1},
  pages={1--7},
  year={2020},
  publisher={Nature Publishing Group}
}

@article{sheller2020federated,
  title={Federated learning in medicine: facilitating multi-institutional collaborations without sharing patient data},
  author={Sheller, Micah J and Edwards, Brandon and Reina, G Anthony and Martin, Jason and Pati, Sarthak and Kotrotsou, Aikaterini and Milchenko, Mikhail and Xu, Weilin and Marcus, Daniel and Colen, Rivka R and others},
  journal={Scientific reports},
  volume={10},
  number={1},
  pages={1--12},
  year={2020},
  publisher={Nature Publishing Group}
}

@article{liu2020secure,
  title={A secure federated transfer learning framework},
  author={Liu, Yang and Kang, Yan and Xing, Chaoping and Chen, Tianjian and Yang, Qiang},
  journal={IEEE Intelligent Systems},
  volume={35},
  number={4},
  pages={70--82},
  year={2020},
  publisher={IEEE}
}

@article{hardy2017private,
  title={Private federated learning on vertically partitioned data via entity resolution and additively homomorphic encryption},
  author={Hardy, Stephen and Henecka, Wilko and Ivey-Law, Hamish and Nock, Richard and Patrini, Giorgio and Smith, Guillaume and Thorne, Brian},
  journal={arXiv:1711.10677},
  year={2017}
}

@inproceedings{gu2020federated,
  title={Federated doubly stochastic kernel learning for vertically partitioned data},
  author={Gu, Bin and Dang, Zhiyuan and Li, Xiang and Huang, Heng},
  booktitle={Proceedings of the 26th ACM SIGKDD International Conference on Knowledge Discovery \& Data Mining},
  pages={2483--2493},
  year={2020}
}

@article{yang2019federated,
  title={Federated machine learning: Concept and applications},
  author={Yang, Qiang and Liu, Yang and Chen, Tianjian and Tong, Yongxin},
  journal={ACM Transactions on Intelligent Systems and Technology (TIST)},
  volume={10},
  number={2},
  pages={1--19},
  year={2019},
  publisher={ACM New York, NY, USA}
}

@article{wei2020federated,
  title={Federated learning with differential privacy: Algorithms and performance analysis},
  author={Wei, Kang and Li, Jun and Ding, Ming and Ma, Chuan and Yang, Howard H and Farokhi, Farhad and Jin, Shi and Quek, Tony QS and Poor, H Vincent},
  journal={IEEE Transactions on Information Forensics and Security},
  volume={15},
  pages={3454--3469},
  year={2020},
  publisher={IEEE}
}

@article{gascon2017privacy,
  title={Privacy-preserving distributed linear regression on high-dimensional data},
  author={Gasc{\'o}n, Adri{\`a} and Schoppmann, Phillipp and Balle, Borja and Raykova, Mariana and Doerner, Jack and Zahur, Samee and Evans, David},
  journal={Proceedings on Privacy Enhancing Technologies},
  volume={2017},
  number={4},
  pages={345--364},
  year={2017},
  publisher={Sciendo}
}

@inproceedings{abadi2016deep,
  title={Deep learning with differential privacy},
  author={Abadi, Martin and Chu, Andy and Goodfellow, Ian and McMahan, H Brendan and Mironov, Ilya and Talwar, Kunal and Zhang, Li},
  booktitle={Proceedings of the 2016 ACM SIGSAC conference on computer and communications security},
  pages={308--318},
  year={2016}
}

@article{fletcher2019decision,
  title={Decision tree classification with differential privacy: A survey},
  author={Fletcher, Sam and Islam, Md Zahidul},
  journal={ACM Computing Surveys (CSUR)},
  volume={52},
  number={4},
  pages={1--33},
  year={2019},
  publisher={ACM New York, NY, USA}
}

@article{chen2011publishing,
  title={Publishing set-valued data via differential privacy},
  author={Chen, Rui and Mohammed, Noman and Fung, Benjamin CM and Desai, Bipin C and Xiong, Li},
  journal={Proceedings of the VLDB Endowment},
  volume={4},
  number={11},
  pages={1087--1098},
  year={2011},
  publisher={VLDB Endowment}
}

@article{wang2016real,
  title={Real-time and spatio-temporal crowd-sourced social network data publishing with differential privacy},
  author={Wang, Qian and Zhang, Yan and Lu, Xiao and Wang, Zhibo and Qin, Zhan and Ren, Kui},
  journal={IEEE Transactions on Dependable and Secure Computing},
  volume={15},
  number={4},
  pages={591--606},
  year={2016},
  publisher={IEEE}
}

@article{hong2014collaborative,
  title={Collaborative search log sanitization: Toward differential privacy and boosted utility},
  author={Hong, Yuan and Vaidya, Jaideep and Lu, Haibing and Karras, Panagiotis and Goel, Sanjay},
  journal={IEEE Transactions on Dependable and Secure Computing},
  volume={12},
  number={5},
  pages={504--518},
  year={2014},
  publisher={IEEE}
}

@article{chen2014correlated,
  title={Correlated network data publication via differential privacy},
  author={Chen, Rui and Fung, Benjamin and Yu, Philip S and Desai, Bipin C},
  journal={The VLDB Journal},
  volume={23},
  number={4},
  pages={653--676},
  year={2014},
  publisher={Springer}
}

@article{ren2018textsf,
  title={LoPub: high-dimensional crowdsourced data publication with local differential privacy},
  author={Ren, Xuebin and Yu, Chia-Mu and Yu, Weiren and Yang, Shusen and Yang, Xinyu and McCann, Julie A and Philip, S Yu},
  journal={IEEE Transactions on Information Forensics and Security},
  volume={13},
  number={9},
  pages={2151--2166},
  year={2018},
  publisher={IEEE}
}

@inproceedings{zhang2016anonymizing,
  title={Anonymizing query logs by differential privacy},
  author={Zhang, Sicong and Yang, Hui and Singh, Lisa},
  booktitle={Proceedings of the 39th International ACM SIGIR conference on Research and Development in Information Retrieval},
  pages={753--756},
  year={2016}
}

@article{dwork2016calibrating,
  title={Calibrating noise to sensitivity in private data analysis},
  author={Dwork, Cynthia and McSherry, Frank and Nissim, Kobbi and Smith, Adam},
  journal={Journal of Privacy and Confidentiality},
  volume={7},
  number={3},
  pages={17--51},
  year={2016}
}

@article{dwork2014algorithmic,
  title={The algorithmic foundations of differential privacy.},
  author={Dwork, Cynthia and Roth, Aaron and others},
  journal={Foundations and Trends in Theoretical Computer Science},
  volume={9},
  number={3-4},
  pages={211--407},
  year={2014}
}

@inproceedings{dwork2006calibrating,
  title={Calibrating noise to sensitivity in private data analysis},
  author={Dwork, Cynthia and McSherry, Frank and Nissim, Kobbi and Smith, Adam},
  booktitle={Theory of cryptography conference},
  pages={265--284},
  year={2006},
  organization={Springer}
}

@inproceedings{ji2015secgraph,
  title={Secgraph: A uniform and open-source evaluation system for graph data anonymization and de-anonymization},
  author={Ji, Shouling and Li, Weiqing and Mittal, Prateek and Hu, Xin and Beyah, Raheem},
  booktitle={24th $\{$USENIX$\}$ Security Symposium ($\{$USENIX$\}$ Security 15)},
  pages={303--318},
  year={2015}
}

@article{ji2016graph,
  title={Graph data anonymization, de-anonymization attacks, and de-anonymizability quantification: A survey},
  author={Ji, Shouling and Mittal, Prateek and Beyah, Raheem},
  journal={IEEE Communications Surveys \& Tutorials},
  volume={19},
  number={2},
  pages={1305--1326},
  year={2016},
  publisher={IEEE}
}

@inproceedings{liu2008towards,
  title={Towards identity anonymization on graphs},
  author={Liu, Kun and Terzi, Evimaria},
  booktitle={Proceedings of the 2008 ACM SIGMOD international conference on Management of data},
  pages={93--106},
  year={2008}
}

@article{el2011systematic,
  title={A systematic review of re-identification attacks on health data},
  author={El Emam, Khaled and Jonker, Elizabeth and Arbuckle, Luk and Malin, Bradley},
  journal={PloS one},
  volume={6},
  number={12},
  pages={e28071},
  year={2011},
  publisher={Public Library of Science San Francisco, USA}
}

@inproceedings{ji2014structural,
  title={Structural data de-anonymization: Quantification, practice, and implications},
  author={Ji, Shouling and Li, Weiqing and Srivatsa, Mudhakar and Beyah, Raheem},
  booktitle={Proceedings of the 2014 ACM SIGSAC conference on computer and communications security},
  pages={1040--1053},
  year={2014}
}

@article{dirik2014analysis,
  title={Analysis of seam-carving-based anonymization of images against PRNU noise pattern-based source attribution},
  author={Dirik, Ahmet Emir and Sencar, H{\"u}srev Taha and Memon, Nasir},
  journal={IEEE Transactions on Information Forensics and Security},
  volume={9},
  number={12},
  pages={2277--2290},
  year={2014},
  publisher={IEEE}
}

@inproceedings{maximov2020ciagan,
  title={Ciagan: Conditional identity anonymization generative adversarial networks},
  author={Maximov, Maxim and Elezi, Ismail and Leal-Taix{\'e}, Laura},
  booktitle={Proceedings of the IEEE/CVF Conference on Computer Vision and Pattern Recognition},
  pages={5447--5456},
  year={2020}
}

@article{he2009anonymization,
  title={Anonymization of set-valued data via top-down, local generalization},
  author={He, Yeye and Naughton, Jeffrey F},
  journal={Proceedings of the VLDB Endowment},
  volume={2},
  number={1},
  pages={934--945},
  year={2009},
  publisher={VLDB Endowment}
}

@article{terrovitis2008privacy,
  title={Privacy-preserving anonymization of set-valued data},
  author={Terrovitis, Manolis and Mamoulis, Nikos and Kalnis, Panos},
  journal={Proceedings of the VLDB Endowment},
  volume={1},
  number={1},
  pages={115--125},
  year={2008},
  publisher={VLDB Endowment}
}

@article{pervaiz2013accuracy,
  title={Accuracy-constrained privacy-preserving access control mechanism for relational data},
  author={Pervaiz, Zahid and Aref, Walid G and Ghafoor, Arif and Prabhu, Nagabhushana},
  journal={IEEE Transactions on Knowledge and Data Engineering},
  volume={26},
  number={4},
  pages={795--807},
  year={2013},
  publisher={IEEE}
}

@article{beigi2020survey,
  title={A survey on privacy in social media: Identification, mitigation, and applications},
  author={Beigi, Ghazaleh and Liu, Huan},
  journal={ACM Transactions on Data Science},
  volume={1},
  number={1},
  pages={1--38},
  year={2020},
  publisher={ACM New York, NY, USA}
}

@inproceedings{dwork2008differential,
  title={Differential privacy: A survey of results},
  author={Dwork, Cynthia},
  booktitle={International conference on theory and applications of models of computation},
  pages={1--19},
  year={2008},
  organization={Springer}
}

@article{ji2014differential,
  title={Differential privacy and machine learning: a survey and review},
  author={Ji, Zhanglong and Lipton, Zachary C and Elkan, Charles},
  journal={arXiv:1412.7584},
  year={2014}
}

@inproceedings{al2015literature,
  title={A literature survey and classifications on data deanonymisation},
  author={Al-Azizy, Dalal and Millard, David and Symeonidis, Iraklis and O’Hara, Kieron and Shadbolt, Nigel},
  booktitle={International Conference on Risks and Security of Internet and Systems},
  pages={36--51},
  year={2015},
  organization={Springer}
}

@article{majeed2020anonymization,
  title={Anonymization techniques for privacy preserving data publishing: A comprehensive survey},
  author={Majeed, Abdul and Lee, Sungchang},
  journal={IEEE Access},
  volume={9},
  pages={8512--8545},
  year={2020},
  publisher={IEEE}
}

@article{zhou2008brief,
  title={A brief survey on anonymization techniques for privacy preserving publishing of social network data},
  author={Zhou, Bin and Pei, Jian and Luk, WoShun},
  journal={ACM Sigkdd Explorations Newsletter},
  volume={10},
  number={2},
  pages={12--22},
  year={2008},
  publisher={ACM New York, NY, USA}
}

@article{jakob2020design,
  title={Design and evaluation of a data anonymization pipeline to promote Open Science on COVID-19},
  author={Jakob, Carolin EM and Kohlmayer, Florian and Meurers, Thierry and Vehreschild, J{\"o}rg Janne and Prasser, Fabian},
  journal={Scientific data},
  volume={7},
  number={1},
  pages={1--10},
  year={2020},
  publisher={Nature Publishing Group}
}

@inproceedings{mohassel2017secureml,
  title={Secureml: A system for scalable privacy-preserving machine learning},
  author={Mohassel, Payman and Zhang, Yupeng},
  booktitle={2017 IEEE Symposium on Security and Privacy (SP)},
  pages={19--38},
  year={2017},
  organization={IEEE}
}

@article{shi2016secure,
  title={Secure multi-pArty computation grid LOgistic REgression (SMAC-GLORE)},
  author={Shi, Haoyi and Jiang, Chao and Dai, Wenrui and Jiang, Xiaoqian and Tang, Yuzhe and Ohno-Machado, Lucila and Wang, Shuang},
  journal={BMC medical informatics and decision making},
  volume={16},
  number={3},
  pages={175--187},
  year={2016},
  publisher={Springer}
}

@inproceedings{erlingsson2014rappor,
  title={Rappor: Randomized aggregatable privacy-preserving ordinal response},
  author={Erlingsson, {\'U}lfar and Pihur, Vasyl and Korolova, Aleksandra},
  booktitle={Proceedings of the 2014 ACM SIGSAC conference on computer and communications security},
  pages={1054--1067},
  year={2014}
}

@article{zhang2013scalable,
  title={A scalable two-phase top-down specialization approach for data anonymization using mapreduce on cloud},
  author={Zhang, Xuyun and Yang, Laurence T and Liu, Chang and Chen, Jinjun},
  journal={IEEE Transactions on Parallel and Distributed Systems},
  volume={25},
  number={2},
  pages={363--373},
  year={2013},
  publisher={IEEE}
}

@article{yuan2011protecting,
  title={Protecting sensitive labels in social network data anonymization},
  author={Yuan, Mingxuan and Chen, Lei and Philip, S Yu and Yu, Ting},
  journal={IEEE Transactions on Knowledge and Data Engineering},
  volume={25},
  number={3},
  pages={633--647},
  year={2011},
  publisher={IEEE}
}

@inproceedings{carlini2021extracting,
  title={Extracting training data from large language models},
  author={Carlini, Nicholas and Tramer, Florian and Wallace, Eric and Jagielski, Matthew and Herbert-Voss, Ariel and Lee, Katherine and Roberts, Adam and Brown, Tom and Song, Dawn and Erlingsson, Ulfar and others},
  booktitle={30th USENIX Security Symposium (USENIX Security 21)},
  pages={2633--2650},
  year={2021}
}

@inproceedings{eykholt2018robust,
  title={Robust physical-world attacks on deep learning visual classification},
  author={Eykholt, Kevin and Evtimov, Ivan and Fernandes, Earlence and Li, Bo and Rahmati, Amir and Xiao, Chaowei and Prakash, Atul and Kohno, Tadayoshi and Song, Dawn},
  booktitle={Proceedings of the IEEE conference on computer vision and pattern recognition},
  pages={1625--1634},
  year={2018}
}

@article{gunning2019darpa,
  title={DARPA’s explainable artificial intelligence (XAI) program},
  author={Gunning, David and Aha, David},
  journal={AI magazine},
  volume={40},
  number={2},
  pages={44--58},
  year={2019}
}

@misc{openai2018openai,
  author={{OpenAI}},
  title = {{OpenAI} Charter},
  howpublished = {\url{https://openai.com/charter/}},
  year={2018},
  note = {Accessed: 2021-02-20}
}

@misc{ibm-trusting,
  author={{IBM}},
  title = {Trusting {AI}},
  howpublished = {\url{https://www.research.ibm.com/artificial-intelligence/trusted-ai/}},
  note = {Accessed: 2021-02-20}
}

@misc{aihleg2018assessment,
  author={{AI HLEG}},
  title = {Assessment List for Trustworthy Artificial Intelligence (ALTAI) for self-assessment},
  year={2020}
}

@misc{aihleg2018ethics,
  author={{AI HLEG}},
  title = {Ethics Guidelines for Trustworthy {AI}},
  year={2018},
  note = {Accessed: 2021-02-20}
}

@misc{hao2019making,
  author={Karen Hao},
  title = {Making face recognition less biased doesn't make it less scary},
  howpublished = {{MIT Technology Review}, \url{https://www.technologyreview.com/2019/01/29/137676/making-face-recognition-less-biased-doesnt-make-it-less-scary/}},
  year={2019},
  journal={MIT Technology Review},
  note = {Accessed: 2021-02-20}
}

@misc{hao2019ai,
  author={Karen Hao},
  title = {AI is sending people to jail-and getting it wrong},
  howpublished = {{MIT Technology Review}, \url{https://www.technologyreview.com/2019/01/21/137783/algorithms-criminal-justice-ai/}},
  year={2019},
  journal={MIT Technology Review},
  note = {Accessed: 2021-02-20}
}

@inproceedings{raji2020closing,
  title={Closing the AI accountability gap: Defining an end-to-end framework for internal algorithmic auditing},
  author={Raji, Inioluwa Deborah and Smart, Andrew and White, Rebecca N and Mitchell, Margaret and Gebru, Timnit and Hutchinson, Ben and Smith-Loud, Jamila and Theron, Daniel and Barnes, Parker},
  booktitle={Proceedings of the 2020 conference on fairness, accountability, and transparency},
  pages={33--44},
  year={2020}
}

@article{brundage2020toward,
  title={Toward trustworthy AI development: mechanisms for supporting verifiable claims},
  author={Brundage, Miles and Avin, Shahar and Wang, Jasmine and Belfield, Haydn and Krueger, Gretchen and Hadfield, Gillian and Khlaaf, Heidy and Yang, Jingying and Toner, Helen and Fong, Ruth and others},
  journal={arXiv:2004.07213},
  year={2020}
}

@inproceedings{boemer2020mp2ml,
  title={MP2ML: A mixed-protocol machine learning framework for private inference},
  author={Boemer, Fabian and Cammarota, Rosario and Demmler, Daniel and Schneider, Thomas and Yalame, Hossein},
  booktitle={Proceedings of the 15th International Conference on Availability, Reliability and Security},
  pages={1--10},
  year={2020}
}

@inproceedings{kumar2020cryptflow,
  title={Cryptflow: Secure tensorflow inference},
  author={Kumar, Nishant and Rathee, Mayank and Chandran, Nishanth and Gupta, Divya and Rastogi, Aseem and Sharma, Rahul},
  booktitle={2020 IEEE Symposium on Security and Privacy (SP)},
  pages={336--353},
  year={2020},
  organization={IEEE}
}

@article{treiber2020cryptospn,
  title={CryptoSPN: Privacy-preserving Sum-Product Network Inference},
  author={Treiber, Amos and Molina, Alejandro and Weinert, Christian and Schneider, Thomas and Kersting, Kristian},
  journal={arXiv:2002.00801},
  year={2020}
}

@article{seshia2016towards,
  title={Towards verified artificial intelligence},
  author={Seshia, Sanjit A and Sadigh, Dorsa and Sastry, S Shankar},
  journal={arXiv:1606.08514},
  year={2016}
}

@inproceedings{dreossi2019verifai,
  title={Verifai: A toolkit for the formal design and analysis of artificial intelligence-based systems},
  author={Dreossi, Tommaso and Fremont, Daniel J and Ghosh, Shromona and Kim, Edward and Ravanbakhsh, Hadi and Vazquez-Chanlatte, Marcell and Seshia, Sanjit A},
  booktitle={International Conference on Computer Aided Verification},
  pages={432--442},
  year={2019},
  organization={Springer}
}

@article{brogle2019hardware,
  title={Hardware-in-the-loop autonomous driving simulation without real-time constraints},
  author={Brogle, Craig and Zhang, Chao and Lim, Kai Li and Br{\"a}unl, Thomas},
  journal={IEEE Transactions on Intelligent Vehicles},
  volume={4},
  number={3},
  pages={375--384},
  year={2019},
  publisher={IEEE}
}

@article{hendrycks2019benchmarking,
  title={Benchmarking neural network robustness to common corruptions and perturbations},
  author={Hendrycks, Dan and Dietterich, Thomas},
  journal={arXiv:1903.12261},
  year={2019}
}

@inproceedings{tae2019data,
  title={Data cleaning for accurate, fair, and robust models: A big data-AI integration approach},
  author={Tae, Ki Hyun and Roh, Yuji and Oh, Young Hun and Kim, Hyunsu and Whang, Steven Euijong},
  booktitle={Proceedings of the 3rd International Workshop on Data Management for End-to-End Machine Learning},
  pages={1--4},
  year={2019}
}

@article{wing2021trustworthy,
  title={Trustworthy ai},
  author={Wing, Jeannette M},
  journal={Communications of the ACM},
  volume={64},
  number={10},
  pages={64--71},
  year={2021},
  publisher={ACM New York, NY, USA}
}

@article{amodei2016concrete,
  title={Concrete problems in AI safety},
  author={Amodei, Dario and Olah, Chris and Steinhardt, Jacob and Christiano, Paul and Schulman, John and Man{\'e}, Dan},
  journal={arXiv:1606.06565},
  year={2016}
}

@article{nori2019interpretml,
  title={InterpretML: A Unified Framework for Machine Learning Interpretability},
  author={Nori, Harsha and Jenkins, Samuel and Koch, Paul and Caruana, Rich},
  journal={arXiv:1909.09223},
  year={2019}
}

@article{askell2019role,
  title={The role of cooperation in responsible AI development},
  author={Askell, Amanda and Brundage, Miles and Hadfield, Gillian},
  journal={arXiv:1907.04534},
  year={2019}
}

@inproceedings{amershi2019software,
  title={Software engineering for machine learning: A case study},
  author={Amershi, Saleema and Begel, Andrew and Bird, Christian and DeLine, Robert and Gall, Harald and Kamar, Ece and Nagappan, Nachiappan and Nushi, Besmira and Zimmermann, Thomas},
  booktitle={2019 IEEE/ACM 41st International Conference on Software Engineering: Software Engineering in Practice (ICSE-SEIP)},
  pages={291--300},
  year={2019},
  organization={IEEE}
}

@article{zhang2020machine,
  title={Machine learning testing: Survey, landscapes and horizons},
  author={Zhang, Jie M and Harman, Mark and Ma, Lei and Liu, Yang},
  journal={IEEE Transactions on Software Engineering},
  year={2020},
  publisher={IEEE}
}

@inproceedings{pei2017deepxplore,
  title={Deepxplore: Automated whitebox testing of deep learning systems},
  author={Pei, Kexin and Cao, Yinzhi and Yang, Junfeng and Jana, Suman},
  booktitle={proceedings of the 26th Symposium on Operating Systems Principles},
  pages={1--18},
  year={2017}
}

@inproceedings{zhang2018deeproad,
  title={Deeproad: Gan-based metamorphic testing and input validation framework for autonomous driving systems},
  author={Zhang, Mengshi and Zhang, Yuqun and Zhang, Lingming and Liu, Cong and Khurshid, Sarfraz},
  booktitle={2018 33rd IEEE/ACM International Conference on Automated Software Engineering (ASE)},
  pages={132--142},
  year={2018},
  organization={IEEE}
}

@inproceedings{mitchell2019model,
  title={Model cards for model reporting},
  author={Mitchell, Margaret and Wu, Simone and Zaldivar, Andrew and Barnes, Parker and Vasserman, Lucy and Hutchinson, Ben and Spitzer, Elena and Raji, Inioluwa Deborah and Gebru, Timnit},
  booktitle={Proceedings of the conference on fairness, accountability, and transparency},
  pages={220--229},
  year={2019}
}

@inproceedings{cysneiros2018software,
  title={Software transparency as a key requirement for self-driving cars},
  author={Cysneiros, Luiz Marcio and Raffi, Majid and do Prado Leite, Julio Cesar Sampaio},
  booktitle={2018 IEEE 26th international requirements engineering conference (RE)},
  pages={382--387},
  year={2018},
  organization={IEEE}
}

@inproceedings{brun2018software,
  title={Software fairness},
  author={Brun, Yuriy and Meliou, Alexandra},
  booktitle={Proceedings of the 2018 26th ACM joint meeting on european software engineering conference and symposium on the foundations of software engineering},
  pages={754--759},
  year={2018}
}

@inproceedings{verma2018fairness,
  title={Fairness definitions explained},
  author={Verma, Sahil and Rubin, Julia},
  booktitle={2018 ieee/acm international workshop on software fairness (fairware)},
  pages={1--7},
  year={2018},
  organization={IEEE}
}

@inproceedings{madaio2020co,
  title={Co-designing checklists to understand organizational challenges and opportunities around fairness in AI},
  author={Madaio, Michael A and Stark, Luke and Wortman Vaughan, Jennifer and Wallach, Hanna},
  booktitle={Proceedings of the 2020 CHI Conference on Human Factors in Computing Systems},
  pages={1--14},
  year={2020}
}

@inproceedings{croce2021robustbench,
  title={RobustBench: a standardized adversarial robustness benchmark},
  author={Croce, Francesco and Andriushchenko, Maksym and Sehwag, Vikash and Debenedetti, Edoardo and Flammarion, Nicolas and Chiang, Mung and Mittal, Prateek and Hein, Matthias},
  booktitle={Thirty-fifth Conference on Neural Information Processing Systems Datasets and Benchmarks Track (Round 2)},
  year={2021}
}

@inproceedings{galhotra2017fairness,
  title={Fairness testing: testing software for discrimination},
  author={Galhotra, Sainyam and Brun, Yuriy and Meliou, Alexandra},
  booktitle={Proceedings of the 2017 11th Joint meeting on foundations of software engineering},
  pages={498--510},
  year={2017}
}

@inproceedings{tramer2017fairtest,
  title={Fairtest: Discovering unwarranted associations in data-driven applications},
  author={Tramer, Florian and Atlidakis, Vaggelis and Geambasu, Roxana and Hsu, Daniel and Hubaux, Jean-Pierre and Humbert, Mathias and Juels, Ari and Lin, Huang},
  booktitle={2017 IEEE European Symposium on Security and Privacy (EuroS\&P)},
  pages={401--416},
  year={2017},
  organization={IEEE}
}

@article{guidotti2018survey,
  title={A survey of methods for explaining black box models},
  author={Guidotti, Riccardo and Monreale, Anna and Ruggieri, Salvatore and Turini, Franco and Giannotti, Fosca and Pedreschi, Dino},
  journal={ACM computing surveys (CSUR)},
  volume={51},
  number={5},
  pages={1--42},
  year={2018},
  publisher={ACM New York, NY, USA}
}

@report{bogen2018help,
  title={HELP WANTED: An Examination of Hiring Algorithms, Equity, and Bias},
  author={Bogen, Miranda and Rieke, Aaron},
  publisher={Upturn},
  year={2018}
}

@article{dressel2018accuracy,
  title={The accuracy, fairness, and limits of predicting recidivism},
  author={Dressel, Julia and Farid, Hany},
  journal={Science advances},
  volume={4},
  number={1},
  pages={eaao5580},
  year={2018},
  publisher={American Association for the Advancement of Science}
}

@article{saleiro2018aequitas,
  title={Aequitas: A bias and fairness audit toolkit},
  author={Saleiro, Pedro and Kuester, Benedict and Hinkson, Loren and London, Jesse and Stevens, Abby and Anisfeld, Ari and Rodolfa, Kit T and Ghani, Rayid},
  journal={arXiv:1811.05577},
  year={2018}
}

@article{caton2020fairness,
  title={Fairness in Machine Learning: A Survey},
  author={Caton, Simon and Haas, Christian},
  journal={arXiv:2010.04053},
  year={2020}
}

@inproceedings{sixta2020fairface,
  title={Fairface challenge at eccv 2020: Analyzing bias in face recognition},
  author={Sixta, Tom{\'a}{\v{s}} and Jacques Junior, Julio and Buch-Cardona, Pau and Vazquez, Eduard and Escalera, Sergio},
  booktitle={European conference on computer vision},
  pages={463--481},
  year={2020},
  organization={Springer}
}

@inproceedings{zemel2013learning,
  title={Learning fair representations},
  author={Zemel, Rich and Wu, Yu and Swersky, Kevin and Pitassi, Toni and Dwork, Cynthia},
  booktitle={International conference on machine learning},
  pages={325--333},
  year={2013},
  organization={PMLR}
}

@article{howard2018ugly,
  title={The ugly truth about ourselves and our robot creations: the problem of bias and social inequity},
  author={Howard, Ayanna and Borenstein, Jason},
  journal={Science and engineering ethics},
  volume={24},
  number={5},
  pages={1521--1536},
  year={2018},
  publisher={Springer}
}

@article{drozdowski2020demographic,
  title={Demographic bias in biometrics: A survey on an emerging challenge},
  author={Drozdowski, Pawel and Rathgeb, Christian and Dantcheva, Antitza and Damer, Naser and Busch, Christoph},
  journal={IEEE Transactions on Technology and Society},
  volume={1},
  number={2},
  pages={89--103},
  year={2020},
  publisher={IEEE}
}

@article{hardt2016equality,
  title={Equality of opportunity in supervised learning},
  author={Hardt, Moritz and Price, Eric and Srebro, Nati},
  journal={Advances in neural information processing systems},
  volume={29},
  year={2016}
}

@inproceedings{iosifidis2019fae,
  title={Fae: A fairness-aware ensemble framework},
  author={Iosifidis, Vasileios and Fetahu, Besnik and Ntoutsi, Eirini},
  booktitle={2019 IEEE International Conference on Big Data (Big Data)},
  pages={1375--1380},
  year={2019},
  organization={IEEE}
}

@article{cao2019adversarial,
  title={Adversarial objects against LiDAR-based autonomous driving systems},
  author={Cao, Yulong and Xiao, Chaowei and Yang, Dawei and Fang, Jing and Yang, Ruigang and Liu, Mingyan and Li, Bo},
  journal={Journal of Environmental Sciences (China) English Ed},
  year={2019},
  publisher={Chinese Academy of Sciences}
}

@article{arrieta2020explainable,
  title={Explainable Artificial Intelligence (XAI): Concepts, taxonomies, opportunities and challenges toward responsible AI},
  author={Arrieta, Alejandro Barredo and D{\'\i}az-Rodr{\'\i}guez, Natalia and Del Ser, Javier and Bennetot, Adrien and Tabik, Siham and Barbado, Alberto and Garc{\'\i}a, Salvador and Gil-L{\'o}pez, Sergio and Molina, Daniel and Benjamins, Richard and others},
  journal={Information fusion},
  volume={58},
  pages={82--115},
  year={2020},
  publisher={Elsevier}
}

@article{rosenfeld2019explainability,
  title={Explainability in human--agent systems},
  author={Rosenfeld, Avi and Richardson, Ariella},
  journal={Autonomous Agents and Multi-Agent Systems},
  volume={33},
  number={6},
  pages={673--705},
  year={2019},
  publisher={Springer}
}

@article{kleinberg2016inherent,
  title={Inherent trade-offs in the fair determination of risk scores},
  author={Kleinberg, Jon and Mullainathan, Sendhil and Raghavan, Manish},
  journal={arXiv:1609.05807},
  year={2016}
}

@article{lipton2018mythos,
  title={The Mythos of Model Interpretability: In machine learning, the concept of interpretability is both important and slippery.},
  author={Lipton, Zachary C},
  journal={Queue},
  volume={16},
  number={3},
  pages={31--57},
  year={2018},
  publisher={ACM New York, NY, USA}
}

@article{hoffman2018metrics,
  title={Metrics for explainable AI: Challenges and prospects},
  author={Hoffman, Robert R and Mueller, Shane T and Klein, Gary and Litman, Jordan},
  journal={arXiv:1812.04608},
  year={2018}
}

@article{mohseni2021multidisciplinary,
  title={A multidisciplinary survey and framework for design and evaluation of explainable AI systems},
  author={Mohseni, Sina and Zarei, Niloofar and Ragan, Eric D},
  journal={ACM Transactions on Interactive Intelligent Systems (TiiS)},
  volume={11},
  number={3-4},
  pages={1--45},
  year={2021},
  publisher={ACM New York, NY}
}

@article{bertrand2004emily,
  title={Are Emily and Greg more employable than Lakisha and Jamal? A field experiment on labor market discrimination},
  author={Bertrand, Marianne and Mullainathan, Sendhil},
  journal={American economic review},
  volume={94},
  number={4},
  pages={991--1013},
  year={2004}
}

@article{chouldechova2017fair,
  title={Fair prediction with disparate impact: A study of bias in recidivism prediction instruments},
  author={Chouldechova, Alexandra},
  journal={Big data},
  volume={5},
  number={2},
  pages={153--163},
  year={2017},
  publisher={Mary Ann Liebert, Inc. 140 Huguenot Street, 3rd Floor New Rochelle, NY 10801 USA}
}

@article{bodria2021benchmarking,
  title={Benchmarking and Survey of Explanation Methods for Black Box Models},
  author={Bodria, Francesco and Giannotti, Fosca and Guidotti, Riccardo and Naretto, Francesca and Pedreschi, Dino and Rinzivillo, Salvatore},
  journal={arXiv:2102.13076},
  year={2021}
}

@article{miller2019explanation,
  title={Explanation in artificial intelligence: Insights from the social sciences},
  author={Miller, Tim},
  journal={Artificial intelligence},
  volume={267},
  pages={1--38},
  year={2019},
  publisher={Elsevier}
}

@inproceedings{kulesza2012tell,
  title={Tell me more? The effects of mental model soundness on personalizing an intelligent agent},
  author={Kulesza, Todd and Stumpf, Simone and Burnett, Margaret and Kwan, Irwin},
  booktitle={Proceedings of the SIGCHI Conference on Human Factors in Computing Systems},
  pages={1--10},
  year={2012}
}

@article{goodall2018situ,
  title={Situ: Identifying and explaining suspicious behavior in networks},
  author={Goodall, John R and Ragan, Eric D and Steed, Chad A and Reed, Joel W and Richardson, G David and Huffer, Kelly MT and Bridges, Robert A and Laska, Jason A},
  journal={IEEE transactions on visualization and computer graphics},
  volume={25},
  number={1},
  pages={204--214},
  year={2018},
  publisher={IEEE}
}

@inproceedings{ribeiro2016should,
  title={" Why should i trust you?" Explaining the predictions of any classifier},
  author={Ribeiro, Marco Tulio and Singh, Sameer and Guestrin, Carlos},
  booktitle={Proceedings of the 22nd ACM SIGKDD international conference on knowledge discovery and data mining},
  pages={1135--1144},
  year={2016}
}

@article{guidotti2018local,
  title={Local rule-based explanations of black box decision systems},
  author={Guidotti, Riccardo and Monreale, Anna and Ruggieri, Salvatore and Pedreschi, Dino and Turini, Franco and Giannotti, Fosca},
  journal={arXiv:1805.10820},
  year={2018}
}

@article{lundberg2017unified,
  title={A Unified Approach to Interpreting Model Predictions},
  author={Lundberg, Scott M and Lee, Su-In},
  journal={Advances in Neural Information Processing Systems},
  volume={30},
  pages={4765--4774},
  year={2017}
}

@inproceedings{ribeiro2018anchors,
  title={Anchors: High-precision model-agnostic explanations},
  author={Ribeiro, Marco Tulio and Singh, Sameer and Guestrin, Carlos},
  booktitle={Proceedings of the AAAI conference on artificial intelligence},
  volume={32},
  number={1},
  year={2018}
}

@inproceedings{alvarez2018towards,
  title={Towards robust interpretability with self-explaining neural networks},
  author={Alvarez-Melis, David and Jaakkola, Tommi S},
  booktitle={Proceedings of the 32nd International Conference on Neural Information Processing Systems},
  pages={7786--7795},
  year={2018}
}

@inproceedings{wang2003measurement,
  title={Measurement of the cognitive functional complexity of software},
  author={Wang, Yingxu and Shao, Jingqiu},
  booktitle={The Second IEEE International Conference on Cognitive Informatics, 2003. Proceedings.},
  pages={67--74},
  year={2003},
  organization={IEEE}
}

@article{kim2018introduction,
  title={Introduction to interpretable machine learning},
  author={Kim, Been and Doshi-Velez, F},
  journal={Proceedings of the CVPR 2018 Tutorial on Interpretable Machine Learning for Computer Vision, Salt Lake City, UT, USA},
  volume={18},
  year={2018}
}

@article{ayvaz2019witness,
  title={Witness of Things: Blockchain-based distributed decision record-keeping system for autonomous vehicles},
  author={Ayvaz, Serkan and Cetin, Salih Cemil},
  journal={International Journal of Intelligent Unmanned Systems},
  year={2019},
  publisher={Emerald Publishing Limited}
}

@inproceedings{cysneiros2009initial,
  title={An Initial Analysis on How Software Transparency and Trust Influence each other.},
  author={Cysneiros, Luiz Marcio and Werneck, Vera Maria Benjamim},
  booktitle={WER},
  year={2009},
  organization={Citeseer}
}

@article{leite2010software,
  title={Software transparency},
  author={Leite, Julio Cesar Sampaio do Prado and Cappelli, Claudia},
  journal={Business \& Information Systems Engineering},
  volume={2},
  number={3},
  pages={127--139},
  year={2010},
  publisher={Springer}
}

@article{gebru2021datasheets,
  title={Datasheets for datasets},
  author={Gebru, Timnit and Morgenstern, Jamie and Vecchione, Briana and Vaughan, Jennifer Wortman and Wallach, Hanna and Iii, Hal Daum{\'e} and Crawford, Kate},
  journal={Communications of the ACM},
  volume={64},
  number={12},
  pages={86--92},
  year={2021},
  publisher={ACM New York, NY, USA}
}

@inproceedings{dovsilovic2018explainable,
  title={Explainable artificial intelligence: A survey},
  author={Do{\v{s}}ilovi{\'c}, Filip Karlo and Br{\v{c}}i{\'c}, Mario and Hlupi{\'c}, Nikica},
  booktitle={2018 41st International convention on information and communication technology, electronics and microelectronics (MIPRO)},
  pages={0210--0215},
  year={2018},
  organization={IEEE}
}

@inproceedings{tramer2016stealing,
  title={Stealing Machine Learning Models via Prediction $\{$APIs$\}$},
  author={Tram{\`e}r, Florian and Zhang, Fan and Juels, Ari and Reiter, Michael K and Ristenpart, Thomas},
  booktitle={25th USENIX security symposium (USENIX Security 16)},
  pages={601--618},
  year={2016}
}

@article{pleiss2017fairness,
  title={On fairness and calibration},
  author={Pleiss, Geoff and Raghavan, Manish and Wu, Felix and Kleinberg, Jon and Weinberger, Kilian Q},
  journal={Advances in neural information processing systems},
  volume={30},
  year={2017}
}

@article{berk2021fairness,
  title={Fairness in criminal justice risk assessments: The state of the art},
  author={Berk, Richard and Heidari, Hoda and Jabbari, Shahin and Kearns, Michael and Roth, Aaron},
  journal={Sociological Methods \& Research},
  volume={50},
  number={1},
  pages={3--44},
  year={2021},
  publisher={Sage Publications Sage CA: Los Angeles, CA}
}

@incollection{kurakin2018adversarial,
  title={Adversarial examples in the physical world},
  author={Kurakin, Alexey and Goodfellow, Ian J and Bengio, Samy},
  booktitle={Artificial intelligence safety and security},
  pages={99--112},
  year={2018},
  publisher={Chapman and Hall/CRC}
}

@inproceedings{carlini2017towards,
  title={Towards evaluating the robustness of neural networks},
  author={Carlini, Nicholas and Wagner, David},
  booktitle={2017 ieee symposium on security and privacy (sp)},
  pages={39--57},
  year={2017},
  organization={IEEE}
}

@inproceedings{wang2019neural,
  title={Neural cleanse: Identifying and mitigating backdoor attacks in neural networks},
  author={Wang, Bolun and Yao, Yuanshun and Shan, Shawn and Li, Huiying and Viswanath, Bimal and Zheng, Haitao and Zhao, Ben Y},
  booktitle={2019 IEEE Symposium on Security and Privacy (SP)},
  pages={707--723},
  year={2019},
  organization={IEEE}
}

@inproceedings{liu2018fine,
  title={Fine-pruning: Defending against backdooring attacks on deep neural networks},
  author={Liu, Kang and Dolan-Gavitt, Brendan and Garg, Siddharth},
  booktitle={International Symposium on Research in Attacks, Intrusions, and Defenses},
  pages={273--294},
  year={2018},
  organization={Springer}
}

@inproceedings{madry2018towards,
  title={Towards Deep Learning Models Resistant to Adversarial Attacks},
  author={Madry, Aleksander and Makelov, Aleksandar and Schmidt, Ludwig and Tsipras, Dimitris and Vladu, Adrian},
  booktitle={International Conference on Learning Representations},
  year={2018}
}

@inproceedings{tramer2018ensemble,
  title={Ensemble Adversarial Training: Attacks and Defenses},
  author={Tram{\`e}r, Florian and Kurakin, Alexey and Papernot, Nicolas and Goodfellow, Ian and Boneh, Dan and McDaniel, Patrick},
  booktitle={International Conference on Learning Representations},
  year={2018}
}

@inproceedings{kim2016examples,
  title={Examples are not enough, learn to criticize! Criticism for Interpretability.},
  author={Kim, Been and Koyejo, Oluwasanmi and Khanna, Rajiv and others},
  booktitle={NIPS},
  pages={2280--2288},
  year={2016}
}

@inproceedings{sundararajan2017axiomatic,
  title={Axiomatic attribution for deep networks},
  author={Sundararajan, Mukund and Taly, Ankur and Yan, Qiqi},
  booktitle={International Conference on Machine Learning},
  pages={3319--3328},
  year={2017},
  organization={PMLR}
}

@inproceedings{shrikumar2017learning,
  title={Learning important features through propagating activation differences},
  author={Shrikumar, Avanti and Greenside, Peyton and Kundaje, Anshul},
  booktitle={International Conference on Machine Learning},
  pages={3145--3153},
  year={2017},
  organization={PMLR}
}

@article{mehrabi2021survey,
  title={A survey on bias and fairness in machine learning},
  author={Mehrabi, Ninareh and Morstatter, Fred and Saxena, Nripsuta and Lerman, Kristina and Galstyan, Aram},
  journal={ACM Computing Surveys (CSUR)},
  volume={54},
  number={6},
  pages={1--35},
  year={2021},
  publisher={ACM New York, NY, USA}
}

@article{blanco2020machine,
  title={Machine learning explainability via microaggregation and shallow decision trees},
  author={Blanco-Justicia, Alberto and Domingo-Ferrer, Josep and Mart{\'\i}nez, Sergio and S{\'a}nchez, David},
  journal={Knowledge-Based Systems},
  volume={194},
  pages={105532},
  year={2020},
  publisher={Elsevier}
}

@inproceedings{feldman2015certifying,
  title={Certifying and removing disparate impact},
  author={Feldman, Michael and Friedler, Sorelle A and Moeller, John and Scheidegger, Carlos and Venkatasubramanian, Suresh},
  booktitle={proceedings of the 21th ACM SIGKDD international conference on knowledge discovery and data mining},
  pages={259--268},
  year={2015}
}

@article{chen2019looks,
  title={This looks like that: deep learning for interpretable image recognition},
  author={Chen, Chaofan and Li, Oscar and Tao, Daniel and Barnett, Alina and Rudin, Cynthia and Su, Jonathan K},
  journal={Advances in neural information processing systems},
  volume={32},
  year={2019}
}

@article{wiegreffe2021teach,
  title={Teach Me to Explain: A Review of Datasets for Explainable NLP},
  author={Wiegreffe, Sarah and Marasovi{\'c}, Ana},
  journal={arXiv:2102.12060},
  year={2021}
}

@inproceedings{zhou2016learning,
  title={Learning deep features for discriminative localization},
  author={Zhou, Bolei and Khosla, Aditya and Lapedriza, Agata and Oliva, Aude and Torralba, Antonio},
  booktitle={Proceedings of the IEEE conference on computer vision and pattern recognition},
  pages={2921--2929},
  year={2016}
}

@article{bahdanau2014neural,
  title={Neural machine translation by jointly learning to align and translate},
  author={Bahdanau, Dzmitry and Cho, Kyunghyun and Bengio, Yoshua},
  journal={arXiv:1409.0473},
  year={2014}
}

@article{craven1995extracting,
  title={Extracting tree-structured representations of trained networks},
  author={Craven, Mark and Shavlik, Jude},
  journal={Advances in neural information processing systems},
  volume={8},
  pages={24--30},
  year={1995}
}

@article{zhou2016interpreting,
  title={Interpreting models via single tree approximation},
  author={Zhou, Yichen and Hooker, Giles},
  journal={arXiv:1610.09036},
  year={2016}
}

@inproceedings{tan2020tree,
  title={Tree space prototypes: Another look at making tree ensembles interpretable},
  author={Tan, Sarah and Soloviev, Matvey and Hooker, Giles and Wells, Martin T},
  booktitle={Proceedings of the 2020 ACM-IMS on Foundations of Data Science Conference},
  pages={23--34},
  year={2020}
}

@article{zhou2003extracting,
  title={Extracting symbolic rules from trained neural network ensembles},
  author={Zhou, Zhi-Hua and Jiang, Yuan and Chen, Shi-Fu},
  journal={Ai Communications},
  volume={16},
  number={1},
  pages={3--15},
  year={2003},
  publisher={IOS Press}
}

@article{augasta2012reverse,
  title={Reverse engineering the neural networks for rule extraction in classification problems},
  author={Augasta, M Gethsiyal and Kathirvalavakumar, Thangairulappan},
  journal={Neural processing letters},
  volume={35},
  number={2},
  pages={131--150},
  year={2012},
  publisher={Springer}
}

@inproceedings{xu2015show,
  title={Show, attend and tell: Neural image caption generation with visual attention},
  author={Xu, Kelvin and Ba, Jimmy and Kiros, Ryan and Cho, Kyunghyun and Courville, Aaron and Salakhudinov, Ruslan and Zemel, Rich and Bengio, Yoshua},
  booktitle={International conference on machine learning},
  pages={2048--2057},
  year={2015},
  organization={PMLR}
}

@inproceedings{zeiler2014visualizing,
  title={Visualizing and understanding convolutional networks},
  author={Zeiler, Matthew D and Fergus, Rob},
  booktitle={European conference on computer vision},
  pages={818--833},
  year={2014},
  organization={Springer}
}

@inproceedings{selvaraju2017grad,
  title={Grad-cam: Visual explanations from deep networks via gradient-based localization},
  author={Selvaraju, Ramprasaath R and Cogswell, Michael and Das, Abhishek and Vedantam, Ramakrishna and Parikh, Devi and Batra, Dhruv},
  booktitle={Proceedings of the IEEE international conference on computer vision},
  pages={618--626},
  year={2017}
}

@book{molnar2020interpretable,
  title={Interpretable machine learning},
  author={Molnar, Christoph},
  year={2020},
  publisher={Lulu. com}
}

@article{angelov2020towards,
  title={Towards explainable deep neural networks (xDNN)},
  author={Angelov, Plamen and Soares, Eduardo},
  journal={Neural Networks},
  volume={130},
  pages={185--194},
  year={2020},
  publisher={Elsevier}
}

@inproceedings{koh2017understanding,
  title={Understanding black-box predictions via influence functions},
  author={Koh, Pang Wei and Liang, Percy},
  booktitle={International Conference on Machine Learning},
  pages={1885--1894},
  year={2017},
  organization={PMLR}
}

@inproceedings{tu2020select,
  title={Select, answer and explain: Interpretable multi-hop reading comprehension over multiple documents},
  author={Tu, Ming and Huang, Kevin and Wang, Guangtao and Huang, Jing and He, Xiaodong and Zhou, Bowen},
  booktitle={Proceedings of the AAAI Conference on Artificial Intelligence},
  volume={34},
  number={05},
  pages={9073--9080},
  year={2020}
}

@inproceedings{yang2018hotpotqa,
  title={HotpotQA: A Dataset for Diverse, Explainable Multi-hop Question Answering},
  author={Yang, Zhilin and Qi, Peng and Zhang, Saizheng and Bengio, Yoshua and Cohen, William and Salakhutdinov, Ruslan and Manning, Christopher D},
  booktitle={Proceedings of the 2018 Conference on Empirical Methods in Natural Language Processing},
  pages={2369--2380},
  year={2018}
}

@article{adadi2018peeking,
  title={Peeking inside the black-box: a survey on explainable artificial intelligence (XAI)},
  author={Adadi, Amina and Berrada, Mohammed},
  journal={IEEE access},
  volume={6},
  pages={52138--52160},
  year={2018},
  publisher={IEEE}
}

@inproceedings{zhang2018interpretable,
  title={Interpretable convolutional neural networks},
  author={Zhang, Quanshi and Wu, Ying Nian and Zhu, Song-Chun},
  booktitle={Proceedings of the IEEE Conference on Computer Vision and Pattern Recognition},
  pages={8827--8836},
  year={2018}
}

@inproceedings{zhang2019interpreting,
  title={Interpreting cnns via decision trees},
  author={Zhang, Quanshi and Yang, Yu and Ma, Haotian and Wu, Ying Nian},
  booktitle={Proceedings of the IEEE/CVF Conference on Computer Vision and Pattern Recognition},
  pages={6261--6270},
  year={2019}
}

@article{frosst2017distilling,
  title={Distilling a neural network into a soft decision tree},
  author={Frosst, Nicholas and Hinton, Geoffrey},
  journal={arXiv:1711.09784},
  year={2017}
}

@article{wachter2017counterfactual,
  title={Counterfactual explanations without opening the black box: Automated decisions and the GDPR},
  author={Wachter, Sandra and Mittelstadt, Brent and Russell, Chris},
  journal={Harv. JL \& Tech.},
  volume={31},
  pages={841},
  year={2017},
  publisher={HeinOnline}
}

@inproceedings{liu2019stgan,
  title={STGAN: A unified selective transfer network for arbitrary image attribute editing},
  author={Liu, Ming and Ding, Yukang and Xia, Min and Liu, Xiao and Ding, Errui and Zuo, Wangmeng and Wen, Shilei},
  booktitle={Proceedings of the IEEE/CVF Conference on Computer Vision and Pattern Recognition},
  pages={3673--3682},
  year={2019}
}

@inproceedings{ghorbani2019interpretation,
  title={Interpretation of neural networks is fragile},
  author={Ghorbani, Amirata and Abid, Abubakar and Zou, James},
  booktitle={Proceedings of the AAAI Conference on Artificial Intelligence},
  volume={33},
  number={01},
  pages={3681--3688},
  year={2019}
}

@article{lakkaraju2017interpretable,
  title={Interpretable \& explorable approximations of black box models},
  author={Lakkaraju, Himabindu and Kamar, Ece and Caruana, Rich and Leskovec, Jure},
  journal={arXiv:1707.01154},
  year={2017}
}

@inproceedings{poursabzi2021manipulating,
  title={Manipulating and measuring model interpretability},
  author={Poursabzi-Sangdeh, Forough and Goldstein, Daniel G and Hofman, Jake M and Wortman Vaughan, Jennifer Wortman and Wallach, Hanna},
  booktitle={Proceedings of the 2021 CHI conference on human factors in computing systems},
  pages={1--52},
  year={2021}
}

@inproceedings{gundersen2018state,
  title={State of the art: Reproducibility in artificial intelligence},
  author={Gundersen, Odd Erik and Kjensmo, Sigbjorn},
  booktitle={Proceedings of the AAAI Conference on Artificial Intelligence},
  volume={32},
  number={1},
  year={2018}
}

@inproceedings{breuer2020measure,
  title={How to Measure the Reproducibility of System-oriented IR Experiments},
  author={Breuer, Timo and Ferro, Nicola and Fuhr, Norbert and Maistro, Maria and Sakai, Tetsuya and Schaer, Philipp and Soboroff, Ian},
  booktitle={Proceedings of the 43rd International ACM SIGIR Conference on Research and Development in Information Retrieval},
  pages={349--358},
  year={2020}
}

@inproceedings{ferro2018overview,
  title={Overview of CENTRE@ CLEF 2018: a first tale in the systematic reproducibility realm},
  author={Ferro, Nicola and Maistro, Maria and Sakai, Tetsuya and Soboroff, Ian},
  booktitle={International Conference of the Cross-Language Evaluation Forum for European Languages},
  pages={239--246},
  year={2018},
  organization={Springer}
}

@article{jobin2019global,
  title={The global landscape of AI ethics guidelines},
  author={Jobin, Anna and Ienca, Marcello and Vayena, Effy},
  journal={Nature Machine Intelligence},
  volume={1},
  number={9},
  pages={389--399},
  year={2019},
  publisher={Nature Publishing Group}
}

@book{sai2020audting,
  title={Auditing machine learning algorithms},
  author={{Supreme Audit Institutions of Finland, Germany, the Netherlands, Norway and the UK}},
  year={2020},
  publisher={\url{https://www.auditingalgorithms.net/index.html}}
}

@article{arnold2019factsheets,
  title={FactSheets: Increasing trust in AI services through supplier's declarations of conformity},
  author={Arnold, Matthew and Bellamy, Rachel KE and Hind, Michael and Houde, Stephanie and Mehta, Sameep and Mojsilovi{\'c}, Aleksandra and Nair, Ravi and Ramamurthy, K Natesan and Olteanu, Alexandra and Piorkowski, David and others},
  journal={IBM Journal of Research and Development},
  volume={63},
  number={4/5},
  pages={6--1},
  year={2019},
  publisher={IBM}
}

@misc{xinhua2020savelife,
  title={Precisely Detecting Desperate Emotions, This {AI} Customer Service Saved A Life by Alerting Help},
  author={{XinhuaNet}},
  howpublished={\url{http://www.xinhuanet.com/tech/2020-06/24/c_1126152953.htm}},
  note={Accessed: January 19, 2021 (in Chinese)},
  year={2020}
}

@misc{google2018reducing,
  title={Reducing gender bias in Google Translate},
  author={James Kuczmarski},
  howpublished={\url{https://blog.google/products/translate/reducing-gender-bias-google-translate/}},
  note={Accessed: April 28, 2021},
  year={2018}
}

@article{sculley2015hidden,
  title={Hidden technical debt in machine learning systems},
  author={Sculley, David and Holt, Gary and Golovin, Daniel and Davydov, Eugene and Phillips, Todd and Ebner, Dietmar and Chaudhary, Vinay and Young, Michael and Crespo, Jean-Francois and Dennison, Dan},
  journal={Advances in neural information processing systems},
  volume={28},
  pages={2503--2511},
  year={2015}
}

@article{huang2020survey,
  title={A survey of safety and trustworthiness of deep neural networks: Verification, testing, adversarial attack and defence, and interpretability},
  author={Huang, Xiaowei and Kroening, Daniel and Ruan, Wenjie and Sharp, James and Sun, Youcheng and Thamo, Emese and Wu, Min and Yi, Xinping},
  journal={Computer Science Review},
  volume={37},
  pages={100270},
  year={2020},
  publisher={Elsevier}
}

@inproceedings{ma2018deepgauge,
  title={Deepgauge: Multi-granularity testing criteria for deep learning systems},
  author={Ma, Lei and Juefei-Xu, Felix and Zhang, Fuyuan and Sun, Jiyuan and Xue, Minhui and Li, Bo and Chen, Chunyang and Su, Ting and Li, Li and Liu, Yang and others},
  booktitle={Proceedings of the 33rd ACM/IEEE International Conference on Automated Software Engineering},
  pages={120--131},
  year={2018}
}

@inproceedings{ribeiro2020beyond,
  title={Beyond Accuracy: Behavioral Testing of NLP Models with CheckList},
  author={Ribeiro, Marco Tulio and Wu, Tongshuang and Guestrin, Carlos and Singh, Sameer},
  booktitle={Proceedings of the 58th Annual Meeting of the Association for Computational Linguistics},
  pages={4902--4912},
  year={2020}
}

@article{rosique2019systematic,
  title={A systematic review of perception system and simulators for autonomous vehicles research},
  author={Rosique, Francisca and Navarro, Pedro J and Fern{\'a}ndez, Carlos and Padilla, Antonio},
  journal={Sensors},
  volume={19},
  number={3},
  pages={648},
  year={2019},
  publisher={Multidisciplinary Digital Publishing Institute}
}

@inproceedings{chen2017multi,
  title={Multi-view 3d object detection network for autonomous driving},
  author={Chen, Xiaozhi and Ma, Huimin and Wan, Ji and Li, Bo and Xia, Tian},
  booktitle={Proceedings of the IEEE conference on Computer Vision and Pattern Recognition},
  pages={1907--1915},
  year={2017}
}

@article{likmeta2020combining,
  title={Combining reinforcement learning with rule-based controllers for transparent and general decision-making in autonomous driving},
  author={Likmeta, Amarildo and Metelli, Alberto Maria and Tirinzoni, Andrea and Giol, Riccardo and Restelli, Marcello and Romano, Danilo},
  journal={Robotics and Autonomous Systems},
  volume={131},
  pages={103568},
  year={2020},
  publisher={Elsevier}
}

@inproceedings{horwick2010strategy,
  title={Strategy and architecture of a safety concept for fully automatic and autonomous driving assistance systems},
  author={H{\"o}rwick, Markus and Siedersberger, Karl-Heinz},
  booktitle={2010 IEEE Intelligent Vehicles Symposium},
  pages={955--960},
  year={2010},
  organization={IEEE}
}

@misc{wood2019safety,
  author={Wood, Matthew and others},
  title = {Safety first for autonomous driving},
  howpublished = {\url{https://www.daimler.com/documents/innovation/other/safety-first-for-automated-driving.pdf}},
  year={2019},
  note = {Accessed: 2021-02-20}
}

@inproceedings{buolamwini2018gender,
  title={Gender shades: Intersectional accuracy disparities in commercial gender classification},
  author={Buolamwini, Joy and Gebru, Timnit},
  booktitle={Conference on fairness, accountability and transparency},
  pages={77--91},
  year={2018},
  organization={PMLR}
}

@article{de2018algorithmic,
  title={Algorithmic decision-making based on machine learning from Big Data: Can transparency restore accountability?},
  author={De Laat, Paul B},
  journal={Philosophy \& technology},
  volume={31},
  number={4},
  pages={525--541},
  year={2018},
  publisher={Springer}
}

@article{piano2020ethical,
  title={Ethical principles in machine learning and artificial intelligence: cases from the field and possible ways forward},
  author={Piano, Samuele Lo},
  journal={Humanities and Social Sciences Communications},
  volume={7},
  number={1},
  pages={1--7},
  year={2020},
  publisher={Palgrave}
}

@article{leslie2019understanding,
  title={Understanding artificial intelligence ethics and safety: A guide for the responsible design and implementation of AI systems in the public sector},
  author={Leslie, David},
  journal={Available at SSRN 3403301},
  year={2019}
}

@article{lavin2021technology,
  title={Technology readiness levels for machine learning systems},
  author={Lavin, Alexander and Gilligan-Lee, Ciar{\`a}n M and Visnjic, Alessya and Ganju, Siddha and Newman, Dava and Ganguly, Sujoy and Lange, Danny and Baydin, At{\i}l{\i}m G{\"u}ne{\c{s}} and Sharma, Amit and Gibson, Adam and others},
  journal={arXiv:2101.03989},
  year={2021}
}

@article{sandvig2014auditing,
  title={Auditing algorithms: Research methods for detecting discrimination on internet platforms},
  author={Sandvig, Christian and Hamilton, Kevin and Karahalios, Karrie and Langbort, Cedric},
  journal={Data and discrimination: converting critical concerns into productive inquiry},
  volume={22},
  pages={4349--4357},
  year={2014}
}

@inproceedings{wilson2021building,
  title={Building and auditing fair algorithms: A case study in candidate screening},
  author={Wilson, Christo and Ghosh, Avijit and Jiang, Shan and Mislove, Alan and Baker, Lewis and Szary, Janelle and Trindel, Kelly and Polli, Frida},
  booktitle={Proceedings of the 2021 ACM Conference on Fairness, Accountability, and Transparency},
  pages={666--677},
  year={2021}
}

@article{reisman2018algorithmic,

  title={Algorithmic impact assessments: A practical framework for public agency accountability},
  author={Reisman, Dillon and Schultz, Jason and Crawford, Kate and Whittaker, Meredith},
  journal={AI Now Institute},
  pages={1--22},
  year={2018}
}

@article{bender2018data,
  title={Data statements for natural language processing: Toward mitigating system bias and enabling better science},
  author={Bender, Emily M and Friedman, Batya},
  journal={Transactions of the Association for Computational Linguistics},
  volume={6},
  pages={587--604},
  year={2018},
  publisher={MIT Press}
}

@article{holland2018dataset,
  title={The dataset nutrition label: A framework to drive higher data quality standards},
  author={Holland, Sarah and Hosny, Ahmed and Newman, Sarah and Joseph, Joshua and Chmielinski, Kasia},
  journal={arXiv:1805.03677},
  year={2018}
}

@article{tay2014stereotypes,
  title={When stereotypes meet robots: the double-edge sword of robot gender and personality in human--robot interaction},
  author={Tay, Benedict and Jung, Younbo and Park, Taezoon},
  journal={Computers in Human Behavior},
  volume={38},
  pages={75--84},
  year={2014},
  publisher={Elsevier}
}

@misc{unesco2019blush,
  title={I'd blush if {I} could: closing gender divides in digital skills through education},
  author={{UNESCO} and {EQUALS Skills Coalition}},
  howpublished={\url{https://unesdoc.unesco.org/ark:/48223/pf0000367416.page=1}},
  note={Accessed: January 21, 2021},
  year={2019},
}

@misc{elgan2018kidsalexa,
  title={The case against teaching kids to be polite to {A}lexa},
  author={Mike Elgan},
  howpublished={\url{https://www.fastcompany.com/40588020/the-case-against-teaching-kids-to-be-polite-to-alexa}},
  year={2018},
  note={Accessed: January 22, 2021}
}

@misc{rudgard2018alexachildren,
  title={`Alexa generation' may be learning bad manners from talking to digital assistants, report warns},
  author={Olivia Rudgard},
  howpublished={\url{https://www.telegraph.co.uk/news/2018/01/31/alexa-generation-could-learning-bad-manners-talking-digital/}},
  note={Accessed: January 10, 2021},
  year={2018}
}

@article{aimar1998introduction,
  title={Introduction to software documentation},
  author={Aimar, Alberto},
  year={1998},
  publisher={CERN}
}

@inproceedings{zielke2020artificial,
  title={Is Artificial Intelligence Ready for Standardization?},
  author={Zielke, Thomas},
  booktitle={European Conference on Software Process Improvement},
  pages={259--274},
  year={2020},
  organization={Springer}
}

@article{schiff2021ai,
  title={AI Ethics in the Public, Private, and NGO Sectors: A Review of a Global Document Collection},
  author={Schiff, Daniel and Borenstein, Jason and Biddle, Justin and Laas, Kelly},
  journal={IEEE Transactions on Technology and Society},
  year={2021},
  publisher={IEEE}
}

@article{hagendorff2020ethics,
  title={The ethics of AI ethics: An evaluation of guidelines},
  author={Hagendorff, Thilo},
  journal={Minds and Machines},
  volume={30},
  number={1},
  pages={99--120},
  year={2020},
  publisher={Springer}
}

@article{lewis2021ontology,
  title={An Ontology for Standardising Trustworthy AI},
  author={Lewis, Dave and Filip, David and Pandit, Harshvardhan J},
  year={2021},
  publisher={IntechOpen}
}

@misc{DavidDao2020,
  title={Awful {AI}},
  author={Dao, David},
  year={2020},
  howpublished = {\url{https://github.com/daviddao/awful-ai}}
}

@article{gundersen2018reproducible,
  title={On reproducible AI: Towards reproducible research, open science, and digital scholarship in AI publications},
  author={Gundersen, Odd Erik and Gil, Yolanda and Aha, David W},
  journal={AI magazine},
  volume={39},
  number={3},
  pages={56--68},
  year={2018}
}

@article{schelenz2020applying,
  title={Applying Transparency in Artificial Intelligence based Personalization Systems},
  author={Schelenz, Laura and Segal, Avi and Gal, Kobi},
  journal={arXiv e-prints},
  pages={arXiv--2004},
  year={2020}
}

@misc{devulder2021world,
  author={De Vulder, Kristof and De Rouck, Florian and Hachez, Gilles},
  title = {The world's first Artificial Intelligence Act: Europe's proposal to lead in human-centered {AI}},
  howpublished = {\url{https://www.technologyslegaledge.com/2021/04/the-worlds-first-artificial-intelligence-act-europes-proposal-to-lead-in-human-centered-ai}},
  year={2021},
  note = {Accessed: 2021-04-30}
}

@inproceedings{marijan2020software,
  title={Software Testing for Machine Learning},
  author={Marijan, Dusica and Gotlieb, Arnaud},
  booktitle={Proceedings of the AAAI Conference on Artificial Intelligence},
  volume={34},
  number={09},
  pages={13576--13582},
  year={2020}
}

@inproceedings{javadi2020monitoring,
  title={Monitoring Misuse for Accountable'Artificial Intelligence as a Service'},
  author={Javadi, Seyyed Ahmad and Cloete, Richard and Cobbe, Jennifer and Lee, Michelle Seng Ah and Singh, Jatinder},
  booktitle={Proceedings of the AAAI/ACM Conference on AI, Ethics, and Society},
  pages={300--306},
  year={2020}
}

@article{hu2021model,
  title={Model Complexity of Deep Learning: A Survey},
  author={Hu, Xia and Chu, Lingyang and Pei, Jian and Liu, Weiqing and Bian, Jiang},
  journal={arXiv:2103.05127},
  year={2021}
}

@article{metzen2017detecting,
  title={ON DETECTING ADVERSARIAL PERTURBATIONS},
  author={Metzen, Jan Hendrik and Genewein, Tim and Fischer, Volker and Bischoff, Bastian},
  journal={stat},
  volume={1050},
  pages={21},
  year={2017}
}

@article{abdallah2016fraud,
  title={Fraud detection system: A survey},
  author={Abdallah, Aisha and Maarof, Mohd Aizaini and Zainal, Anazida},
  journal={Journal of Network and Computer Applications},
  volume={68},
  pages={90--113},
  year={2016},
  publisher={Elsevier}
}

@article{rabanser2019failing,
  title={Failing loudly: An empirical study of methods for detecting dataset shift},
  author={Rabanser, Stephan and G{\"u}nnemann, Stephan and Lipton, Zachary},
  journal={Advances in Neural Information Processing Systems},
  volume={32},
  year={2019}
}

@misc{samuylova2020machine,
  author={Samuylova, Elena},
  title = {Machine Learning Monitoring: What It Is, and What We Are Missing},
  howpublished = {\url{https://towardsdatascience.com/machine-learning-monitoring-what-it-is-and-what-we-are-missing-e644268023ba}},
  year={2020},
  note = {Accessed: 2021-05-18}
}

@article{perez2010argos,
  title={Argos: An advanced in-vehicle data recorder on a massively sensorized vehicle for car driver behavior experimentation},
  author={P{\'e}rez, Antonio and Garc{\'\i}a, M Isabel and Nieto, Manuel and Pedraza, Jos{\'e} L and Rodr{\'\i}guez, Santiago and Zamorano, Juan},
  journal={IEEE Transactions on Intelligent Transportation Systems},
  volume={11},
  number={2},
  pages={463--473},
  year={2010},
  publisher={IEEE}
}

@article{yao2020smart,
  title={The smart black box: A value-driven high-bandwidth automotive event data recorder},
  author={Yao, Yu and Atkins, Ella},
  journal={IEEE Transactions on Intelligent Transportation Systems},
  year={2020},
  publisher={IEEE}
}

@inproceedings{friedler2019comparative,
  title={A comparative study of fairness-enhancing interventions in machine learning},
  author={Friedler, Sorelle A and Scheidegger, Carlos and Venkatasubramanian, Suresh and Choudhary, Sonam and Hamilton, Evan P and Roth, Derek},
  booktitle={Proceedings of the conference on fairness, accountability, and transparency},
  pages={329--338},
  year={2019}
}

@inproceedings{koenig2004design,
  title={Design and use paradigms for gazebo, an open-source multi-robot simulator},
  author={Koenig, Nathan and Howard, Andrew},
  booktitle={2004 IEEE/RSJ International Conference on Intelligent Robots and Systems (IROS)(IEEE Cat. No. 04CH37566)},
  volume={3},
  pages={2149--2154},
  organization={IEEE}
}

@inproceedings{todorov2012mujoco,
  title={Mujoco: A physics engine for model-based control},
  author={Todorov, Emanuel and Erez, Tom and Tassa, Yuval},
  booktitle={2012 IEEE/RSJ International Conference on Intelligent Robots and Systems},
  pages={5026--5033},
  year={2012},
  organization={IEEE}
}

@article{wymann2000torcs,
  title={Torcs, the open racing car simulator},
  author={Wymann, Bernhard and Espi{\'e}, Eric and Guionneau, Christophe and Dimitrakakis, Christos and Coulom, R{\'e}mi and Sumner, Andrew},
  journal={Software available at http://torcs. sourceforge. net},
  volume={4},
  number={6},
  pages={2},
  year={2000},
  publisher={Citeseer}
}

@article{benekohal1988carsim,
  title={CARSIM: Car-following model for simulation of traffic in normal and stop-and-go conditions},
  author={Benekohal, Rahim F and Treiterer, Joseph},
  journal={Transportation research record},
  volume={1194},
  pages={99--111},
  year={1988},
  publisher={SAGE Publishing}
}

@article{tideman2010scenario,
  title={Scenario-based simulation environment for assistance systems},
  author={Tideman, Martijn},
  journal={ATZautotechnology},
  volume={10},
  number={1},
  pages={28--32},
  year={2010},
  publisher={Springer}
}

@inproceedings{bokc2007validation,
 title={Validation of the vehicle in the loop (vil); a milestone for the simulation of driver assistance systems},
 author={Bokc, Thomas and Maurer, Markus and Farber, Georg},
 booktitle={2007 IEEE Intelligent vehicles symposium},
 pages={612--617},
 year={2007},
 organization={IEEE}
}

@article{ashmore2021assuring,
  title={Assuring the machine learning lifecycle: Desiderata, methods, and challenges},
  author={Ashmore, Rob and Calinescu, Radu and Paterson, Colin},
  journal={ACM Computing Surveys (CSUR)},
  volume={54},
  number={5},
  pages={1--39},
  year={2021},
  publisher={ACM New York, NY, USA}
}

@article{bellamy2018ai,
  title={AI Fairness 360: An Extensible Toolkit for Detecting},
  author={Bellamy, Rachel KE and Dey, Kuntal and Hind, Michael and Hoffman, Samuel C and Houde, Stephanie and Kannan, Kalapriya and Lohia, Pranay and Martino, Jacquelyn and Mehta, Sameep and Mojsilovic, Aleksandra and others},
  journal={Understanding, and Mitigating Unwanted Algorithmic Bias},
  year={2018}
}

@article{deyoung2020eraser,
  title={ERASER: A Benchmark to Evaluate Rationalized NLP Models},
  author={DeYoung, Jay and Jain, Sarthak and Rajani, Nazneen Fatema and Lehman, Eric and Xiong, Caiming and Socher, Richard and Wallace, Byron C},
  journal={Transactions of the Association for Computational Linguistics},
  year={2020}
}

@inproceedings{mathew2021hatexplain,
  title={HateXplain: A Benchmark Dataset for Explainable Hate Speech Detection},
  author={Mathew, Binny and Saha, Punyajoy and Yimam, Seid Muhie and Biemann, Chris and Goyal, Pawan and Mukherjee, Animesh},
  booktitle={Proceedings of the AAAI Conference on Artificial Intelligence},
  volume={35},
  number={17},
  pages={14867--14875},
  year={2021}
}

@article{mohseni2018human,
  title={A human-grounded evaluation benchmark for local explanations of machine learning},
  author={Mohseni, Sina and Block, Jeremy E and Ragan, Eric D},
  journal={arXiv:1801.05075},
  year={2018}
}

@inproceedings{lee2019convlab,
    title = "{C}onv{L}ab: Multi-Domain End-to-End Dialog System Platform",
    author = "Lee, Sungjin  and
      Zhu, Qi  and
      Takanobu, Ryuichi  and
      Zhang, Zheng  and
      Zhang, Yaoqin  and
      Li, Xiang  and
      Li, Jinchao  and
      Peng, Baolin  and
      Li, Xiujun  and
      Huang, Minlie  and
      Gao, Jianfeng",
    booktitle = "Proceedings of the 57th Annual Meeting of the Association for Computational Linguistics: System Demonstrations",
    month = jul,
    year = "2019",
    address = "Florence, Italy",
    publisher = "Association for Computational Linguistics",
    url = "https://www.aclweb.org/anthology/P19-3011",
    doi = "10.18653/v1/P19-3011",
    pages = "64--69",
}

@article{dafoe2020open,
  title={Open Problems in Cooperative AI},
  author={Dafoe, Allan and Hughes, Edward and Bachrach, Yoram and Collins, Tantum and McKee, Kevin R and Leibo, Joel Z and Larson, Kate and Graepel, Thore},
  journal={arXiv:2012.08630},
  year={2020}
}

@inproceedings{amershi2019guidelines,
  title={Guidelines for human-AI interaction},
  author={Amershi, Saleema and Weld, Dan and Vorvoreanu, Mihaela and Fourney, Adam and Nushi, Besmira and Collisson, Penny and Suh, Jina and Iqbal, Shamsi and Bennett, Paul N and Inkpen, Kori and others},
  booktitle={Proceedings of the 2019 chi conference on human factors in computing systems},
  pages={1--13},
  year={2019}
}

@inproceedings{stumpf2016explanations,
  title={Explanations considered harmful? user interactions with machine learning systems},
  author={Stumpf, Simone and Bussone, Adrian and O'sullivan, Dympna},
  booktitle={Proceedings of the ACM SIGCHI Conference on Human Factors in Computing Systems (CHI)},
  year={2016}
}

@article{weld2019challenge,
  title={The challenge of crafting intelligible intelligence},
  author={Weld, Daniel S and Bansal, Gagan},
  journal={Communications of the ACM},
  volume={62},
  number={6},
  pages={70--79},
  year={2019},
  publisher={ACM New York, NY, USA}
}

@article{shneiderman2020bridging,
  title={Bridging the gap between ethics and practice: Guidelines for reliable, safe, and trustworthy Human-Centered AI systems},
  author={Shneiderman, Ben},
  journal={ACM Transactions on Interactive Intelligent Systems (TiiS)},
  volume={10},
  number={4},
  pages={1--31},
  year={2020},
  publisher={ACM New York, NY, USA}
}

@article{schmidt2017intervention,
  title={Intervention user interfaces: a new interaction paradigm for automated systems},
  author={Schmidt, Albrecht and Herrmann, Thomas},
  journal={Interactions},
  volume={24},
  number={5},
  pages={40--45},
  year={2017},
  publisher={ACM New York, NY, USA}
}

@inproceedings{abdul2018trends,
  title={Trends and trajectories for explainable, accountable and intelligible systems: An hci research agenda},
  author={Abdul, Ashraf and Vermeulen, Jo and Wang, Danding and Lim, Brian Y and Kankanhalli, Mohan},
  booktitle={Proceedings of the 2018 CHI conference on human factors in computing systems},
  pages={1--18},
  year={2018}
}

@inproceedings{chatzimparmpas2020state,
  title={The state of the art in enhancing trust in machine learning models with the use of visualizations},
  author={Chatzimparmpas, Angelos and Martins, Rafael M and Jusufi, Ilir and Kucher, Kostiantyn and Rossi, Fabrice and Kerren, Andreas},
  booktitle={Computer Graphics Forum},
  volume={39},
  number={3},
  pages={713--756},
  year={2020},
  organization={Wiley Online Library}
}

@inproceedings{dodge2019explaining,
  title={Explaining models: an empirical study of how explanations impact fairness judgment},
  author={Dodge, Jonathan and Liao, Q Vera and Zhang, Yunfeng and Bellamy, Rachel KE and Dugan, Casey},
  booktitle={Proceedings of the 24th International Conference on Intelligent User Interfaces},
  pages={275--285},
  year={2019}
}

@article{kafle2020artificial,
  title={Artificial intelligence fairness in the context of accessibility research on intelligent systems for people who are deaf or hard of hearing},
  author={Kafle, Sushant and Glasser, Abraham and Al-khazraji, Sedeeq and Berke, Larwan and Seita, Matthew and Huenerfauth, Matt},
  journal={ACM SIGACCESS Accessibility and Computing},
  number={125},
  pages={1--1},
  year={2020},
  publisher={ACM New York, NY, USA}
}

@inproceedings{torres2018accessibility,
  title={Accessibility in Chatbots: The State of the Art in Favor of Users with Visual Impairment},
  author={Torres, Cec{\'\i}lia and Franklin, Walter and Martins, Laura},
  booktitle={International Conference on Applied Human Factors and Ergonomics},
  pages={623--635},
  year={2018},
  organization={Springer}
}

@article{holzinger2016interactive,
  title={Interactive machine learning for health informatics: when do we need the human-in-the-loop?},
  author={Holzinger, Andreas},
  journal={Brain Informatics},
  volume={3},
  number={2},
  pages={119--131},
  year={2016},
  publisher={Springer}
}

@article{singh2018decision,
  title={Decision provenance: Harnessing data flow for accountable systems},
  author={Singh, Jatinder and Cobbe, Jennifer and Norval, Chris},
  journal={IEEE Access},
  volume={7},
  pages={6562--6574},
  year={2018},
  publisher={IEEE}
}

@inproceedings{magdici2016fail,
  title={Fail-safe motion planning of autonomous vehicles},
  author={Magdici, Silvia and Althoff, Matthias},
  booktitle={2016 IEEE 19th International Conference on Intelligent Transportation Systems (ITSC)},
  pages={452--458},
  year={2016},
  organization={IEEE}
}

@article{fu2014monocular,
  title={Monocular visual-inertial SLAM-based collision avoidance strategy for fail-safe UAV using fuzzy logic controllers},
  author={Fu, Changhong and Olivares-Mendez, Miguel A and Suarez-Fernandez, Ramon and Campoy, Pascual},
  journal={Journal of Intelligent \& Robotic Systems},
  volume={73},
  number={1},
  pages={513--533},
  year={2014},
  publisher={Springer}
}

@phdthesis{muller2016increased,
  title={Increased autonomy for quadrocopter systems: trajectory generation, fail-safe strategies and state estimation},
  author={M{\"u}ller, Mark Wilfried},
  year={2016},
  school={ETH Zurich}
}

@article{vzliobaite2016overview,
  title={An overview of concept drift applications},
  author={{\v{Z}}liobait{\.e}, Indr{\.e} and Pechenizkiy, Mykola and Gama, Joao},
  journal={Big data analysis: new algorithms for a new society},
  pages={91--114},
  year={2016},
  publisher={Springer}
}

@inproceedings{hummer2019modelops,
  title={Modelops: Cloud-based lifecycle management for reliable and trusted ai},
  author={Hummer, Waldemar and Muthusamy, Vinod and Rausch, Thomas and Dube, Parijat and El Maghraoui, Kaoutar and Murthi, Anupama and Oum, Punleuk},
  booktitle={2019 IEEE International Conference on Cloud Engineering (IC2E)},
  pages={113--120},
  year={2019},
  organization={IEEE}
}

@article{siddique2020safetyops,
  title={SafetyOps},
  author={Siddique, Umair},
  journal={arXiv:2008.04461},
  year={2020}
}

@article{urban2021review,
  title={A Review of Formal Methods applied to Machine Learning},
  author={Urban, Caterina and Min{\'e}, Antoine},
  journal={arXiv:2104.02466},
  year={2021}
}

@article{zhang2018efficient,
  title={Efficient neural network robustness certification with general activation functions},
  author={Zhang, Huan and Weng, Tsui-Wei and Chen, Pin-Yu and Hsieh, Cho-Jui and Daniel, Luca},
  journal={Advances in neural information processing systems},
  volume={31},
  year={2018}
}

@inproceedings{weng2018evaluating,
  title={Evaluating the Robustness of Neural Networks: An Extreme Value Theory Approach},
  author={Weng, Tsui-Wei and Zhang, Huan and Chen, Pin-Yu and Yi, Jinfeng and Su, Dong and Gao, Yupeng and Hsieh, Cho-Jui and Daniel, Luca},
  booktitle={International Conference on Learning Representations},
  year={2018}
}

@article{clarke1996formal,
  title={Formal methods: State of the art and future directions},
  author={Clarke, Edmund M and Wing, Jeannette M},
  journal={ACM Computing Surveys (CSUR)},
  volume={28},
  number={4},
  pages={626--643},
  year={1996},
  publisher={ACM New York, NY, USA}
}

@inproceedings{boopathy2019cnn,
  title={Cnn-cert: An efficient framework for certifying robustness of convolutional neural networks},
  author={Boopathy, Akhilan and Weng, Tsui-Wei and Chen, Pin-Yu and Liu, Sijia and Daniel, Luca},
  booktitle={Proceedings of the AAAI Conference on Artificial Intelligence},
  volume={33},
  number={01},
  pages={3240--3247},
  year={2019}
}

@inproceedings{arpit2017closer,
  title={A closer look at memorization in deep networks},
  author={Arpit, Devansh and Jastrzebski, Stanislaw and Ballas, Nicolas and Krueger, David and Bengio, Emmanuel and Kanwal, Maxinder S and Maharaj, Tegan and Fischer, Asja and Courville, Aaron and Bengio, Yoshua and others},
  booktitle={International Conference on Machine Learning},
  pages={233--242},
  year={2017},
  organization={PMLR}
}

@article{krueger2017deep,
  title={Deep nets don't learn via memorization},
  author={Krueger, David and Ballas, Nicolas and Jastrzebski, Stanislaw and Arpit, Devansh and Kanwal, Maxinder S and Maharaj, Tegan and Bengio, Emmanuel and Fischer, Asja and Courville, Aaron},
  year={2017}
}

@article{kawaguchi2017generalization,
  title={Generalization in deep learning},
  author={Kawaguchi, Kenji and Kaelbling, Leslie Pack and Bengio, Yoshua},
  journal={arXiv:1710.05468},
  year={2017}
}

@article{xu2012robustness,
  title={Robustness and generalization},
  author={Xu, Huan and Mannor, Shie},
  journal={Machine learning},
  volume={86},
  number={3},
  pages={391--423},
  year={2012},
  publisher={Springer}
}

@article{tsipras103robustness,
  title={Robustness May Be at Odds with Accuracy},
  author={Tsipras, Dimitris and Santurkar, Shibani and Engstrom, Logan and Turner, Alexander and M{\k{a}}dry, Aleksander},
  journal={Training},
  volume={103},
  number={104},
  pages={105}
}

@article{raghunathan2019adversarial,
  title={Adversarial training can hurt generalization},
  author={Raghunathan, Aditi and Xie, Sang Michael and Yang, Fanny and Duchi, John C and Liang, Percy},
  journal={arXiv:1906.06032},
  year={2019}
}

@article{fawzi2018adversarial,
  title={Adversarial vulnerability for any classifier},
  author={Fawzi, Alhussein and Fawzi, Hamza and Fawzi, Omar},
  journal={Advances in neural information processing systems},
  volume={31},
  year={2018}
}

@article{turilli2009ethics,
  title={The ethics of information transparency},
  author={Turilli, Matteo and Floridi, Luciano},
  journal={Ethics and Information Technology},
  volume={11},
  number={2},
  pages={105--112},
  year={2009},
  publisher={Springer}
}

@article{piorkowski2020towards,
  title={Towards evaluating and eliciting high-quality documentation for intelligent systems},
  author={Piorkowski, David and Gonz{\'a}lez, Daniel and Richards, John and Houde, Stephanie},
  journal={arXiv:2011.08774},
  year={2020}
}

@article{brown2020language,
  title={Language models are few-shot learners},
  author={Brown, Tom and Mann, Benjamin and Ryder, Nick and Subbiah, Melanie and Kaplan, Jared D and Dhariwal, Prafulla and Neelakantan, Arvind and Shyam, Pranav and Sastry, Girish and Askell, Amanda and others},
  journal={Advances in neural information processing systems},
  volume={33},
  pages={1877--1901},
  year={2020}
}

@inproceedings{knowles2021sanction,
  title={The Sanction of Authority: Promoting Public Trust in AI},
  author={Knowles, Bran and Richards, John T},
  booktitle={Proceedings of the 2021 ACM Conference on Fairness, Accountability, and Transparency},
  pages={262--271},
  year={2021}
}

@article{raji2019ml,
  title={About ml: Annotation and benchmarking on understanding and transparency of machine learning lifecycles},
  author={Raji, Inioluwa Deborah and Yang, Jingying},
  journal={arXiv:1912.06166},
  year={2019}
}

@inproceedings{erhan2010does,
  title={Why does unsupervised pre-training help deep learning?},
  author={Erhan, Dumitru and Courville, Aaron and Bengio, Yoshua and Vincent, Pascal},
  booktitle={Proceedings of the thirteenth international conference on artificial intelligence and statistics},
  pages={201--208},
  year={2010},
  organization={JMLR Workshop and Conference Proceedings}
}

@inproceedings{long2015fully,
  title={Fully convolutional networks for semantic segmentation},
  author={Long, Jonathan and Shelhamer, Evan and Darrell, Trevor},
  booktitle={Proceedings of the IEEE conference on computer vision and pattern recognition},
  pages={3431--3440},
  year={2015}
}

@inproceedings{girshick2014rich,
  title={Rich feature hierarchies for accurate object detection and semantic segmentation},
  author={Girshick, Ross and Donahue, Jeff and Darrell, Trevor and Malik, Jitendra},
  booktitle={Proceedings of the IEEE conference on computer vision and pattern recognition},
  pages={580--587},
  year={2014}
}

@inproceedings{ioffe2015batch,
  title={Batch normalization: Accelerating deep network training by reducing internal covariate shift},
  author={Ioffe, Sergey and Szegedy, Christian},
  booktitle={International conference on machine learning},
  pages={448--456},
  year={2015},
  organization={PMLR}
}

@article{srivastava2014dropout,
  title={Dropout: a simple way to prevent neural networks from overfitting},
  author={Srivastava, Nitish and Hinton, Geoffrey and Krizhevsky, Alex and Sutskever, Ilya and Salakhutdinov, Ruslan},
  journal={The journal of machine learning research},
  volume={15},
  number={1},
  pages={1929--1958},
  year={2014},
  publisher={JMLR. org}
}

@inproceedings{krogh1992simple,
  title={A simple weight decay can improve generalization},
  author={Krogh, Anders and Hertz, John A},
  booktitle={Advances in neural information processing systems},
  pages={950--957},
  year={1992}
}

@article{yao2007early,
  title={On early stopping in gradient descent learning},
  author={Yao, Yuan and Rosasco, Lorenzo and Caponnetto, Andrea},
  journal={Constructive Approximation},
  volume={26},
  number={2},
  pages={289--315},
  year={2007},
  publisher={Springer}
}

@article{belkin2019reconciling,
  title={Reconciling modern machine-learning practice and the classical bias--variance trade-off},
  author={Belkin, Mikhail and Hsu, Daniel and Ma, Siyuan and Mandal, Soumik},
  journal={Proceedings of the National Academy of Sciences},
  volume={116},
  number={32},
  pages={15849--15854},
  year={2019},
  publisher={National Acad Sciences}
}

@book{friedman2001elements,
  title={The elements of statistical learning},
  author={Friedman, Jerome and Hastie, Trevor and Tibshirani, Robert and others},
  volume={1},
  number={10},
  year={2001},
  publisher={Springer series in statistics New York}
}

@book{vapnik2013nature,
  title={The nature of statistical learning theory},
  author={Vapnik, Vladimir},
  year={2013},
  publisher={Springer science \& business media}
}

@inproceedings{santoro2016meta,
  title={Meta-learning with memory-augmented neural networks},
  author={Santoro, Adam and Bartunov, Sergey and Botvinick, Matthew and Wierstra, Daan and Lillicrap, Timothy},
  booktitle={International conference on machine learning},
  pages={1842--1850},
  year={2016},
  organization={PMLR}
}

@inproceedings{finn2017model,
  title={Model-agnostic meta-learning for fast adaptation of deep networks},
  author={Finn, Chelsea and Abbeel, Pieter and Levine, Sergey},
  booktitle={International Conference on Machine Learning},
  pages={1126--1135},
  year={2017},
  organization={PMLR}
}

@article{li2016learning,
  title={Learning to optimize},
  author={Li, Ke and Malik, Jitendra},
  journal={arXiv:1606.01885},
  year={2016}
}

@article{snell2017prototypical,
  title={Prototypical networks for few-shot learning},
  author={Snell, Jake and Swersky, Kevin and Zemel, Richard},
  journal={Advances in neural information processing systems},
  volume={30},
  year={2017}
}

@inproceedings{koch2015siamese,
  title={Siamese neural networks for one-shot image recognition},
  author={Koch, Gregory and Zemel, Richard and Salakhutdinov, Ruslan},
  booktitle={ICML deep learning workshop},
  volume={2},
  year={2015},
  organization={Lille}
}

@article{vanschoren2018meta,
  title={Meta-learning: A survey},
  author={Vanschoren, Joaquin},
  journal={arXiv:1810.03548},
  year={2018}
}

@article{wang2020generalizing,
  title={Generalizing from a few examples: A survey on few-shot learning},
  author={Wang, Yaqing and Yao, Quanming and Kwok, James T and Ni, Lionel M},
  journal={ACM Computing Surveys (CSUR)},
  volume={53},
  number={3},
  pages={1--34},
  year={2020},
  publisher={ACM New York, NY, USA}
}

@inproceedings{su2019vl,
  title={VL-BERT: Pre-training of Generic Visual-Linguistic Representations},
  author={Su, Weijie and Zhu, Xizhou and Cao, Yue and Li, Bin and Lu, Lewei and Wei, Furu and Dai, Jifeng},
  booktitle={International Conference on Learning Representations},
  year={2019}
}

@article{pan2020auto,
  title={Auto-captions on GIF: A Large-scale Video-sentence Dataset for Vision-language Pre-training},
  author={Pan, Yingwei and Li, Yehao and Luo, Jianjie and Xu, Jun and Yao, Ting and Mei, Tao},
  journal={arXiv:2007.02375},
  year={2020}
}

@inproceedings{yao2021seco,
  title={Seco: Exploring sequence supervision for unsupervised representation learning},
  author={Yao, Ting and Zhang, Yiheng and Qiu, Zhaofan and Pan, Yingwei and Mei, Tao},
  booktitle={AAAI},
  volume={2},
  pages={7},
  year={2021}
}

@inproceedings{he2020momentum,
  title={Momentum contrast for unsupervised visual representation learning},
  author={He, Kaiming and Fan, Haoqi and Wu, Yuxin and Xie, Saining and Girshick, Ross},
  booktitle={Proceedings of the IEEE/CVF Conference on Computer Vision and Pattern Recognition},
  pages={9729--9738},
  year={2020}
}

@article{radford2018improving,
  title={Improving language understanding by generative pre-training},
  author={Radford, Alec and Narasimhan, Karthik and Salimans, Tim and Sutskever, Ilya},
  year={2018}
}

@inproceedings{devlin2019bert,
  title={{BERT}: Pre-training of Deep Bidirectional Transformers for Language Understanding},
  author={Devlin, Jacob and  Chang, Ming-Wei Chang and Kenton, Lee and Toutanova, Kristina},
  booktitle={Proceedings of the 2019 Conference of the North American Chapter of the Association for Computational Linguistics: Human Language Technologies, Volume 1 (Long and Short Papers)},
  year={2019}
}

@article{bartlett2002rademacher,
  title={Rademacher and Gaussian complexities: Risk bounds and structural results},
  author={Bartlett, Peter L and Mendelson, Shahar},
  journal={Journal of Machine Learning Research},
  volume={3},
  number={Nov},
  pages={463--482},
  year={2002}
}

@article{zhou2021domain,
  title={Domain generalization: A survey},
  author={Zhou, Kaiyang and Liu, Ziwei and Qiao, Yu and Xiang, Tao and Loy, Chen Change},
  journal={arXiv:2103.02503},
  year={2021}
}

@article{diakopoulos2016accountability,
  title={Accountability in algorithmic decision making},
  author={Diakopoulos, Nicholas},
  journal={Communications of the ACM},
  volume={59},
  number={2},
  pages={56--62},
  year={2016},
  publisher={ACM New York, NY, USA}
}

\makeatletter\@input{mainaux.tex}\makeatother
\makeatletter\@input{supaux.tex}\makeatother

\end{document}


\maketitle

\section{An Experimental Case Study of Trustworthy AI: Face Verification}
\begin{figure}
    
    \centering
    \includegraphics[width=\linewidth]{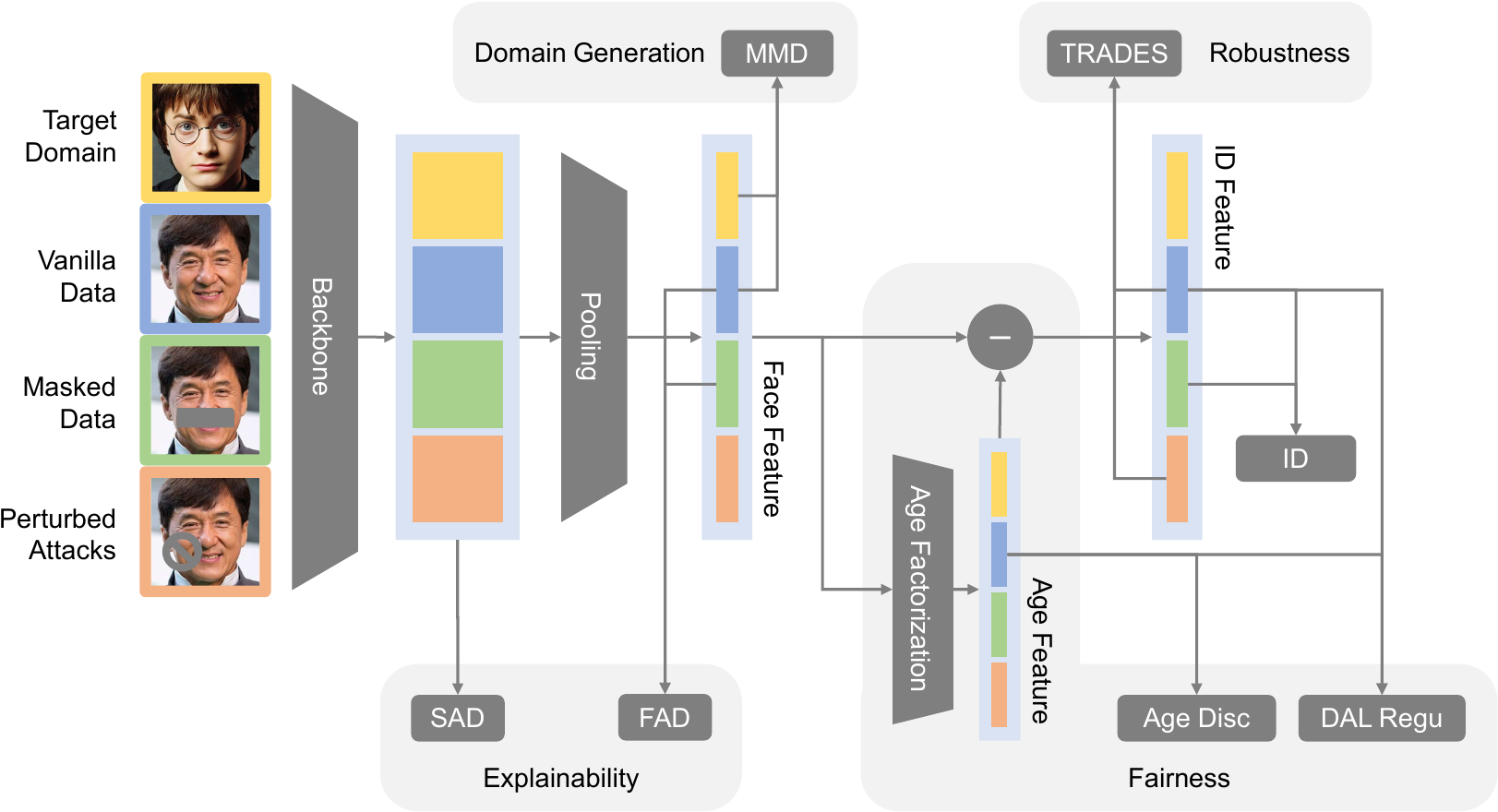}
    \caption{Network structure for a face verification model which integrates four trustworthiness approaches. See Section~\ref{ssec:integration} for details.}
    \label{fig:network}
\end{figure}

\subsection{Scenario}

We consider a scenario of face verification, which is applicable to, for instance, unlocking a smartphone. In this scenario, we examine the interactive effects between four aspects of AI trustworthiness -- robustness, explainability, fairness, and generalization to novel domains. Specifically, to prevent adversarial attacks, we utilize adversarial learning algorithms to enhance model robustness. We assume the age of the user group ranges from young to senior. To prevent algorithmic bias, we demand the recognition accuracy to be fair to different age groups by employing two different algorithmic improvements, we consider two types of algorithmic improvements. The first type is to exploit age invariant face recognition algorithms to enhance fairness across ages. The second type is to adopt transfer learning as a domain generalization approach to improve accuracy by extra unlabeled data from the disadvantageous group (which is younger faces for the conventional models we experimented with). Last but not least, to enhance model explainability, we explore replacing the conventional backbone with models with more explainable design.

\subsection{Datasets}

We use three public datasets to simulate our application scenario. We use ECCV FairFace Challenge Dataset (FFC)~\cite{sixta2020fairface} as our main training and testing dataset. FFC contains identity labels and age labels so that we can divide the test dataset by young (age $<30$) and sensor (age $\geq 30$) groups to examine the algorithmic fairness. In the domain generalization procedure, we aim to exploit unlabeled data to enhance model accuracy for the young group. We construct an unlabeled youngth face dataset from FairFace Dataset (FFD)~\cite{karkkainen2019fairface} by filtering the age labels. To further enhance the model performance, we use MS1M Dataset~\cite{guo2016ms} to pre-train our backbone.

\subsection{Baseline Model and Training Strategy}
Our experiments start from a baseline face recognition model with a backbone feature extractor of ResNet50 and a FC layer to classify identities. The backbone extractor is pre-trained on MS1M. The training loss is constructed following ArcFace~\cite{deng2019arcface}. The full model is firstly fine-tuned at its FC layers while freezing the backbone and then fine-tuned on all its weights.


\subsection{Submodule Approaches}

We consider four AI trustworthiness dimensions in our experiments and select the following algorithms to improve the model in each dimension:

\begin{itemize}
    \item \textbf{Robustness.} We consider black-box attacks using the M-PGD~\cite{kurakin2016adversarial} method. We assume attackers have access to the baseline model to generate adversarial samples. We consider three approaches in defense, i.e. image pre-processing by blurring, adversarial training with M-PGD~\cite{kurakin2016adversarial} and TRADES~\cite{zhang2019theoretically}.
    
    \item \textbf{Explainability.} We consider two face recognition approaches that enhance explainability. XCos~\cite{lin2020xcos} constructs an attention map to illustrate the spatial similarity between a verified pair. IFR~\cite{yin2019towards} added extra loss to enforce the model to learn feature maps that are interpretable to humans.
    
    \item \textbf{Fairness.} We consider three face recognition approaches to enhance fairness. 
    DAL~\cite{wang2019decorrelated} decomposes the face feature into an identity-related component and an age-related component, so as to reduce the unfairness caused by aging.
    MTLFace~\cite{huang2021age} introduces attention mechanism to decompose deep face features, and then uses multi-task training and continuous domain adaption for model learning. CIFP~\cite{xu2021consistent} applies a false positive rate penalty loss to mitigate bias in face recognition.
    
    \item \textbf{Domain Generalization.} We consider three transfer learning methods for domain generalization. In our experiments, the target domain (young group) is unlabeled, which requires unsupervised adaptation of knowledge from the source domain (senior group). Both MMD~\cite{long2015learning} and DeepCoral~\cite{sun2016deep} decrease the distance of the feature distribution between the source domain and the target domain by a metric function. DANN~\cite{ganin2015unsupervised} distinguishes the feature of the source domain and the target domain with a discriminator.
    
\end{itemize}

\subsection{Evaluation Metrics}

We consider the following metrics to evaluate the system behavior at different aspects.
\begin{itemize}
    \item \textbf{Performance.} We use ROC (receiver operating characteristic curve) and AUC (area under the ROC curve) to evaluate the model accuracy at different discrimination thresholds. The ROC are reported by TPR (true positve rate) / FPR (false positive rate) samples at FPR of $10^{-5}, 10^{-4}$, and $10^{-3}$.
    
    \item \textbf{Explainability.} It is difficult to quantitatively evaluate the explainability. For IFR, we use the average response location spreadness~(ARLS)~\cite{yin2019towards} as a surrogate measurement for the goodness of explainability. Larger ARLS implies that a feature response denotes better spatial consistency over faces, and thus is more semantically interpretable to humans.
    \item \textbf{Group Performance.} We divide the testing dataset into groups of young and seniors, and evaluate the TPR/FPR for each group respectively to illustrate the intra-group performance and fairness. 
    \item \textbf{Fairness.} We use the bias metric proposed in~\cite{sixta2020fairface} to further measure the fairness performance. Lower value implies lower bias and better fairness.
    \item \textbf{Robustness.} We use the attack success rate (ASR) to measure the robustness of a given model. The attack success rate is evaluated at the same FPR of $10^{-5}, 10^{-4}$, and $10^{-3}$ as for model accuracy performance.
\end{itemize}

We modify the above selected approaches to have the same backbone as the baseline model. In Table~\ref{tab:submodule}, we compare these approaches with baseline using the metrics aforementioned. 


\subsection{Performance Evaluation on Single Approaches}

\begin{table}
    \centering
    \begin{subtable}[h]{0.45\linewidth}
    \centering
    \includegraphics[scale=0.8]{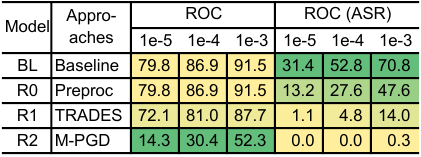}
    \caption{Robustness Algorithms Trade-off}
    \label{fig:rob-perf}
    \end{subtable}
    \begin{subtable}[h]{0.45\linewidth}
    \centering
    \includegraphics[scale=0.8]{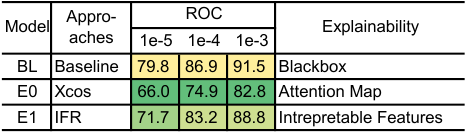}
    \caption{Explainability Algorithms Trade-off}
    \label{fig:expl-perf}
    \end{subtable}\\
    \begin{subtable}[h]{0.6\linewidth}
    \includegraphics[scale=0.8]{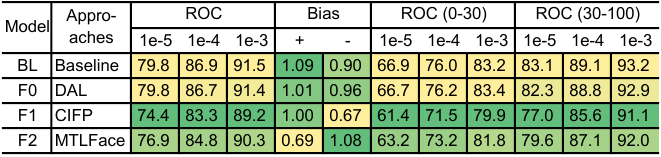}
    \caption{Fairness Algorithms Trade-off}
    \label{fig:fair-perf}
    \end{subtable}\\
    \begin{subtable}[h]{0.6\linewidth}
    \includegraphics[scale=0.8]{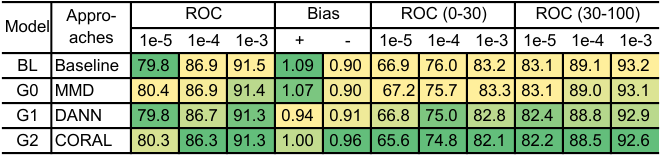}
    \caption{Domain Generalization Algorithms Trade-off}
    \label{fig:gen-perf}
    \end{subtable}
    \caption{Performance evaluation on each single submodule approaches.}
    \label{tab:submodule}
\end{table}

We first evaluate each selected approach respectively and illustrate its accuracy-trustworthiness trade-off with the baseline approach. 

In Table~\ref{fig:rob-perf}, we compare the performance of the selected robustness algorithms. As can be noticed, pre-processing approach preserves the accuracy of the baseline model but is not sufficiently robust to the attacks. TRADES and M-PGD both retain good robustness and TRADES has a less accuracy drop. 

In Table~\ref{fig:expl-perf}, we compare the performance of two selected explainable face recognition approaches. Accuracy drop can be noticed in both approaches. The two approaches utilize different explanation mechanisms. XCos learns a similarity function with an attention map which highlights similar parts. IFR enforces the model to produce interpretable feature maps. We did not compare the superiority between the two explanation mechanisms here since it is related to subjective judgment. The aforementioned ARLS is not used in this comparison, because it is only applicable for IFR.

In Table~\ref{fig:fair-perf}, we compare the performance of the selected fairness algorithms. To illustrate the change of system behavior, we consider two metrics. The first is the TPR/FPR accuracy on two groups of ages. As is shown, accuracy on the young group is significantly lower than the overall accuracy. The second is the bias measurement proposed in~\cite{sixta2020fairface}. As can be noticed the trade-off between fairness and accuracy performance differs across different approaches. Some fairness algorithms sacrifice the overall performance. Among the selected approaches, we observe that only DAL achieves very minor improvement on accuracy the disadvantageous young group. 

In Table~\ref{fig:gen-perf}, we compare three transfer learning approaches in the task of improving model performance on the young group. We use the same metrics with the fairness algorithms as in~Table~\ref{fig:fair-perf}, i.e. accuracy on two sub-groups and bias measurement. Though domain adaption using MMD on unlabeled data of the young group slightly improves the performance on both the young group, we find this improvement not significant. Other experiments even degrade the performance on all groups. We attribute this to the limited size of the adaption dataset.

\subsection{Integration over Trustworthiness Dimensions}
\label{ssec:integration}

To further compare the joint trade-off between approaches which improve different aspects of trustworthiness, we select one representative approach from each aspect of the aforementioned algorithms and integrate them in a single model to analyze the trade-off behavior. The selected algorithms are TRADES (for robustness), DAL (for fairness), MMD (for generalization), and IFR (for explainability). The integrated model is structured as Figure~\ref{fig:network}. The network takes a batch with four types of inputs, which are FFC labeled training data (blue), FFD unlabeled training data for domain generalization (yellow), masked FFC data required by IFR (green), and adversarial attack data (red) generated from FFC. The batched data are fed into a backbone used in IFR and turned into downsampled feature maps. The SAD loss used in IFR~\cite{yin2019towards} is computed from the feature maps. The feature maps are then pooled as face features. The FAD loss in IFR and MMD loss is then computed from face features. The fairness algorithm DAL is an age-invariant face recognition approach. It disentangles face features into age features which are independent from identity and identity features which are invariant against ages. The age discriminator loss and de-correlation loss (DAL regularization) are then computed on disentangled features. The identity feature is used as the final feature embedding for face verification. TRADES training is also based on adversarial examples attacking the identity features. 

Table~\ref{tab:overall} enumerates all the 16 combinations of the selected approaches and presents the performance on each trustworthy dimension. 
We pre-trained E1 on MS1M to retain good feature explainability. In Table~\ref{tab:overall}, all models with IFR enabled are fine-tuned from this pre-trained E1 backbone, and the rest are fine-tuned from the pre-trained BL backbone. 
We see several interesting phenomena from the experiment results.

\begin{itemize}
    \item The robustness algorithm is shown to have a trade-off with the accuracy performance in, e.g., R1, M20, and M21. This is consistent with our observations when robustness approaches are applied alone. This phenomenon is also consistent with previous literature such as~\cite{zhang2019theoretically} while our task is metric learning instead of classification. 
    \item The robustness algorithm is shown to increase the algorithmic bias in e.g., R1, M20, and M21. This is consistent with the known trade-off between fairness and robustness as reported in literature~\cite{xu2020robust, roh2020fr}, but the change is not significant in our experiments. 
    \item The explainability algorithm IFR (e.g., E1, M24, and M25) is shown to have good robustness to our black-box attacks. But we attribute this to the different pre-trained backbone.
    \item The fairness and generalization algorithms, DAL and MMD, shows very minor improvement on fairness and the accuracy performance of the young group. 
\end{itemize}

Though trade-offs can be observed between different combinations of algorithms, we see that the performance difference is not significant in some cases. We attribute this to the insufficient sizes of FFC and FFD, compared to MS1M. FFC is the only publicly available dataset which we find to provide identity and age labels. Thus compared to the pre-training stage, the dataset for the fine-tune stage is limited. The phenomena are observed on these specific experiment configurations and might change with larger training datasets. On this other hand, this fact also illustrates that data collection and annotation are equally crucial to build trustworthy AI compared to algorithms.

\begin{table}
    \centering
    \includegraphics[scale=0.8]{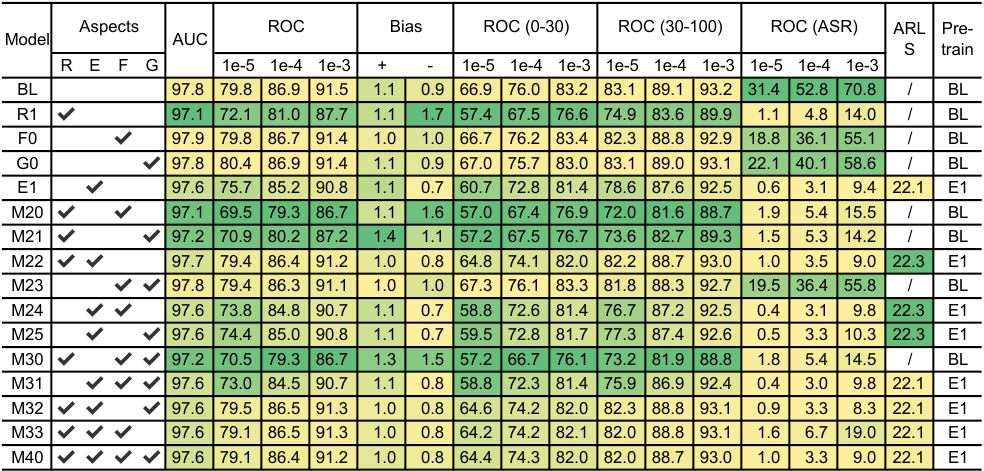}
    \caption{Performance evaluation on combinations of multiple trustworthiness aspects including robustness (R), explainability (E), fairness (F), and domain generalization (G).}
    \label{tab:overall}
\end{table}
\bibliography{main}
\bibliographystyle{plain}


\maketitle

\section{Recent Practices Trustworthy AI Practitioners}\label{sec:practices}

Diverse AI stakeholders have made efforts to enhance AI trustworthiness in the last decade. We review their recent practices in this section. We begin by introducing how various aspects of trustworthiness are being implemented in case studies of different AI applications. We then briefly review the recent progress toward trustworthy AI in multi-disciplinary areas including research, engineering, and regulation. We then show that a systematic framework is necessary to organize all these efforts. 


\begin{table}[]
\begin{tabular}{l|l|l}
\hline
Applications & Requirements & Literature \\ \hline

\multirow{5}{*}{\begin{tabular}[c]{@{}l@{}}Face\\ Recognition\end{tabular}}                                         
& Robustness & \cite{ramachandra2017presentation} \cite{chakraborty2014overview}  \cite{xiao2021improving, guo2021meaningful, komkov2021advhat, sharif2019general} \\ \cline{2-3}

& Generalization  & \cite{yang2019learning} \cite{wang2019racial} \\ \cline{2-3}

& Fairness  & \cite{drozdowski2020demographic} \cite{gong2021mitigating}  \cite{wang2019decorrelated} \\ \cline{2-3}

& Explainability  & \cite{yin2019towards, williford2020explainable} \\ \cline{2-3}

& Privacy Protection  & \cite{aggarwal2021fedface, bai2021federated}  \\ \hline 

\multirow{5}{*}{\begin{tabular}[c]{@{}l@{}}Autonomous\\ Driving\end{tabular}}

& Robustness & \cite{huang2016autonomous, horwick2010strategy} \cite{eykholt2018robust, cao2019adversarial} \cite{wood2019safety, horwick2010strategy} \\ \cline{2-3}

& Generalization  & \cite{zhang2018deeproad, sakaridis2019guided}         \\ \cline{2-3} 

& Transparency    & \cite{cysneiros2018software} \\ \cline{2-3} 

& Explainability  & \cite{zablocki2021explainability, likmeta2020combining} \\ \cline{2-3} 
& Accountability  & \cite{perez2010argos, yao2020smart} \\ \hline
\multirow{4}{*}{NLP}                                                          & Robustness      &      \cite{jia2020building}      \\ \cline{2-3} 
                                                                              & Explainability  &            \cite{wiegreffe2021teach}\\ \cline{2-3} 
                                                                              & Fairness        &           \cite{google2018reducing, tay2014stereotypes, unesco2019blush} \\ \cline{2-3} 
                                                                              & Value Alignment & \cite{chkroun2018did, park2021use, chai2020keep, elgan2018kidsalexa, rudgard2018alexachildren, xinhua2020savelife}\\ \hline
\end{tabular}
\caption{Trustworthiness is raising concern in various scenarios of AI applications. We summarize representative papers addressing different aspects of trustworthy AI in applications of face recognition, autonomous driving, and NLP.}
\label{tab:surveys}
\end{table}

\subsection{Industrial Case Study}\label{ssec:case-study}

\subsubsection{Face Recognition}
Face recognition has been one of the earliest AI techniques widely adopted in real-world scenarios. Compared to other biometric measurements, facial feature provides a much more convenient interface for users to leverage for identification. However, the recent widespread applications of face recognition bring new challenges and risks such as malicious attacks and privacy breaches. From the perspective of building trustworthy AI, we show recent representative problems and their solutions.

Face recognition has been widely adopted in security-critical scenarios such as border control and e-bank authentication. These applications raise concerns about the robustness of face recognition against malicious attacks. Naive face recognition can be easily spoofed by a photo of a person targeted for identity theft. This is known as the presentation attack on the face recognition system~\cite{ramachandra2017presentation}. Liveness detection~\cite{chakraborty2014overview} is a typical approach to defend against presentation attacks by photos or displays. However, liveness detection is still vulnerable against new types of attacks by fake videos generated by AI techniques like DeepFake~\cite{nguyen2019deep}. To comprehensively defend against presentation attacks, a number of new research directions have aroused attention from the AI community, including liveness detection, face anti-spoofing, and synthetic image detection~\cite{rossler2019faceforensics}. Besides presentation attacks, adversarial attacks against face recognition have recently been proved as a new type of threat. Research has shown that face recognition systems can be easily spoofed by impersonation attackers wearing a small printed adversarial patch~\cite{xiao2021improving, guo2021meaningful, komkov2021advhat, sharif2019general}. Concerns on this new type of risk have made the adversarial attack and defense in face recognition an important applicational research direction of adversarial ML.

Modern face recognition systems are known to be sensitive to domain shift of input data such as the change of acquisition sensors or racial groups. Conventionally the mitigation of this performance degeneracy requires a large of annotated data from different domains. To overcome the problem of data scarcity, domain generalization techniques have been adopted in face recognition scenarios to enable unsupervised training. For example, some approaches cluster unlabeled face images to generate pseudo labels for training~\cite{yang2019learning}. Domain adaption techniques are also studied to enhance the generalizability of face recognition models between different data domains~\cite{wang2019racial}.

Fairness issue is another challenge with which modern face recognition systems are confronted~\cite{hao2019making}. It is known that conventional face recognition algorithms are biased in performance on different groups of genders, races or ages~\cite{drozdowski2020demographic}. This problem not only harms the user experience of specific groups, but also harms the societal trust on AI. To mitigate bias, debiasing algorithms have recently aroused interest in face recognition research. Representatives either develop training strategies to balance performance between groups~\cite{gong2021mitigating} or extract features that are invariant across groups~\cite{wang2019decorrelated}.

Compared to the conventional image classification problem, the explainability of face recognition is less studied in the past years. However, the value of explainable face recognition should not be neglected in practical high-stake applications where the explanation of model decision helps build trust. Research has explored some potential directions to interpret face recognition procedures~\cite{yin2019towards, williford2020explainable} by enforcing internal features to express explainable structures.

Last but not the least, the sensitivity of face data arouses the concern of privacy leakage for modern face recognition systems. A typical case is that deep learning models are known to leak training face data when attacked by ``deep leakage''~\cite{zhu2019deep}. Privacy-preserving techniques like federated learning provide a solution to extract knowledge from distributed user data while preserving the identification-sensitive information~\cite{aggarwal2021fedface, bai2021federated}.


\subsubsection{Autonomous Driving}
Autonomous driving is one of the most representative AI systems applied in safety crucial scenarios. Its significant impact on social life raises global concern on ensuring its trustworthiness. Various efforts have been made in different aspects to address this problem.

Robustness is one of the most critical requirements for an autonomous driving vehicle. Poor model performance as well as external attacks~\cite{eykholt2018robust, cao2019adversarial} both threaten its safety. To address this issue, the combination of multiple sensors, including cameras and lidars, is widely adopted when designing a state-of-the-art autonomous driving vehicle~\cite{chen2017multi}. At the development stage, to validate the system robustness in a large number of traffic scenarios, simulation-based testing is widely used. Such simulation not only includes software simulation like CarSim and CARLA. but also includes hybrid realization with hardware-in-the-loop or vehicle-in-the-loop design~\cite{rosique2019systematic}. In addition, a fallback plan~\cite{wood2019safety, horwick2010strategy} is also required as an essential component on autonomous driving products. Inherited from the strict safety requirement in the automotive industry, standards like ISO 26262 also provide effective guidelines on ensuring the robustness and safety of the autonomous driving system. The safety-critical character of autonomous driving further raises the requirement of auditability. Inspired by the success of flight data recorders in aviation, event data recorders for autonomous driving~\cite{perez2010argos, yao2020smart} are proposed to collect audit trails.

Transparency and explainability are also important requirements for autonomous driving systems. At the product design stage, a clear specification of the functionality, e.g., traffic jam pilot (TJP), and autonomy level, i.e., L1--L5, provide the basic transparency of the functionality boundary of an autonomous driving system. At the algorithm level, though DNN has achieved impressive performance in autonomous driving tasks like perception and planning, its lack of explainability still places a demand to require explicit explanations or impose rules to ensure that the vehicle is behaving in a regulated manner~\cite{zablocki2021explainability, likmeta2020combining}.

Last but not least, ethics and value alignment have always been concerns along with the evolving autonomous driving technology. For example, besides human drivers, the AI driver is now also often challenged by the public with the well-known \textit{trolley dilemma}~\cite{piano2020ethical}. Based on different objectives, autonomous driving system might make different decisions to minimize the overall physical harm or protect those who obey traffic rules. Debates on different decision objectives reflect the ambiguity of human morality, especially fairness in this case, and lead to new challenges to build the trustworthiness of autonomous driving systems to the society. 


\subsubsection{Natural Language Processing}
Unlike autonomous driving, natural language processing (NLP) applications are often less safety-critical. 
However, its widely available applications have challenged trustworthiness from different aspects, especially the alignment of human values and ethics. 
As more research and media attention is directed to these issues, industry practitioners have also introduced methods to address these concerns recently.

Society has noticed a series of negative events caused by the lack of ethical consideration in conversational AI products. A notable example of the breach in trust is the prevalent gender bias in machine translation.
Specifically, when translating from languages that do not have gender distinction for third-person pronouns (e.g., Turkish) into one that does (e.g., English), translations usually exhibit bias that might stem from the data (e.g., ``He is a doctor. She is a nurse.'' even if gender cannot be inferred from the original text).
Providers such as Google have taken steps to provide alternative gendered translations in these cases to help mitigate such issues \cite{google2018reducing}. 

On the other hand, systems that are designed by taking human values into consideration can also help avoid unforeseen risks.
For instance, conversational AI systems that help serve customers in e-businesses could be equipped with flagging mechanisms for risky inquiries or unstable mental conditions of its users, and help us intervene in suicide attempts and work to improve social well-being \cite{xinhua2020savelife}.

Last but not least, voice assistants are becoming more and more prevalent in our daily lives, helping us operate devices in smart homes, as well as conveniently automating our daily workflows.
However, most commercially available voice assistants (Apple Siri, Amazon Alexa, Google Assistant, etc.) are made exclusively available in female voices, or use female personas by default \cite{tay2014stereotypes, unesco2019blush}.
This could potentially strengthen gender stereotypes, and potentially associate aggressive user behaviors with feminine service personnel in real life, which is especially concerning when a new generation of children grow up around these devices \cite{elgan2018kidsalexa, rudgard2018alexachildren}.
More recently, most major voice assistant service providers have made them available in more voice options, which helps these systems better align with human values by reducing the implicit gender inequity.

\subsection{Recent Multi-disciplinary Progresses}\label{sec:recent-multidisciplinary}

As can be seen from the above recent practices in different applications, establishing AI trustworthiness requires consideration, design, development, and governance throughout different stages of the system lifecycle to fulfill the requirements as mentioned in Section~\ref{sec:definition-evaluation}. This fact has motivated efforts from various fields to building trustworthy AI. In this section, we review the recent representative progresses, and demonstrate that a framework to organize these fields along the AI system lifecycle is a necessity.

\subsubsection{Multi-disciplinary research}
Trustworthy AI is more than a subdiscipline of computer science or ML. Its profound impact on human society has derived various research topics demanding prompt solutions.

\textbf{Societal understanding of trustworthy AI.}
The impact of AI on human society is profound and double-edged. It becomes an important problem to understand and forecast both the negative and positive impact of AI. Literature such as~\cite{kaloudi2020ai, king2020artificial, yampolskiy2019predicting} has collected and studied representative AI misuse cases, and summarized high-stake patterns that AI failures tend to follow~\cite{yampolskiy2019predicting}, such as vulnerability to attacks, underperformance on noises, and biased prediction. Understanding these patterns naturally derives a number of specific requirements of AI trustworthiness, such as robustness, generalization, and fairness. In addition to lessons learned from the AI failures, potential areas for AI to make positive impact are also important to study~\cite{tomavsev2020ai, hager2019artificial}. For example, as noted in~\cite{tomavsev2020ai, qi2021conversational}, various scenarios have been studied to apply AI to advance the United Nations' Sustainable Development Goals (SDGs) for social good.

In addition to the social impact of AI application, the societal and psychological outline of trustworthiness should be a prerequisite for the establishment of trustworthy AI. A main aspect of this study starts by analyzing the risk, and threat of AI and derives the requirements of trustworthiness~\cite{thiebes2020trustworthy}. Beyond empirical analysis, the formalization of AI trustworthiness is also studied. For example, \cite{jacovi2021formalizing} formalizes human--AI trust by considering and extending the concepts of interpersonal trust. In terms of specific dimensions of trustworthy AI, previous works such as~\cite{gunning2019darpa, mohseni2021multidisciplinary, miller2019explanation} have studied the problem of explainability and transparency from philosophy, cognitive psychology, and social psychology. Analysis of ethical values regarding AI application is also an important problem in the aspect of social science and can be found in recent literature such as~\cite{piano2020ethical, greene2019better, bankins2021ethical, mittelstadt2019principles}.








\textbf{Technical approaches.} As is shown in Section~\ref{sec:definition-evaluation}, AI trustworthiness is decomposed as the fulfillment of a series of requirements. It is recognized that technology improvement serves as a core solution to many of these requirements, such as robustness, generalization, explainability, fairness, and privacy. In the last decade, these requirements have derived a variety of active directions in AI research. Recently there has been some literature that briefly surveys the related technology to facilitate the establishment of trustworthy AI. For example, \cite{toreini2020relationship, liu2021trustworthy} list example trustworthy research directions in the area of ML algorithms. \cite{wickramasinghe2020trustworthy} surveys trustworthy AI development techniques from the aspect of human--system interaction. \cite{kumar2020trustworthy} comprehensively decomposes requirements of trustworthy AI to executable techniques that are of the current interest in the community of AI research. Due to the wide coverage of trustworthy AI related techniques, it is difficult to exhaustively survey its every dimension in a single paper. Fortunately, there have been surveys available that covers several subdivisions of research mentioned in this paper. We refer to these surveys in Sections~\ref{sec:definition-evaluation} and \ref{sec:systematic-approach}.

The establishment of trustworthy AI systems involves the development of both AI algorithms and engineering techniques. Take the requirement of robustness as an example. Adversarial attack and defense are becoming popular research directions. They have been studied in various high-stake tasks such as face recognition~\cite{komkov2021advhat}, autonomous driving~\cite{cao2019adversarial}, and recommendation systems ~\cite{li2016data}. Besides algorithms to enhance model robustness against attacks, engineering techniques are increasingly raising concern to enhance system robustness. At the development stage of the system, comprehensive testing on malicious inputs serves as an essential guarantee to avoid malfunction on attacks. At the deployment stage, a fail-safe mechanism prevents the system to cause loss even if malfunction happens. 



\textbf{Non-technical approaches.}
The achievement of trustworthy AI requires more than just technical approaches. The creation and operation of AI systems involve the participation of multiple roles including researchers, software developers, business operators, and regulators. The systems' lifecycle ranges over the steps of product design, algorithm design, system development, deployment, and business operation. It becomes a prerequisite of trustworthy AI to align the understanding and efforts of these various roles and steps, which raises the demand for appropriate management strategies.

Literature such as~\cite{shneiderman2020bridging, baker2021trustworthy, rakova2021responsible} has pointed a series of reforms to the organization of  participants and procedures to build trustworthy AI. These reforms include responsible product design aligned with trustworthiness principles~\cite{amershi2019guidelines}, new human resource strategies~\cite{shneiderman2020bridging, rakova2021responsible}, and justification of trustworthiness via continuous auditing~\cite{shneiderman2020bridging}. We continue to consider the aforementioned example of enhancing AI robustness. Algorithm researchers and software engineers should be aligned with the same requirement of robustness so that they apply appropriate techniques in their area to prevent the risk of attacks. In high-stake applications, safety engineers~\cite{siddique2020safetyops} should be involved to oversight the lifecycle of a system and assess potential risks. Further, business operators should be aware of the working condition and limitation of the system to avoid exceptional use cases. When a new type of attacks happen, feedback from business operators to system manufacturers serves as a prerequisite to upgrade the system against the problem. Besides direct participants in the system, third-party auditors provide independent certification of robustness and security, which serves as an external driving force to the fulfillment of trustworthiness requirements~\cite{zicari2021z, raji2019ml}.



\subsubsection{Engineering practices}
With the development of the study on trustworthy AI, a number of common approaches in building AI trustworthiness have been organized as tools to address various aspects of AI trustworthiness. 

\textbf{Development tools.} Various toolboxes have been released in the past years to assist the algorithm experimentation and software development to enhance AI trustworthiness in different aspects such as fairness~\cite{bellamy2018ai, bird2020fairlearn, saleiro2018aequitas}, XAI~\cite{nori2019interpretml, github-xai,arya2019one}, and adversarial robustness~\cite{nicolae2018adversarial, github-robustness}. For the management stage, tools and references are released to facilitate the improvement of documentation and thus enhance transparency~\cite{arnold2019factsheets}. Development platforms such as CodaLab provide workspaces for reproducible AI experimentation and development. Tools to test and audit the system trustworthiness in specific requirements such as fairness~\cite{wilson2021building} provide further verification to the system development.


To manage and evaluate different tools, the OECD Network of Experts on Implementing Trustworthy AI is working on a generic framework to help practitioners choose suitable tools for their specific use case~\cite{OECD-tools}. The OECD has conducted a survey and shows that three types of tools are of the common focus: 1) Technical tools aiming to address technical aspects such as the aforementioned robustness, explainability, and bias detection; 2) Procedural tools providing implementation guidance; 3) Educational tools training stakeholders of the AI systems.  



\textbf{Development frameworks.}
As can be seen from the above examples and categorization, existing tools reflect the consideration of trustworthy AI at different stages in the lifecycle of an AI system, such as algorithm design, development, and management~\cite{aihleg2018ethics}. Each of these tools mainly focuses on a specific requirement of trustworthy AI and improves a specific stage of the lifecycle. To build a complete trustworthy AI system, there is a necessity to select and organize these tools systematically.

Some institutions have released their collection of tools as a framework to facilitate the development of trustworthy AI systems, e.g., Amazon SageMaker, IBM OpenScale. We take Google's Responsible AI with TensorFlow as a representative practical collection of modules considering requirements of trustworthy AI at multiple stages when building AI systems~\cite{google-ai}. The People+AI Research (PAIR) Guidebook provides guidelines and best practices to better understand the human factors when designing an AI product. At the model development stage, libraries and tools are provided within the framework to enhance the explainability, fairness, and privacy of the model. Tool-chains are also available to track the meta information regarding the training process~\cite{mitchell2019model}. Responsible AI reflects the consideration of AI trustworthiness from the perspective of the AI system lifecycle. It provides well-developed libraries that can be directly adopted at the stages of problem definition, data preparation, model training, model evaluation, and model deployment. However, considering the case studies aforementioned in this section, this collection of tools is still insufficient to guide and support the industrial practices to address trustworthy AI through the system lifecycle.

\subsubsection{Institutionalization}
    \label{sec:institutionalization}
Though the value of trustworthy AI has been widely recognized, the motivation, interpretation, and consideration on AI trustworthiness vary over different stakeholders. Such inconsistency is even more evident in the industry due to the commercial consideration. Even within a single company, different departments of such as business, technology, and legality are likely to hold different interpretation on requirements of AI trustworthiness. To address this issue, guidelines, standards, and regulations become necessary to unify the understanding of different AI stakeholders.

\textbf{Guidelines.} To regulate the development of AI, a variety of guidelines and principles have been proposed by governmental organizations in recent years~\cite{aihleg2018ethics, holdren2016preparing, beijing2019ethics, union2017top, lewis2021ontology}. Representatives of these guidelines include the EU Ethics Guidelines for Trustworthy AI~\cite{aihleg2018ethics}. Besides governmental organizations, a wealth of companies also release their guidelines or principles to demonstrate their compliance with AI trustworthiness~\cite{google-ai, openai2018openai, ibm-trusting, cammarota2020trustworthy, brundage2020toward, varshney2019trustworthy}. Thorough review and discussion of these guidelines can be found from \cite{jobin2019global, schiff2021ai, hagendorff2020ethics}. Though differences exist, it is shown that global convergence on the requirement of AI trustworthiness is recognized as summarized in Section~\ref{sec:definition-evaluation}.


\textbf{Standardization.} With various ethics guidelines released globally as soft recommendations for AI development and converges to consensus principles, standards to concretize these principles in specific AI application areas serve as the further means to regulate the trustworthiness of AI products~\cite{zielke2020artificial}. Multiple standardization activities are ongoing parallelly, including IEEE P7000 and ISO/IEC SC 42~\cite{lewis2021ontology}, which specifically focuses on AI trustworthiness. There are multiple different aspects addressed in the current standardization. For example, at the holistic level, ISO/IEC SC 42 Working Group~3 (WG3) works on establishing trustworthiness for general ML systems through transparency, robustness, explainability, and so on~\cite{lewis2021ontology}.
Besides generic AI-centered standards, certain applications of AI should also seek compliance with domain-specific standards to build trustworthy AI products, e.g., ISO 26262 for the functional safety of intelligent vehicles and ISO 14155 for clinical practice on medical devices.


\textbf{Regulation.} Governmental legislation is catching up its pace with the rapid development of AI technology and industrial application. Recently EU releases its proposal for Artificial Intelligence Act~\cite{devulder2021world}. A notable point of the act is the requirement to execute risk management approaches throughout different stages in the lifecycle of an AI system. These approaches include but not limited to data governance, testing for risks, detailed documentation, logging, and human oversight. Besides regulation on AI systems, it should also be noted that a wealth of regulation in the area of general information technology is also bringing significant impact on the AI industry, such as GDPR, the Chinese Cybersecurity Law, and the Personal Information Security Specification. 

\subsubsection{Toward systematic organization}
As can be seen from the above recent practices, the establishment of trustworthy AI requires efforts at different stages in the lifecycle of an AI system. This fact motivates the need to organize all these fragmented efforts in a unified framework. A number of works have conducted studies on such frameworks, mainly on a specific area such as technique or management. For example, \cite{kumar2020trustworthy} briefly reviews executive techniques and organizes them by the specified requirement dimensions of trustworthy AI.   \cite{wickramasinghe2020trustworthy} categorizes the interactions between AI systems, developers, and users to organize the technical approaches and responsibility in the lifecycle of trustworthy AI. \cite{thiebes2020trustworthy} simplifies the system lifecycle as the flow of data and technically analyzes threats existing at each flow stage. The aforementioned tool-chains such as Google Responsible AI also reflect the idea of organizing existing techniques in the frame of system development lifecycle. In terms of management, \cite{baker2021trustworthy} studies a management workflow to guarantee that ethics are aligned throughout the creation and update of an AI system. \cite{raji2020closing} specifically organizes auditing approaches for the management of system lifecycle.

Whilst the above existing literature has studied the organization of technical and management approaches to build trustworthy AI, we argue that these works are still insufficient from the industrial perspective. Considering the case study of e.g., autonomous driving in Section~\ref{ssec:case-study}, this research of frameworks has not yet covered all aspects that should be addressed.
As depicted in Section~\ref{sec:introduction} and Figure~\ref{fig:lifecycle}, the AI industry holds a connecting position to organize efforts from multi-disciplinary fields in the establishment of trustworthy AI. This position requires industrial stakeholders to learn and organize the diverse approaches from areas of academic research, engineering, and management so that trustworthy AI can be ensured throughout the lifecycle of products. By surveying the above previous works, we find a framework covering all these areas is still lacking. 
In Section~\ref{sec:systematic-approach}, we push a step forward toward this requirement by a multi-disciplinary survey of approaches to build trustworthy AI. These approaches cover active research areas, engineering techniques, and management strategies, and are organized in the frame of an industrial system lifecycle. A workflow is then proposed to integrate these approaches and facilitate dynamic feedback and upgrade from academia and governmental regulation over these approaches.

\bibliography{main}
\bibliographystyle{plain}